\documentclass[10pt,twocolumn,letterpaper]{article}

\usepackage[pagenumbers]{cvpr} % To force page numbers, e.g. for an arXiv version

\usepackage{graphicx}
\usepackage{amsmath}
\usepackage{amssymb}
\usepackage{booktabs}

\usepackage{mathtools}
\usepackage{amsthm}

\usepackage{multirow}
\usepackage[dvipsnames]{xcolor}

\theoremstyle{plain}

\theoremstyle{definition}

\theoremstyle{remark}

\definecolor{cvprblue}{rgb}{0.21,0.49,0.74}
\usepackage[pagebackref,breaklinks,colorlinks,citecolor=cvprblue]{hyperref}

\usepackage[accsupp]{axessibility}  % Improves PDF readability for those with disabilities.

\usepackage{xcolor, soul}
\sethlcolor{pink}

\usepackage{pgfplots}
    \usepgfplotslibrary{colorbrewer}
    \pgfplotsset{
        cycle list/Dark2,
        cycle multiindex* list={
            mark list*\nextlist
            Dark2\nextlist
        },
    }
\pgfplotsset{compat=1.14}

\definecolor{ForestGreen}{RGB}{34,139,34}
\definecolor{Cerulean}{rgb}{0.0, 0.48, 0.65}
\definecolor{OrangeRed}{RGB}{255,69,0}
\definecolor{OrangeRed2}{RGB}{240,9,30}
\definecolor{goldenpoppy}{rgb}{0.99, 0.76, 0.0}
\definecolor{goldenpoppy3}{rgb}{0.99, 0.46, 0.1}
\definecolor{skyblue}{rgb}{0.53, 0.81, 0.92}
\definecolor{skyblue2}{rgb}{0.55, 0.83, 0.94}
\definecolor{red-violet}{rgb}{0.78, 0.08, 0.52}
\definecolor{darkcerulean}{rgb}{0.03, 0.33, 0.55}
\definecolor{darkcerulean2}{rgb}{0.00, 0.20, 0.35}
\definecolor{flamingopink}{rgb}{0.99, 0.56, 0.67}
\definecolor{caribbeangreen}{rgb}{0.0, 0.8, 0.6}
\definecolor{caribbeangreen2}{rgb}{0.0, 0.825, 0.625}
\definecolor{darkpastelpurple}{rgb}{0.69, 0.34, 0.84}
\definecolor{darkpastelpurple2}{rgb}{0.68, 0.33, 0.83}
\definecolor{smokyblack}{rgb}{0.1, 0.1, 0.1}
\usepackage{standalone}

\usepackage{rotating}

\usepackage{booktabs}
\usepackage{array}
\newcolumntype{P}[1]{>{\raggedright\arraybackslash}p{#1}}
\usepackage{makecell}      % easy multi-line cells
\title{Benchmarking Pathology Foundation Models\\ for Breast Cancer Survival Prediction}

\author{Fredrik K. Gustafsson$^{1,2}$
\and
Constance Boissin$^{1}$
\and
Johan Vallon-Christersson$^{3}$
\and
David A. Clifton$^{2,4}$
\and
Mattias Rantalainen$^{*,1}$\vspace{1.0mm}
\and
\normalsize$^{1}$Department of Medical Epidemiology and Biostatistics, Karolinska Institutet, Stockholm, Sweden\vspace{-0.85mm}\\
\normalsize$^{2}$Department of Engineering Science, University of Oxford, Oxford, UK\vspace{-0.85mm}\\
\normalsize$^{3}$Division of Oncology, Department of Clinical Sciences Lund, Lund University, Lund, Sweden\vspace{-0.85mm}\\
\normalsize$^{4}$Oxford Suzhou Centre for Advanced Research, University of Oxford, Suzhou, China\vspace{-0.6mm}\\
\small$^{*}$Corresponding Author, {\tt mattias.rantalainen@ki.se}
}

\begin{document}

\maketitle

\begin{abstract}
    Pathology foundation models (PFMs) have recently emerged as powerful pretrained encoders for computational pathology, enabling transfer learning across a wide range of downstream tasks. However, systematic comparisons of these models for clinically meaningful prediction problems remain limited, especially in the context of survival prediction under external validation. In this study, we benchmark widely used and recently proposed PFMs for breast cancer survival prediction from whole-slide histopathology images. Using a standardized pipeline based on patch-level feature extraction and a unified survival modeling framework, we evaluate model representations across three independent clinical cohorts comprising more than 5{,}400 patients with long-term follow-up. Models are trained on one cohort and evaluated on two independent external cohorts, enabling a rigorous assessment of cross-dataset generalization. Overall, H-optimus-1 achieves the strongest survival prediction performance. More broadly, we observe consistent generational improvements across model families, with second-generation PFMs outperforming their first-generation counterparts. However, absolute performance differences between many recent PFMs remain modest, suggesting diminishing returns from further scaling of pretraining data or model size alone. Notably, the compact distilled model H0-mini slightly outperforms its larger teacher model H-optimus-0, despite using fewer than 8\% of the parameters and enabling significantly faster feature extraction. Together, these results provide the first large-scale, externally validated benchmark of PFMs for breast cancer survival prediction, and offer practical guidance for efficient deployment of PFMs in clinical workflows.
\end{abstract}
\vspace{-12.0mm}

\section*{}

% Intro/background/contributions:
Computational pathology has recently undergone a rapid transformation driven by the emergence of large-scale pathology foundation models (PFMs)~\citep{gigapath2024, chen2024uni, virchow2024, conch2024, hoptimus0, ding2025multimodal, bilal2025foundation, li2025survey}. Trained on hundreds of thousands to millions of whole-slide images (WSIs), these models aim to learn generalizable visual representations that can be transferred across a wide range of downstream tasks, including tumor grading~\citep{wang2022improved, bulten2022artificial}, biomarker prediction~\citep{niehues2023generalizable, arslan2024systematic}, and patient prognosis~\citep{hohn2023colorectal, jiang2024end, volinsky2024prediction}. This paradigm shift mirrors the success of foundation models in natural language processing and computer vision~\citep{bommasani2021opportunities, azad2023foundational, moor2023foundation}, positioning PFMs as a key component of modern computational pathology workflows.

While numerous PFMs have recently been proposed, their relative performance on clinically meaningful prediction tasks remains incompletely understood. Recent benchmarking efforts have begun to address this gap, but with certain limitations. \citet{marza2025thunder} provide a comprehensive comparison of more than 20 models but exclusively focus on patch-level analysis, which limits relevance for downstream clinical prediction tasks. \citet{kasireddy2026comprehensive} evaluate PFMs on both patch- and slide-level kidney pathology tasks, but the slide-level cohorts remain relatively small (at most $\approx$200 WSIs per task). \citet{breen2025comprehensive} provide a rigorous single-task evaluation of PFMs for ovarian cancer subtyping, but focus exclusively on a classification setting rather than broader clinical endpoints. Similarly, \citet{bareja2025evaluating} conduct a large-scale comparison of models across various datasets and tasks, but restrict their evaluation to classification-based endpoints, without considering time-to-event modeling for prognosis. 

\begin{figure*}[t]
    \centering
    \begin{tikzpicture}
  \centering
  \begin{axis}[
        ybar=4.0pt, axis on top,
        title={},
        % height=4.525cm, width=18.0cm, %%%%%%%%%%%%%%%%%%%%%%%%%%%%%%%%%%%%%
        % height=5.5cm, width=24.75cm, %%%%%%%%%%%%%%%%%%%%%%%%%%%%%%%%%%%%%
        height=5.75cm, width=24.75cm, %%%%%%%%%%%%%%%%%%%%%%%%%%%%%%%%%%%%%
        bar width=0.525cm, %%%%%%%%%%%%%%%%%%%%%%%%%%%%%%%%%%%%%%%%%
        ymajorgrids, tick align=inside,
        enlarge y limits={value=.1,upper},
        ymin=0.50, ymax=0.7175,
        set layers,
        ytick={0.50, 0.55, 0.6, 0.65, 0.70},
        yticklabels={0.50, 0.55, 0.60, 0.65, 0.70},
        axis x line*=bottom,
        axis y line*=left,
        y axis line style={opacity=0},
        tickwidth=0pt,
        enlarge x limits=0.575, % (makes the different bar groups closer to eachother!) %%%%%
        extra y tick style={
            grid style={
                % black,
                solid,
                line width = 2.5pt,
                /pgfplots/on layer=axis background,
            },
        },
        tick style={
            grid style={
                dashed,
                /pgfplots/on layer=axis background,
            },
        },
        legend style={
            at={(0.5,-0.20)},
            anchor=north,
            legend columns=-1,
            /tikz/every even column/.append style={column sep=0.175cm},
            font=\footnotesize
        },
        ylabel={C-index ($\uparrow$)},
        ylabel style={font=\normalsize},
        symbolic x coords={
           RFS -- All Patients,
           RFS -- Patient Subgroup (ER+ \& HER2-)},
       xtick=data,
       % xticklabel=\empty,
       % x tick label style={text width = 7.0cm, align=center, font={\footnotesize\bfseries}},
       x tick label style={text width = 7.0cm, align=center, font={\bfseries}},
    ]
    \addplot [fill=gray, draw=gray, line width=0mm, error bars/.cd, y dir=both, y explicit, error bar style={thick}, error bar style=black] coordinates { 
      (RFS -- All Patients, 0.6119) += (0, 0.0229) -= (0, 0.0236)
      (RFS -- Patient Subgroup (ER+ \& HER2-), 0.6001) += (0, 0.0269) -= (0, 0.0276)};
    \addplot [fill=OrangeRed2, draw=OrangeRed2, error bars/.cd, y dir=both, y explicit, error bar style={thick}, error bar style=black] coordinates {
      (RFS -- All Patients, 0.6321) += (0, 0.0227) -= (0, 0.0230)
      (RFS -- Patient Subgroup (ER+ \& HER2-), 0.6269) += (0, 0.0254) -= (0, 0.0261)};
    \addplot [fill=darkgray, draw=darkgray, error bars/.cd, y dir=both, y explicit, error bar style={thick}, error bar style=black] coordinates {
      (RFS -- All Patients, 0.6449) += (0, 0.0228) -= (0, 0.0229)
      (RFS -- Patient Subgroup (ER+ \& HER2-), 0.6357) += (0, 0.0258) -= (0, 0.0263)};
    \addplot [fill=flamingopink, draw=flamingopink, error bars/.cd, y dir=both, y explicit, error bar style={thick}, error bar style=black] coordinates {
      (RFS -- All Patients, 0.6573) += (0, 0.0227) -= (0, 0.0226)
      (RFS -- Patient Subgroup (ER+ \& HER2-), 0.6480) += (0, 0.0257) -= (0, 0.0264)};
    \addplot [fill=red-violet, draw=red-violet, error bars/.cd, y dir=both, y explicit, error bar style={thick}, error bar style=black] coordinates {
      (RFS -- All Patients, 0.6665) += (0, 0.0226) -= (0, 0.0228)
      (RFS -- Patient Subgroup (ER+ \& HER2-), 0.6629) += (0, 0.0254) -= (0, 0.0261)};
    \addplot [fill=skyblue2, draw=skyblue2, error bars/.cd, y dir=both, y explicit, error bar style={thick}, error bar style=black] coordinates {
      (RFS -- All Patients, 0.6639) += (0, 0.0223) -= (0, 0.0224)
      (RFS -- Patient Subgroup (ER+ \& HER2-), 0.6595) += (0, 0.0251) -= (0, 0.0263)};
    \addplot [fill=Cerulean, draw=Cerulean, error bars/.cd, y dir=both, y explicit, error bar style={thick}, error bar style=black] coordinates {
      (RFS -- All Patients, 0.6783) += (0, 0.0218) -= (0, 0.0220)
      (RFS -- Patient Subgroup (ER+ \& HER2-), 0.6703) += (0, 0.0248) -= (0, 0.0258)};
    \addplot [fill=darkcerulean2, draw=darkcerulean2, error bars/.cd, y dir=both, y explicit, error bar style={thick}, error bar style=black] coordinates {
      (RFS -- All Patients, 0.6758) += (0, 0.0227) -= (0, 0.0230)
      (RFS -- Patient Subgroup (ER+ \& HER2-), 0.6677) += (0, 0.0252) -= (0, 0.0258)};
    \addplot [fill=darkpastelpurple2, draw=darkpastelpurple2, error bars/.cd, y dir=both, y explicit, error bar style={thick}, error bar style=black] coordinates {
      (RFS -- All Patients, 0.6629) += (0, 0.0231) -= (0, 0.0224)
      (RFS -- Patient Subgroup (ER+ \& HER2-), 0.6566) += (0, 0.0251) -= (0, 0.0254)};
    \addplot [fill=goldenpoppy, draw=goldenpoppy, error bars/.cd, y dir=both, y explicit, error bar style={thick}, error bar style=black] coordinates {
      (RFS -- All Patients, 0.6556) += (0, 0.0228) -= (0, 0.0231)
      (RFS -- Patient Subgroup (ER+ \& HER2-), 0.6392) += (0, 0.0260) -= (0, 0.0266)};
    \addplot [fill=goldenpoppy3, draw=goldenpoppy3, error bars/.cd, y dir=both, y explicit, error bar style={thick}, error bar style=black] coordinates {
      (RFS -- All Patients, 0.6752) += (0, 0.0225) -= (0, 0.0227)
      (RFS -- Patient Subgroup (ER+ \& HER2-), 0.6648) += (0, 0.0252) -= (0, 0.0258)};
    \addplot [fill=caribbeangreen2, draw=caribbeangreen2, error bars/.cd, y dir=both, y explicit, error bar style={thick}, error bar style=black] coordinates {
      (RFS -- All Patients, 0.6631) += (0, 0.0222) -= (0, 0.0222)
      (RFS -- Patient Subgroup (ER+ \& HER2-), 0.6566) += (0, 0.0248) -= (0, 0.0252)};
    \addplot [fill=ForestGreen, draw=ForestGreen, error bars/.cd, y dir=both, y explicit, error bar style={thick}, error bar style=black] coordinates {
      (RFS -- All Patients, 0.6601) += (0, 0.0225) -= (0, 0.0227)
      (RFS -- Patient Subgroup (ER+ \& HER2-), 0.6508) += (0, 0.0248) -= (0, 0.0255)};
  \end{axis}
  \end{tikzpicture}\vspace{0.0mm}
    \begin{tikzpicture}
  \centering
  \begin{axis}[
        ybar=4.0pt, axis on top,
        title={},
        % height=4.525cm, width=18.0cm, %%%%%%%%%%%%%%%%%%%%%%%%%%%%%%%%%%%%%
        % height=5.5cm, width=24.75cm, %%%%%%%%%%%%%%%%%%%%%%%%%%%%%%%%%%%%%
        height=5.75cm, width=24.75cm, %%%%%%%%%%%%%%%%%%%%%%%%%%%%%%%%%%%%%
        bar width=0.525cm, %%%%%%%%%%%%%%%%%%%%%%%%%%%%%%%%%%%%%%%%%
        ymajorgrids, tick align=inside,
        enlarge y limits={value=.1,upper},
        ymin=0.50, ymax=0.7175,
        set layers,
        ytick={0.50, 0.55, 0.6, 0.65, 0.70},
        yticklabels={0.50, 0.55, 0.60, 0.65, 0.70},
        axis x line*=bottom,
        axis y line*=left,
        y axis line style={opacity=0},
        tickwidth=0pt,
        enlarge x limits=0.575, % (makes the different bar groups closer to eachother!) %%%%%
        extra y tick style={
            grid style={
                % black,
                solid,
                line width = 2.5pt,
                /pgfplots/on layer=axis background,
            },
        },
        tick style={
            grid style={
                dashed,
                /pgfplots/on layer=axis background,
            },
        },
        % ticklabel style = {font=\large},
        legend style={
            at={(0.5,-0.175)},
            anchor=north,
            legend columns=-1,
            /tikz/every even column/.append style={column sep=0.175cm},
            font=\footnotesize
        },
        ylabel={C-index ($\uparrow$)},
        ylabel style={font=\normalsize},
        symbolic x coords={
           PFS -- All Patients,
           PFS -- Patient Subgroup (ER+ \& HER2-)},
       xtick=data,
       % x tick label style={text width = 7.0cm, align=center, font={\footnotesize\bfseries}},
       x tick label style={text width = 7.0cm, align=center, font={\bfseries}},
    ]
    \addplot [fill=gray, draw=gray, line width=0mm, error bars/.cd, y dir=both, y explicit, error bar style={thick}, error bar style=black] coordinates { 
      (PFS -- All Patients, 0.6280) += (0, 0.0378) -= (0, 0.0383)
      (PFS -- Patient Subgroup (ER+ \& HER2-), 0.6060) += (0, 0.0488) -= (0, 0.0489)};
    \addplot [fill=OrangeRed2, draw=OrangeRed2, error bars/.cd, y dir=both, y explicit, error bar style={thick}, error bar style=black] coordinates {
      (PFS -- All Patients, 0.6470) += (0, 0.0385) -= (0, 0.0380)
      (PFS -- Patient Subgroup (ER+ \& HER2-), 0.6220) += (0, 0.0478) -= (0, 0.0490)};
    \addplot [fill=darkgray, draw=darkgray, error bars/.cd, y dir=both, y explicit, error bar style={thick}, error bar style=black] coordinates {
      (PFS -- All Patients, 0.6273) += (0, 0.0386) -= (0, 0.0383)
      (PFS -- Patient Subgroup (ER+ \& HER2-), 0.6042) += (0, 0.0484) -= (0, 0.0492)};
    \addplot [fill=flamingopink, draw=flamingopink, error bars/.cd, y dir=both, y explicit, error bar style={thick}, error bar style=black] coordinates {
      (PFS -- All Patients, 0.6708) += (0, 0.0368) -= (0, 0.0370)
      (PFS -- Patient Subgroup (ER+ \& HER2-), 0.6377) += (0, 0.0452) -= (0, 0.0457)};
    \addplot [fill=red-violet, draw=red-violet, error bars/.cd, y dir=both, y explicit, error bar style={thick}, error bar style=black] coordinates {
      (PFS -- All Patients, 0.6832) += (0, 0.0355) -= (0, 0.0361)
      (PFS -- Patient Subgroup (ER+ \& HER2-), 0.6570) += (0, 0.0446) -= (0, 0.0455)};
    \addplot [fill=skyblue2, draw=skyblue2, error bars/.cd, y dir=both, y explicit, error bar style={thick}, error bar style=black] coordinates {
      (PFS -- All Patients, 0.6860) += (0, 0.0356) -= (0, 0.0354)
      (PFS -- Patient Subgroup (ER+ \& HER2-), 0.6695) += (0, 0.0447) -= (0, 0.0452)};
    \addplot [fill=Cerulean, draw=Cerulean, error bars/.cd, y dir=both, y explicit, error bar style={thick}, error bar style=black] coordinates {
      (PFS -- All Patients, 0.7019) += (0, 0.0352) -= (0, 0.0357)
      (PFS -- Patient Subgroup (ER+ \& HER2-), 0.6703) += (0, 0.0446) -= (0, 0.0448)};
    \addplot [fill=darkcerulean2, draw=darkcerulean2, error bars/.cd, y dir=both, y explicit, error bar style={thick}, error bar style=black] coordinates {
      (PFS -- All Patients, 0.6915) += (0, 0.0361) -= (0, 0.0363)
      (PFS -- Patient Subgroup (ER+ \& HER2-), 0.6655) += (0, 0.0458) -= (0, 0.0464)};
    \addplot [fill=darkpastelpurple2, draw=darkpastelpurple2, error bars/.cd, y dir=both, y explicit, error bar style={thick}, error bar style=black] coordinates {
      (PFS -- All Patients, 0.6732) += (0, 0.0365) -= (0, 0.0370)
      (PFS -- Patient Subgroup (ER+ \& HER2-), 0.6409) += (0, 0.0457) -= (0, 0.0466)};
    \addplot [fill=goldenpoppy, draw=goldenpoppy, error bars/.cd, y dir=both, y explicit, error bar style={thick}, error bar style=black] coordinates {
      (PFS -- All Patients, 0.6948) += (0, 0.0343) -= (0, 0.0346)
      (PFS -- Patient Subgroup (ER+ \& HER2-), 0.6781) += (0, 0.0420) -= (0, 0.0441)};
    \addplot [fill=goldenpoppy3, draw=goldenpoppy3, error bars/.cd, y dir=both, y explicit, error bar style={thick}, error bar style=black] coordinates {
      (PFS -- All Patients, 0.6813) += (0, 0.0364) -= (0, 0.0365)
      (PFS -- Patient Subgroup (ER+ \& HER2-), 0.6546) += (0, 0.0451) -= (0, 0.0468)};
    \addplot [fill=caribbeangreen2, draw=caribbeangreen2, error bars/.cd, y dir=both, y explicit, error bar style={thick}, error bar style=black] coordinates {
      (PFS -- All Patients, 0.6751)+= (0, 0.0353) -= (0, 0.0357)
      (PFS -- Patient Subgroup (ER+ \& HER2-), 0.6561) += (0, 0.0439) -= (0, 0.0442)};
    \addplot [fill=ForestGreen, draw=ForestGreen, error bars/.cd, y dir=both, y explicit, error bar style={thick}, error bar style=black] coordinates {
      (PFS -- All Patients, 0.6948) += (0, 0.0345) -= (0, 0.0353)
      (PFS -- Patient Subgroup (ER+ \& HER2-), 0.6797) += (0, 0.0416) -= (0, 0.0438)};
    \legend{
            Resnet-IN,
            CTransPath,
            RetCCL,
            UNI,
            UNI2-h,
            H-optimus-0,
            H-optimus-1,
            H0-mini,
            Prov-GigaPath,
            Virchow,
            Virchow2,
            CONCH,
            CONCHv1.5
            }
  \end{axis}
  \end{tikzpicture}

% Resnet-IN,
% Phikon-v2,
% UNI,
% UNI2-h,
% H-optimus-0,
% H-optimus-1,
% H0-mini,
% Prov-Gigapath,
% Virchow,
% Virchow2,
% CONCH,
% CONCHv1.5
\vspace{-1.0mm}
    \caption{\textbf{Main model comparison across evaluation settings}, showing performance in terms of C-index ($\uparrow$) for all thirteen evaluated models. Models are evaluated for recurrence-free survival (RFS) and progression-free survival (PFS), each assessed both for the full cohort (`All Patients') and the `ER+ \& HER2-' patient subgroup. Bars show the bootstrap mean C-index with 95\% confidence intervals.}
    \label{fig:main_results}
\end{figure*}
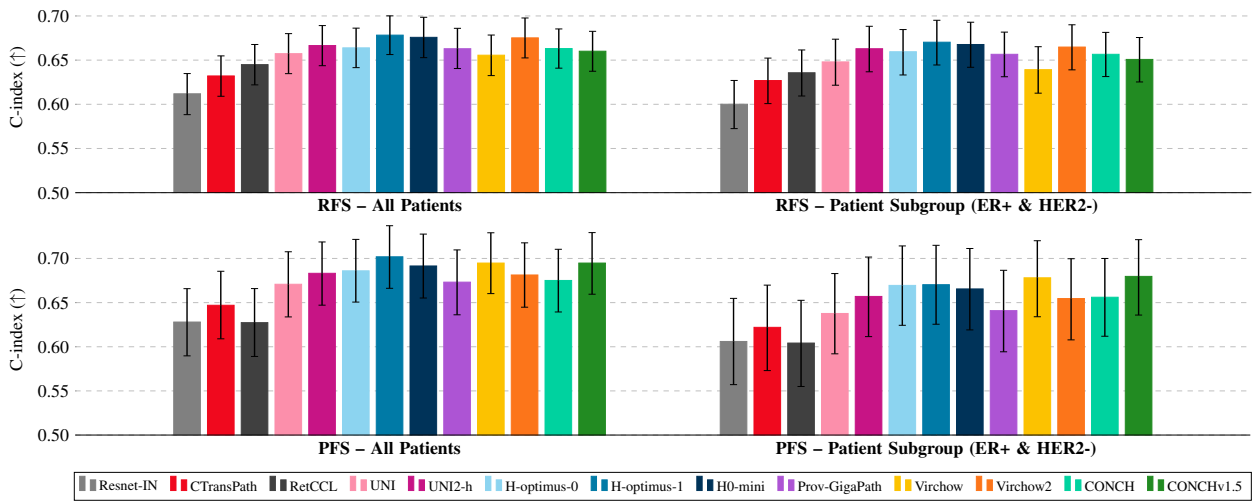

\begin{figure*}[t]
    \centering
    \begin{minipage}{0.85\linewidth}
        \centering
        \captionof{table}{\textbf{Main model comparison with numerical results and model rankings across evaluation settings}, reporting performance in terms of C-index (bootstrap mean with 95\% confidence intervals) for all thirteen evaluated models. Models are separately ranked based on the bootstrap mean within each of the four evaluation settings.}\vspace{-1.5mm}
        \label{table:main_results}
        \begin{subtable}{0.49\linewidth}
        \centering
        \vskip -0.05in
        \resizebox{1.0\linewidth}{!}{
\begin{tabular}{cl@{\hspace{1.2cm}}c}
\toprule
Rank &Model Name &C-index ($\uparrow$)\\
\midrule
1 &H-optimus-1     &0.678\hspace{2.5mm}(0.656 -- 0.700)\\
2 &H0-mini         &0.676\hspace{2.5mm}(0.653 -- 0.698)\\
3 &Virchow2        &0.675\hspace{2.5mm}(0.653 -- 0.698)\\
4 &UNI2-h          &0.667\hspace{2.5mm}(0.644 -- 0.689)\\
5 &H-optimus-0     &0.664\hspace{2.5mm}(0.642 -- 0.686)\\
6 &CONCH           &0.663\hspace{2.5mm}(0.641 -- 0.685)\\
6 &Prov-GigaPath   &0.663\hspace{2.5mm}(0.640 -- 0.686)\\
8 &CONCHv1.5       &0.660\hspace{2.5mm}(0.637 -- 0.683)\\
9 &UNI             &0.657\hspace{2.5mm}(0.635 -- 0.680)\\
10 &Virchow        &0.656\hspace{2.5mm}(0.633 -- 0.679)\\
11 &RetCCL         &0.645\hspace{2.5mm}(0.622 -- 0.668)\\
12 &CTransPath     &0.632\hspace{2.5mm}(0.609 -- 0.655)\\
13 &Resnet-IN      &0.612\hspace{2.5mm}(0.588 -- 0.635)\\
\bottomrule
\end{tabular}
        }
        \caption{\textbf{RFS -- All Patients}.}
        \label{table:main_results_rfs_all}
        \end{subtable}
        \hfill
        \begin{subtable}{0.49\linewidth}
        \centering
        \vskip -0.05in
        \resizebox{1.0\linewidth}{!}{
\begin{tabular}{cl@{\hspace{1.2cm}}c}
\toprule
Rank &Model Name &C-index ($\uparrow$)\\
\midrule
1 &H-optimus-1     &0.670\hspace{2.5mm}(0.645 -- 0.695)\\
2 &H0-mini         &0.668\hspace{2.5mm}(0.642 -- 0.693)\\
3 &Virchow2        &0.665\hspace{2.5mm}(0.639 -- 0.690)\\
4 &UNI2-h          &0.663\hspace{2.5mm}(0.637 -- 0.688)\\
5 &H-optimus-0     &0.660\hspace{2.5mm}(0.633 -- 0.685)\\
6 &CONCH           &0.657\hspace{2.5mm}(0.631 -- 0.681)\\
6 &Prov-GigaPath   &0.657\hspace{2.5mm}(0.631 -- 0.682)\\
8 &CONCHv1.5       &0.651\hspace{2.5mm}(0.625 -- 0.676)\\
9 &UNI             &0.648\hspace{2.5mm}(0.622 -- 0.674)\\
10 &Virchow        &0.639\hspace{2.5mm}(0.613 -- 0.665)\\
11 &RetCCL         &0.636\hspace{2.5mm}(0.609 -- 0.662)\\
12 &CTransPath     &0.627\hspace{2.5mm}(0.601 -- 0.652)\\
13 &Resnet-IN      &0.600\hspace{2.5mm}(0.572 -- 0.627)\\
\bottomrule
\end{tabular}
        }
        \caption{\textbf{RFS -- Patient Subgroup (ER+ \& HER2-)}.}
        \label{table:main_results_rfs_subgroup}
        \end{subtable}
        \vskip 0.075in
        \begin{subtable}{0.49\linewidth}
        \centering
        \vskip -0.05in
        \resizebox{1.0\linewidth}{!}{
\begin{tabular}{cl@{\hspace{1.2cm}}c}
\toprule
Rank &Model Name &C-index ($\uparrow$)\\
\midrule
1 &H-optimus-1     &0.702\hspace{2.5mm}(0.666 -- 0.737)\\
2 &Virchow         &0.695\hspace{2.5mm}(0.660 -- 0.729)\\
2 &CONCHv1.5       &0.695\hspace{2.5mm}(0.660 -- 0.729)\\
4 &H0-mini         &0.692\hspace{2.5mm}(0.655 -- 0.728)\\
5 &H-optimus-0     &0.686\hspace{2.5mm}(0.651 -- 0.722)\\
6 &UNI2-h          &0.683\hspace{2.5mm}(0.647 -- 0.719)\\
7 &Virchow2        &0.681\hspace{2.5mm}(0.645 -- 0.718)\\
8 &CONCH           &0.675\hspace{2.5mm}(0.639 -- 0.710)\\
9 &Prov-GigaPath   &0.673\hspace{2.5mm}(0.636 -- 0.710)\\
10 &UNI            &0.671\hspace{2.5mm}(0.634 -- 0.708)\\
11 &CTransPath     &0.647\hspace{2.5mm}(0.609 -- 0.686)\\
12 &Resnet-IN      &0.628\hspace{2.5mm}(0.590 -- 0.666)\\
13 &RetCCL         &0.627\hspace{2.5mm}(0.589 -- 0.666)\\
\bottomrule
\end{tabular}
        }
        \caption{\textbf{PFS -- All Patients}.}
        \label{table:main_results_pfs_all}
        \end{subtable}
        \hfill
        \begin{subtable}{0.49\linewidth}
        \centering
        \vskip -0.05in
        \resizebox{1.0\linewidth}{!}{
\begin{tabular}{cl@{\hspace{1.2cm}}c}
\toprule
Rank &Model Name &C-index ($\uparrow$)\\
\midrule
1 &CONCHv1.5       &0.680\hspace{2.5mm}(0.636 -- 0.721)\\
2 &Virchow         &0.678\hspace{2.5mm}(0.634 -- 0.720)\\
3 &H-optimus-1     &0.670\hspace{2.5mm}(0.626 -- 0.715)\\
3 &H-optimus-0     &0.670\hspace{2.5mm}(0.624 -- 0.714)\\
5 &H0-mini         &0.666\hspace{2.5mm}(0.619 -- 0.711)\\
6 &UNI2-h          &0.657\hspace{2.5mm}(0.611 -- 0.702)\\
7 &CONCH           &0.656\hspace{2.5mm}(0.612 -- 0.700)\\
8 &Virchow2        &0.655\hspace{2.5mm}(0.608 -- 0.700)\\
9 &Prov-GigaPath   &0.641\hspace{2.5mm}(0.594 -- 0.687)\\
10 &UNI            &0.638\hspace{2.5mm}(0.592 -- 0.683)\\
11 &CTransPath     &0.622\hspace{2.5mm}(0.573 -- 0.670)\\
12 &Resnet-IN      &0.606\hspace{2.5mm}(0.557 -- 0.655)\\
13 &RetCCL         &0.604\hspace{2.5mm}(0.555 -- 0.653)\\
\bottomrule
\end{tabular}
        }
        \caption{\textbf{PFS -- Patient Subgroup (ER+ \& HER2-)}.}
        \label{table:main_results_pfs_subgroup}
        \end{subtable}
    \end{minipage}
\end{figure*}

\citet{campanella2025clinical} present a large-scale benchmark of publicly available PFMs across multiple datasets, but their evaluation is limited to binary detection and biomarker prediction tasks, and relies on internal cross-validation rather than systematic external validation across independent cohorts, limiting insight into real-world generalization performance. \citet{neidlinger2025benchmarking} perform a comprehensive multi-cohort evaluation across a wide range of weakly supervised tasks, including prognostic endpoints. However, these are formulated as binary classification problems rather than time-to-event modeling, thereby not fully capturing the complexity of survival prediction. Moreover, many of their evaluations are conducted in relatively small or low-sample settings. PathBench~\citep{ma2025pathbench} is a more comprehensive framework that extends benchmarking across multiple cancer types and a wide spectrum of tasks, including diagnosis, molecular prediction, and survival prediction. However, survival prediction is evaluated exclusively using cross-validation, without independent external validation, and with moderate cohort sizes for all considered cancer types (at most 451 patients for breast cancer, 260 for gastric cancer, and 608 for colorectal cancer, across different endpoints). While external datasets are used for several diagnostic and classification tasks in PathBench, they are not applied to survival prediction.

Consequently, despite these important advances, there remains a lack of large-scale, multi-cohort evaluations specifically targeting survival prediction from WSIs under rigorous external validation. Existing benchmarks provide only a partial view of model utility, since clinically relevant applications require robust generalization across different patient populations, institutions, and scanning conditions~\citep{bilal2025foundation, lin2025institutionbias, Jahanifar2025DomainGen, thiringer2026scanner}. Survival prediction from histopathology images represents a particularly challenging and clinically important task: prognostic signals are often subtle, spatially heterogeneous, and confounded by clinical and biological variability. At the same time, accurate survival prediction has clear potential for improving risk stratification, treatment planning, and clinical decision support. Systematic evaluation of PFMs in this setting is therefore essential for understanding their practical value.

\begin{table}[t]
	\caption{\textbf{Main model ranking}, aggregated as the mean model rank ($\downarrow$) across the four evaluation settings based on Table~\ref{table:main_results}.}\vspace{-2.25mm}
    \label{table:main_results_rank}
    \centering
	\resizebox{0.4775\textwidth}{!}{%
		% \begin{tabular}{cl@{\hspace{1.2cm}}cc}
% \toprule
% Rank &Model Name &Model Rankings &Average Model Ranking ($\downarrow$)\\
% \midrule
% X &Virchow2       &3,3,7,8      &5.25\\
% X &Virchow        &10,10,2,2    &6\\
% X &CONCHv1.5      &8,8,2,1      &4.75\\
% X &CONCH          &6,6,8,7      &6.75\\
% X &Prov-GigaPath  &6,6,9,9      &7.5\\
% X &H0-mini        &2,2,4,5      &3.25\\
% X &H-optimus-1    &1,1,1,3      &1.5\\
% X &H-optimus-0    &5,5,5,3      &4.5\\
% X &UNI2-h         &4,4,6,6      &5\\
% X &UNI            &9,9,10,10    &9.5\\
% X &RetCCL         &11,11,13,13  &12\\
% X &CTransPath     &12,12,11,11  &11.5\\
% X &Resnet-IN      &13,13,12,12  &12.5\\
% \bottomrule
% \end{tabular}
%%%%%%%%%%%%%%%%%%%%%%%%%%%%%%%%%%%%
\begin{tabular}{cl@{\hspace{0.8cm}}cl}
\toprule
Rank &Model Name &Model Ranks &Mean Model Rank ($\downarrow$)\\
\midrule
1 &H-optimus-1     &1,1,1,3      &\hspace{13.0mm}1.5\\
2 &H0-mini         &2,2,4,5      &\hspace{13.0mm}3.25\\
3 &H-optimus-0     &5,5,5,3      &\hspace{13.0mm}4.5\\
4 &CONCHv1.5       &8,8,2,1      &\hspace{13.0mm}4.75\\
5 &UNI2-h          &4,4,6,6      &\hspace{13.0mm}5\\
6 &Virchow2        &3,3,7,8      &\hspace{13.0mm}5.25\\
7 &Virchow         &10,10,2,2    &\hspace{13.0mm}6\\
8 &CONCH           &6,6,8,7      &\hspace{13.0mm}6.75\\
9 &Prov-GigaPath   &6,6,9,9      &\hspace{13.0mm}7.5\\
10 &UNI            &9,9,10,10    &\hspace{13.0mm}9.5\\
11 &CTransPath     &12,12,11,11  &\hspace{13.0mm}11.5\\
12 &RetCCL         &11,11,13,13  &\hspace{13.0mm}12\\
13 &Resnet-IN      &13,13,12,12  &\hspace{13.0mm}12.5\\
\bottomrule
\end{tabular}
	}
\end{table}

To address this gap, we benchmark a diverse set of widely used and recently proposed PFMs (Table~\ref{table:fms}) for breast cancer survival prediction from WSIs. Our evaluation is conducted on three independent clinical cohorts comprising 5{,}434 patients with long-term follow-up. Models are trained on one cohort and evaluated on two independent external cohorts, enabling a rigorous assessment of cross-dataset generalization. To the best of our knowledge, this constitutes the largest multi-cohort benchmark for histopathology-based survival prediction with independent external validation. The benchmark spans multiple generations of pathology representation learning, including early pathology-specific models trained using self-supervised learning, and state-of-the-art PFMs trained on more than one million WSIs.

\textit{Our study makes four main contributions:}
(1) We present a large-scale, multi-cohort benchmark for breast cancer survival prediction, leveraging over 5,400 patients with independent external validation to enable robust assessment of model generalization.
(2) We provide a systematic head-to-head evaluation of a natural-image baseline, early pathology-specific models, state-of-the-art PFMs, and vision-language PFMs.
(3) We characterize performance trends across model families, showing consistent generational improvements but only modest absolute gains, suggesting diminishing returns from further scaling of pretraining data or model size alone.
(4) We demonstrate that compact distilled models can match or even exceed the prognostic risk-stratification performance of significantly larger teacher models, highlighting knowledge distillation as a promising approach for efficient deployment of PFMs.

\begin{table}[t]
	\caption{\textbf{Overview of all thirteen evaluated models}, including architecture, model size (number of parameters), feature dimension, and pretraining data scale. The models span a natural-image baseline, two early pathology-specific models, seven state-of-the-art PFMs, a compact distilled PFM, and two vision-language PFMs.}\vspace{-2.25mm}	
    \label{table:fms}
    \centering
	\resizebox{1.0\linewidth}{!}{%
        \begin{tabular}{lcccc}
\toprule
Model Name &Architecture &Size &\makecell{Feature\\ Dimension} &Pretraining Data\\
\midrule
Resnet-IN~\citep{he2016deep}          &\small{ResNet-50}     &25M      &1024 &1.3M natural images\\
\midrule
CTransPath~\citep{Wang2023ctranspath} &\small{CNN + Swin-T}  &22M      &768  &30K WSIs\\
RetCCL~\citep{wang2023retccl}         &\small{ResNet-50}     &25M      &2048 &32K WSIs\\
\midrule
UNI~\citep{chen2024uni}               &ViT-L         &307M     &1024 &100K WSIs\\
UNI2-h~\citep{UNI2h2024}              &ViT-H         &682M     &1536 &350K WSIs\\
H-optimus-0~\citep{hoptimus0}         &ViT-G         &1.1B &1536 &500K WSIs\\
H-optimus-1~\citep{hoptimus1}         &ViT-G         &1.1B &1536 &1M WSIs\\
Prov-GigaPath~\citep{gigapath2024}    &ViT-G         &1.1B &1536 &170K WSIs\\
Virchow~\citep{virchow2024}           &ViT-H         &632M     &2560 &1.5M WSIs\\
Virchow2~\citep{virchow22024}         &ViT-H         &632M     &2560 &3.1M WSIs\\
\midrule
H0-mini~\citep{filiot2025h0mini}      &ViT-B         &86M      &768  &500K + 6K WSIs\\
\midrule
CONCH~\citep{conch2024}               &ViT-B         &86M      &512  &\small{21K WSIs + 1.1M image-text pairs}\\
CONCHv1.5~\citep{ding2025multimodal}  &ViT-L         &307M     &768  &N/A\\
\bottomrule
\end{tabular}

% (number of params based on "Distilling foundation models for robust and efficient models in digital pathology", both the arxiv version and the MICCAI version)
	}
\end{table}

% ma2025pathbench: 
% Survival prediction for breast cancer, gastric cancer, and colorectal cancer.

% Breast cancer:
% Overall survival: 392 censored patients (1,089 slides) and 59 deceased patients (181 slides).
% - 451 patients in total
% Disease Free Survival: 80 disease-free patients (1,066 slides) and 71 recurred patients (204 slides).
% - 151 patients in total

% Gastric cancer:
% Overall survival: 260 slides from 260 cases: 172 censored patients and 88 uncensored patients.
% - 260 patients
% Disease Free Survival: 260 slides from 260 cases: 157 disease-free patients and 103 recurred patients.
% - 260 patients

% Colorectal cancer:
% Overall survival: 608 patients (2,779 slides). It contains 440 living patients (2,081 slides) and 168 deceased patients (698 slides).
% - 608 patients
% Disease Free Survival: 2,779 slides from 608 cases. It contains 389 disease-free patients (1,874 slides) and 219 recurred/progressed patients (904 slides).
% - 608 patients
% Disease Specific Survival: 294 patients (301 slides). It contains 252 living, or dead but tumor-free patients (259 slides) and 42 dead patients with tumor (42 slides).
% - 294 patients

% And just cross-validation resutls for all survival prediction.

% In contrast, we do:
% Survival models are trained on a dataset of 2,315 patients (SöS-BC-4) and evaluated on two independent external datasets (KS-Solna and SCAN-B-Lund) comprising 3,119 patients in total.

\section*{Results}

% Methods/evaluation summary:
We evaluate thirteen representative models spanning a natural-image baseline (Resnet-IN), two early pathology-specific models (CTransPath~\citep{Wang2023ctranspath}, RetCCL~\citep{wang2023retccl}), seven state-of-the-art PFMs (Prov-GigaPath~\citep{gigapath2024}, UNI~\citep{chen2024uni}, UNI2-h~\citep{UNI2h2024}, Virchow~\citep{virchow2024}, Virchow2~\citep{virchow22024}, H-optimus-0~\citep{hoptimus0}, H-optimus-1~\citep{hoptimus1}), a compact distilled PFM (H0-mini~\citep{filiot2025h0mini}, distilled from H-optimus-0), and two vision-language PFMs (CONCH~\citep{conch2024}, CONCHv1.5~\citep{ding2025multimodal}).

Models are evaluated using a unified pipeline for WSI-based survival prediction (see \textit{Methods} for details). Each model is used as a frozen feature extractor to compute patch-level feature vectors for the given WSI, which are aggregated into a predicted patient-level risk score using the PANTHER~\citep{song2024morphological} survival modelling framework. Survival models are trained on a dataset of 2{,}315 patients (SöS-BC-4) and evaluated on two independent external datasets (KS-Solna and SCAN-B-Lund) comprising 3{,}119 patients in total. None of the evaluated PFMs were pretrained on any of these datasets, ensuring a strict separation between pretraining data and the benchmark.

Performance is measured using the concordance index (C-index) and Kaplan-Meier survival analysis. Models are evaluated under four complementary settings: recurrence-free survival (RFS) and progression-free survival (PFS), each assessed both for the full cohort (`All Patients') and for the clinically relevant `ER+ \& HER2-' patient subgroup. Model rankings are computed for each of the four settings and aggregated to obtain a final overall ranking. The combined evaluation set of 3{,}119 patients contains 615 RFS events and 233 PFS events, with 2{,}524 patients (80.9\% of the full set), 475 RFS events (77.2\%) and 157 PFS events (67.4\%) in the `ER+ \& HER2-' patient subgroup.

\subsubsection*{Main Model Comparison}
Figure~\ref{fig:main_results} shows the C-index performance of all thirteen models across the four evaluation settings, with corresponding numerical results and model rankings in Table~\ref{table:main_results}. The aggregated overall ranking is summarized in Table~\ref{table:main_results_rank}.

H-optimus-1 achieves the highest C-index in three out of four settings (\textit{RFS -- All Patients}, \textit{RFS -- Patient Subgroup}, \textit{PFS -- All Patients}) and attains the best overall model ranking. In absolute terms, this corresponds to a C-index of 0.678 for \textit{RFS -- All Patients} and 0.702 for \textit{PFS -- All Patients}. Notably, the compact distilled model H0-mini achieves the second-best overall ranking, and slightly outperforms its teacher model H-optimus-0 in three out of four settings. CONCHv1.5, UNI2-h and Virchow2 also demonstrate strong overall performance.

In contrast, the natural-image baseline Resnet-IN achieves the lowest overall performance and ranks in the bottom two across all four settings. The two early pathology-specific models CTransPath and RetCCL also perform poorly, consistently occupying the bottom three ranks together with Resnet-IN. Among recent PFMs, UNI achieves the lowest overall performance, only beating Resnet-IN, CTransPath and RetCCL.

Across model families, a consistent pattern is observed in which each second-generation PFM (H-optimus-1, CONCHv1.5, UNI2-h, Virchow2) slightly outperforms its corresponding first-generation counterpart (H-optimus-0, CONCH, UNI, Virchow) in the aggregated ranking.

\begin{figure*}[t]
\centering
    \begin{subfigure}[t]{1.0\textwidth}
        \centering%
        \begin{subfigure}[t]{0.33\textwidth}
            \centering%
            \includegraphics[clip, trim=0.5cm 1.25cm 0.25cm 0.0cm, width=1.0\linewidth]{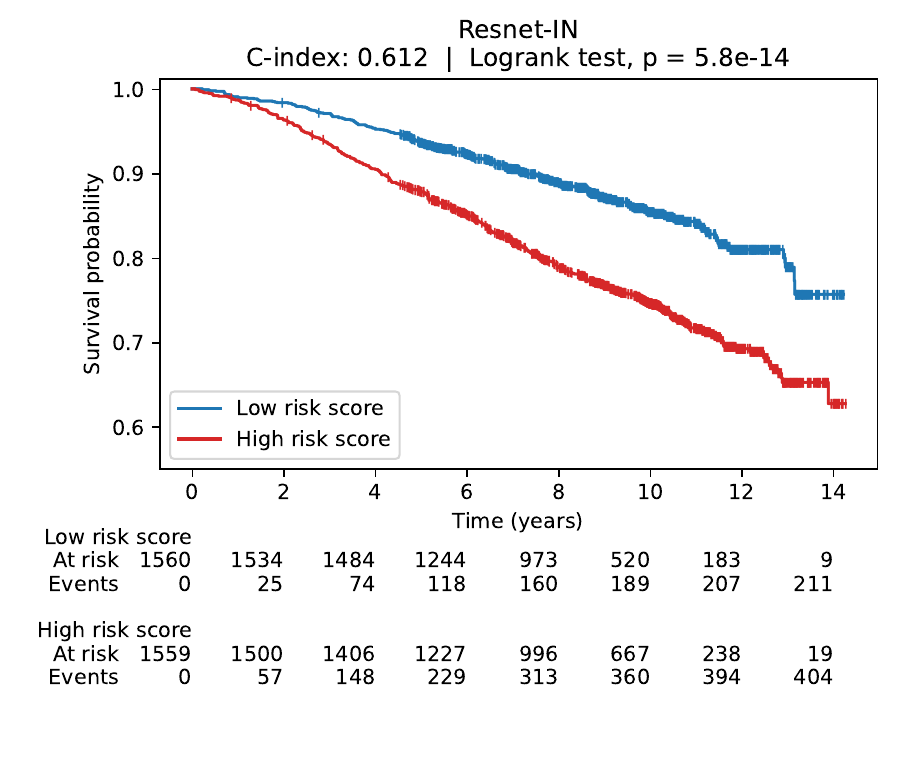}
        \end{subfigure}
        \begin{subfigure}[t]{0.33\textwidth}
            \centering%
            \includegraphics[clip, trim=0.5cm 1.25cm 0.25cm 0.0cm, width=1.0\linewidth]{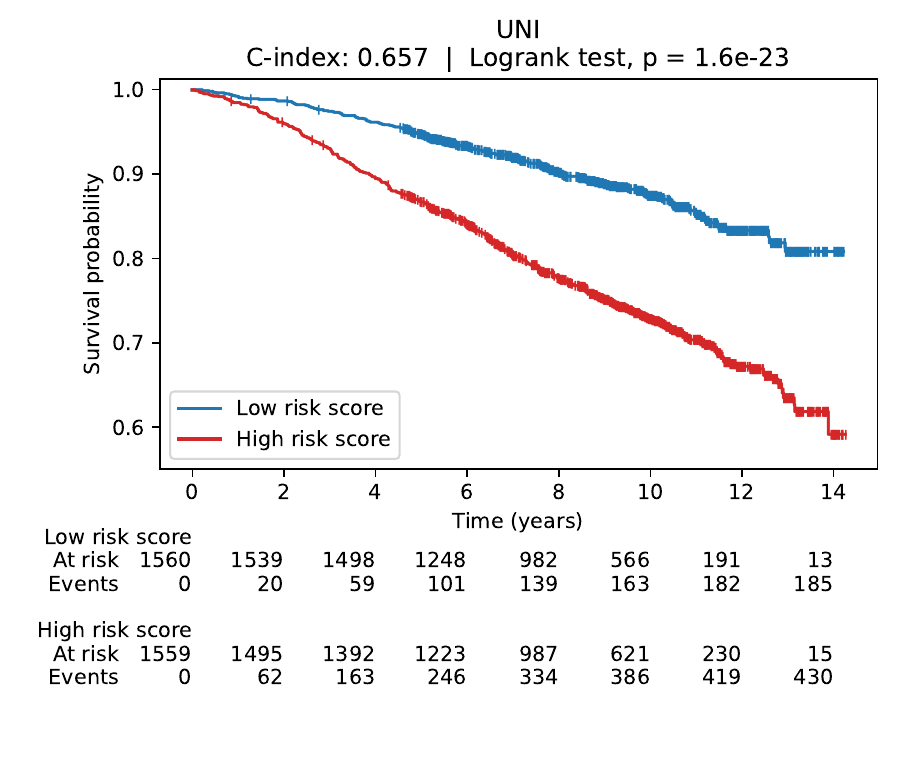}
        \end{subfigure}
        \begin{subfigure}[t]{0.33\textwidth}
            \centering%
            \includegraphics[clip, trim=0.5cm 1.25cm 0.25cm 0.0cm, width=1.0\linewidth]{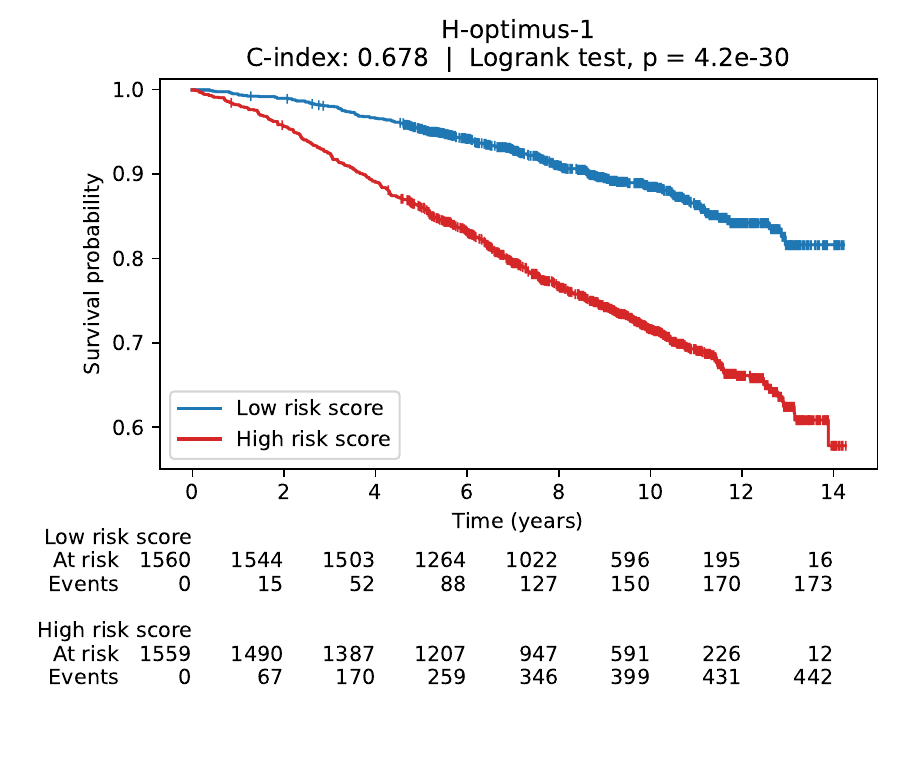}
        \end{subfigure}\vspace{0.0mm}
        \caption{\textbf{RFS -- All Patients}.}\vspace{0.0mm}
        \label{fig:km_plots_rfs_2groups_counts_all-patients}
    \end{subfigure}
    \begin{subfigure}[t]{1.0\textwidth}
        \centering%
        \begin{subfigure}[t]{0.33\textwidth}
            \centering%
            \includegraphics[clip, trim=0.5cm 1.25cm 0.25cm 0.0cm, width=1.0\linewidth]{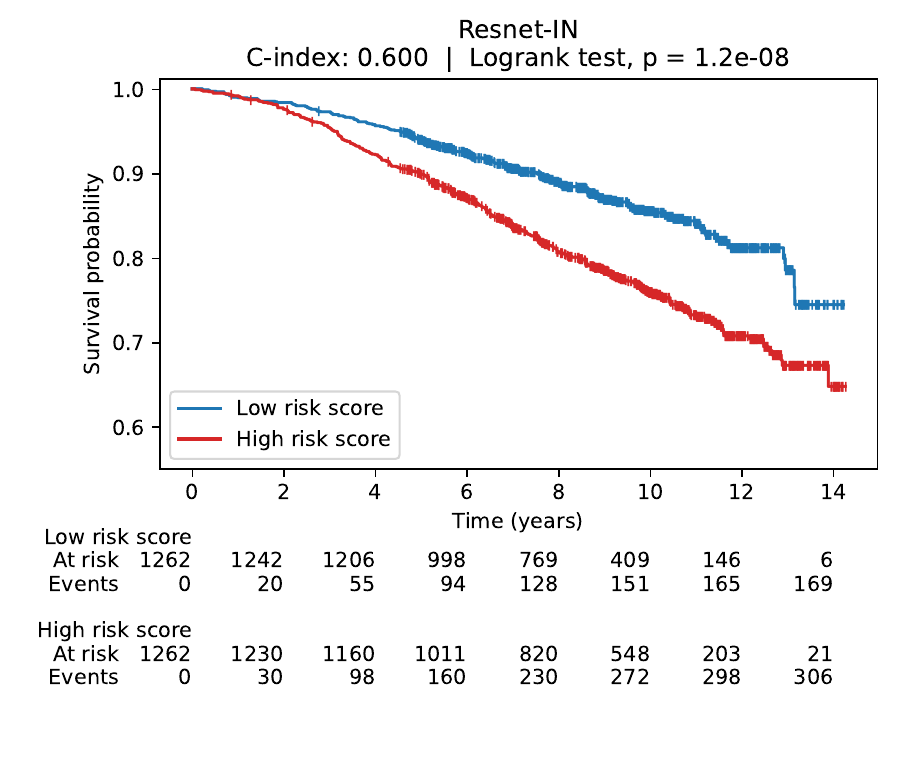}
        \end{subfigure}
        \begin{subfigure}[t]{0.33\textwidth}
            \centering%
            \includegraphics[clip, trim=0.5cm 1.25cm 0.25cm 0.0cm, width=1.0\linewidth]{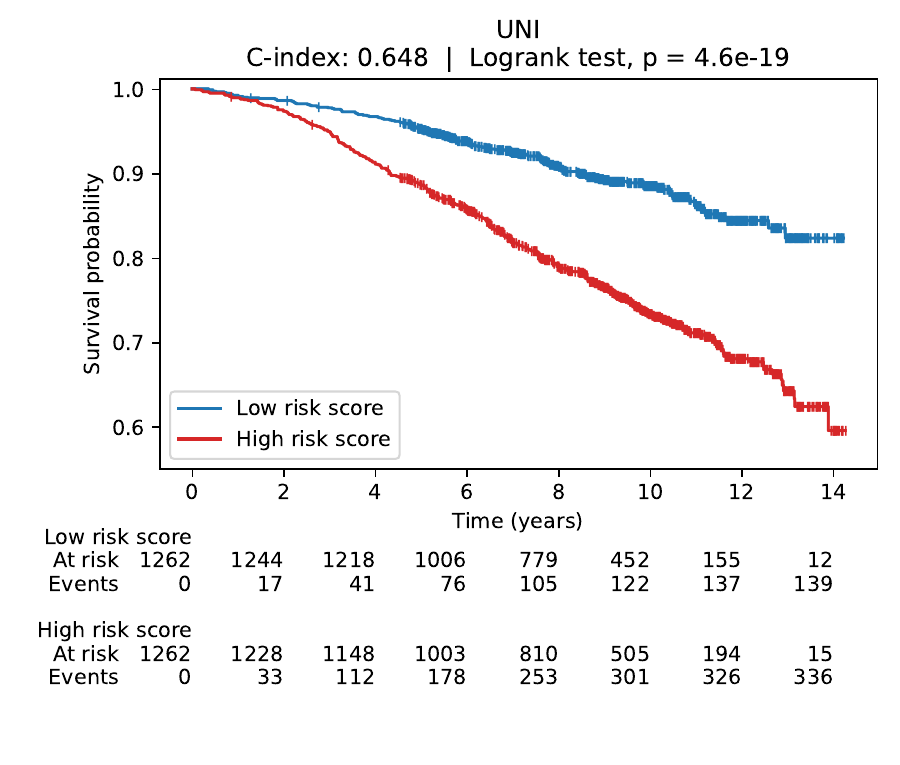}
        \end{subfigure}
        \begin{subfigure}[t]{0.33\textwidth}
            \centering%
            \includegraphics[clip, trim=0.5cm 1.25cm 0.25cm 0.0cm, width=1.0\linewidth]{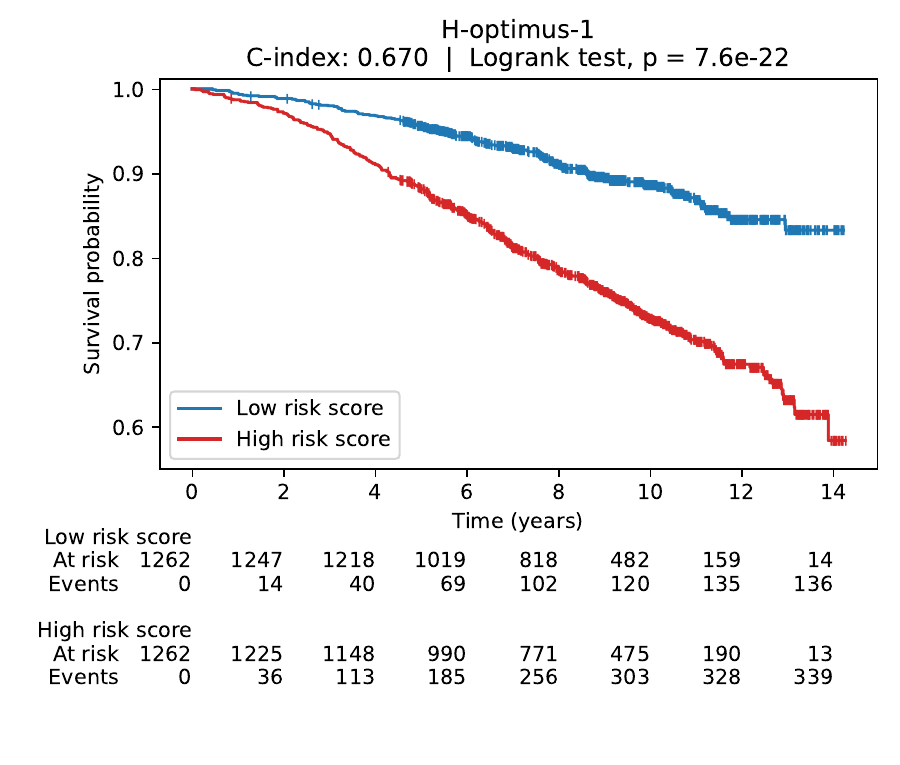}
        \end{subfigure}\vspace{0.0mm}
        \caption{\textbf{RFS -- Patient Subgroup (ER+ \& HER2-)}.}\vspace{0.0mm}
        \label{fig:km_plots_rfs_2groups_counts_subgroup}
    \end{subfigure}
    \begin{subfigure}[t]{1.0\textwidth}
        \centering%
        \begin{subfigure}[t]{0.33\textwidth}
            \centering%
            \includegraphics[clip, trim=0.5cm 1.25cm 0.25cm 0.0cm, width=1.0\linewidth]{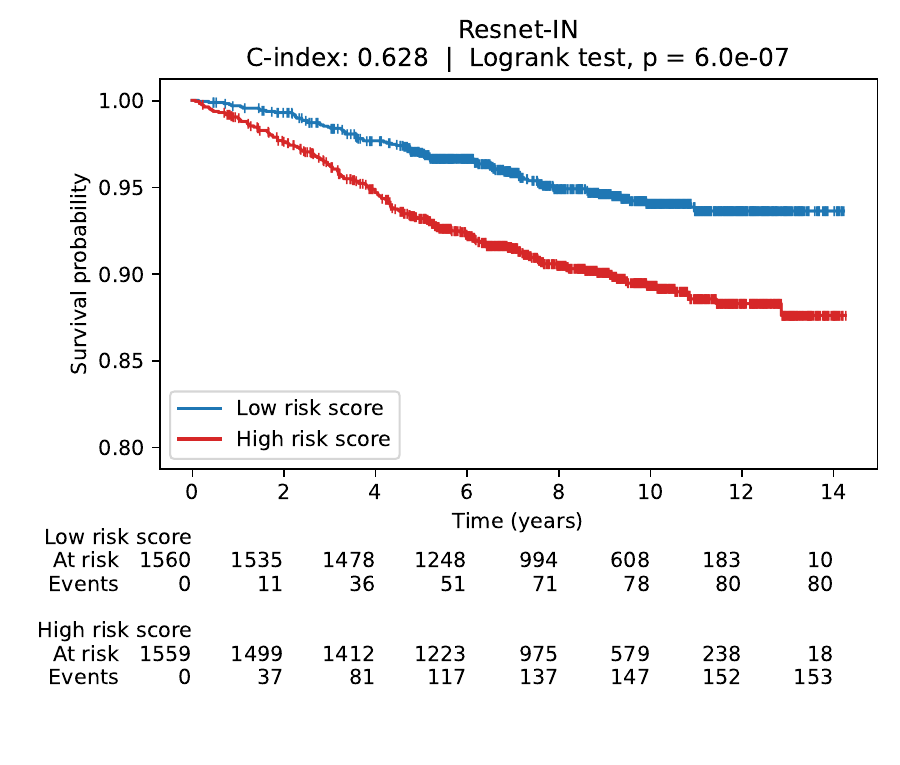}
        \end{subfigure}
        \begin{subfigure}[t]{0.33\textwidth}
            \centering%
            \includegraphics[clip, trim=0.5cm 1.25cm 0.25cm 0.0cm, width=1.0\linewidth]{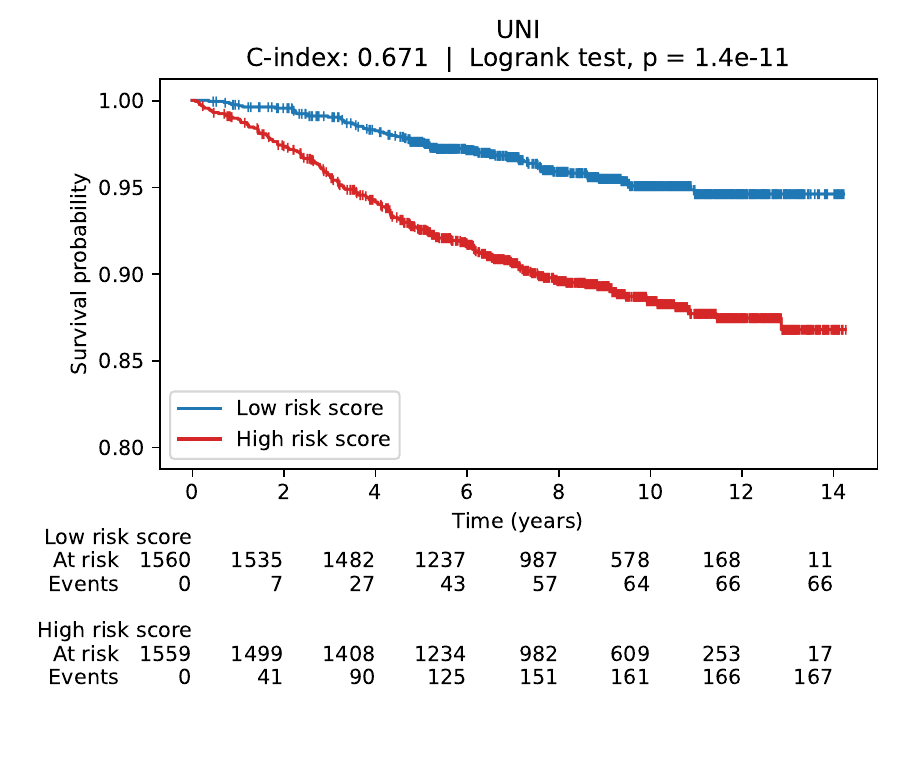}
        \end{subfigure}
        \begin{subfigure}[t]{0.33\textwidth}
            \centering%
            \includegraphics[clip, trim=0.5cm 1.25cm 0.25cm 0.0cm, width=1.0\linewidth]{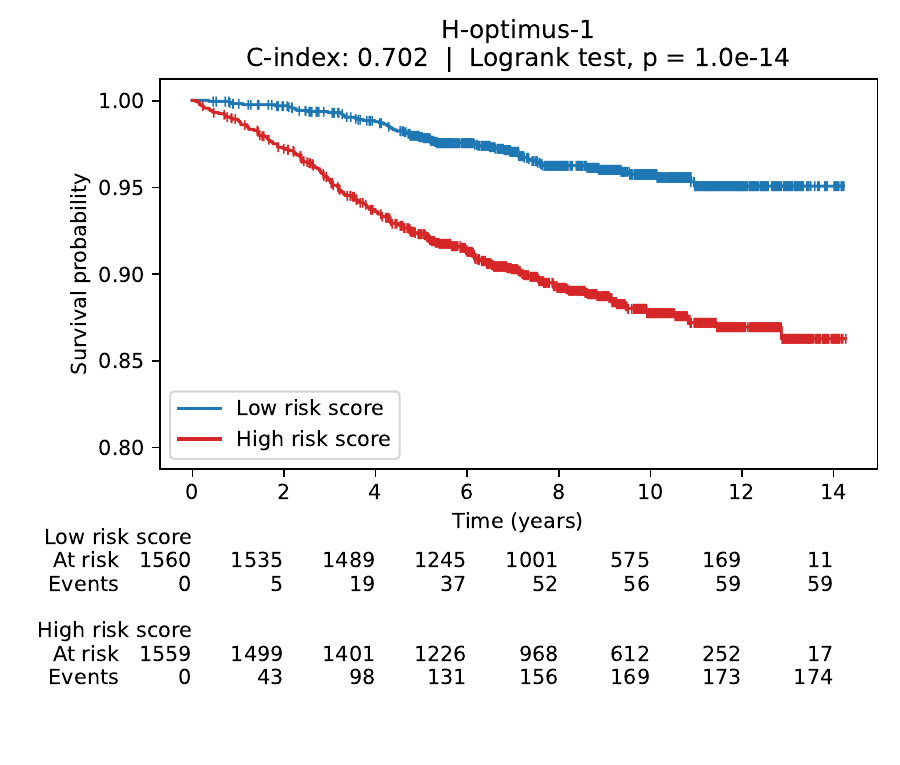}
        \end{subfigure}\vspace{0.0mm}
        \caption{\textbf{PFS -- All Patients}.}\vspace{0.0mm}
        \label{fig:km_plots_pfs_2groups_counts_all-patients}
    \end{subfigure}
    \begin{subfigure}[t]{1.0\textwidth}
        \centering%
        \begin{subfigure}[t]{0.33\textwidth}
            \centering%
            \includegraphics[clip, trim=0.5cm 1.25cm 0.25cm 0.0cm, width=1.0\linewidth]{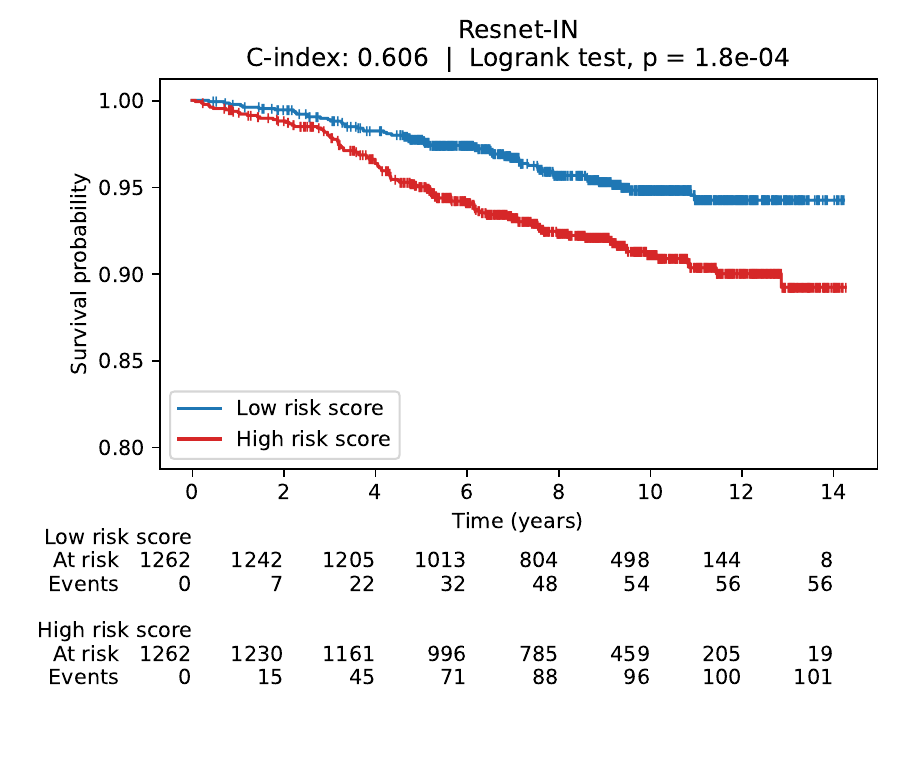}
        \end{subfigure}
        \begin{subfigure}[t]{0.33\textwidth}
            \centering%
            \includegraphics[clip, trim=0.5cm 1.25cm 0.25cm 0.0cm, width=1.0\linewidth]{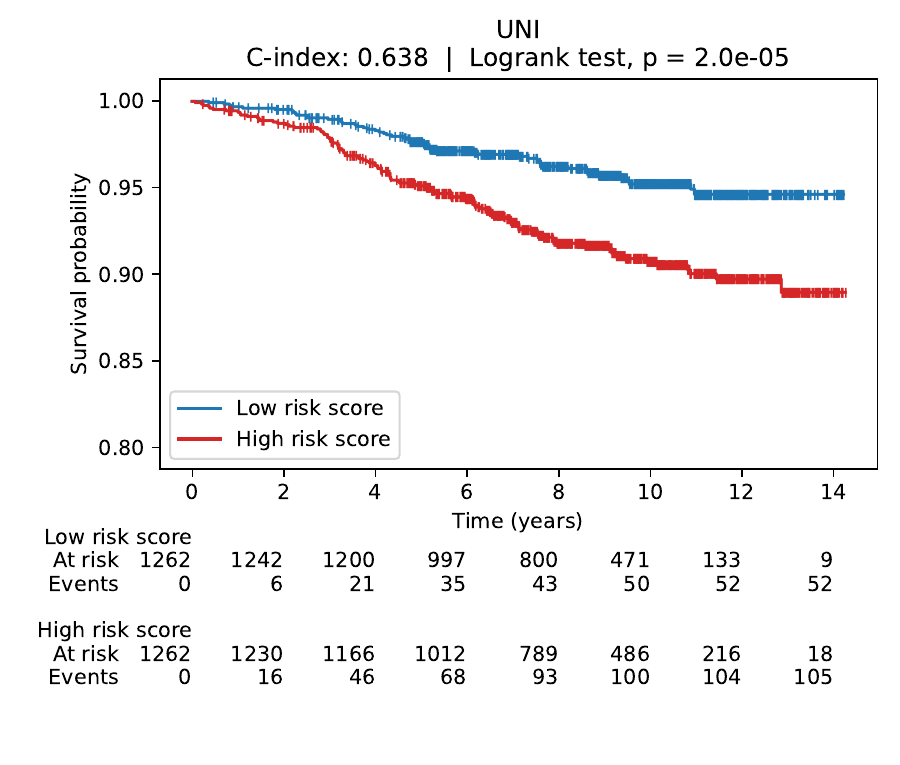}
        \end{subfigure}
        \begin{subfigure}[t]{0.33\textwidth}
            \centering%
            \includegraphics[clip, trim=0.5cm 1.25cm 0.25cm 0.0cm, width=1.0\linewidth]{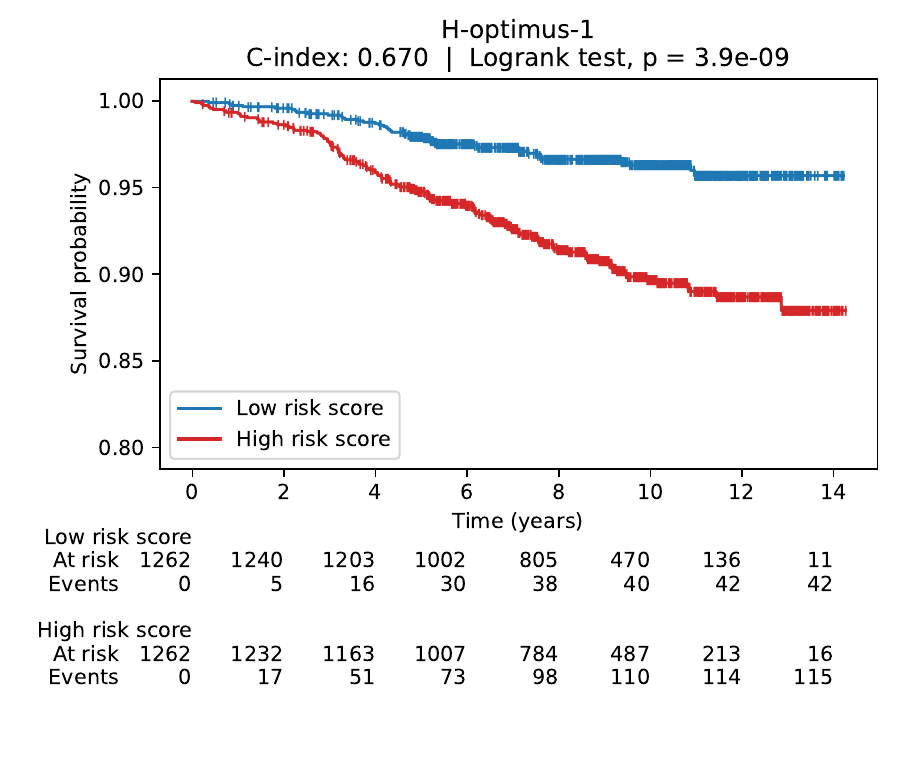}
        \end{subfigure}\vspace{0.0mm}
        \caption{\textbf{PFS -- Patient Subgroup (ER+ \& HER2-)}.}
        \label{fig:km_plots_pfs_2groups_counts_subgroup}
    \end{subfigure}\vspace{-1.0mm}
  \caption{\textbf{Two-group Kaplan-Meier risk stratification for three representative models}. KM survival curves showing stratification into low- and high-risk groups for RFS and PFS, each assessed for the full cohort (`All Patients') and the `ER+ \& HER2-' patient subgroup. Results for \textit{Resnet-IN} (left column), \textit{UNI} (middle), and \textit{H-optimus-1} (right). Each plot includes the C-index, log-rank test p-value, and the number of patients at risk and events over time (0-14 years). Note the difference in range of the y-axis between RFS and PFS.}
  \label{fig:km_plots_2groups_counts}
\end{figure*}

\begin{figure*}[t]
\centering
    \begin{subfigure}[t]{1.0\textwidth}
        \centering%
        \begin{subfigure}[t]{0.33\textwidth}
            \centering%
            \includegraphics[clip, trim=0.5cm 0.25cm 1.25cm 0.5cm, width=1.0\linewidth]{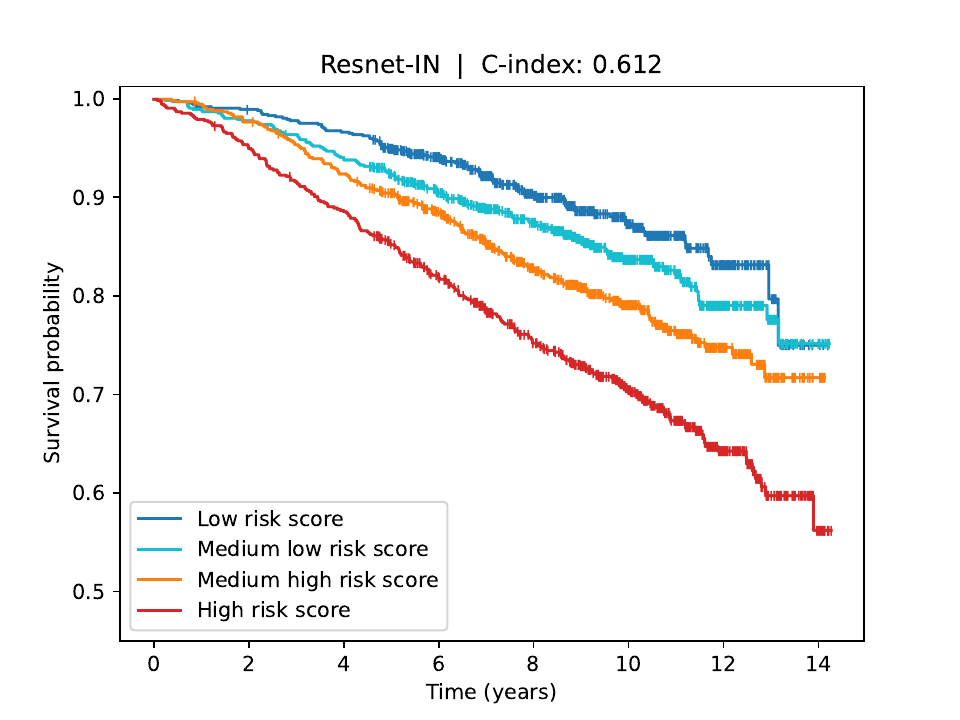}
        \end{subfigure}
        \begin{subfigure}[t]{0.33\textwidth}
            \centering%
            \includegraphics[clip, trim=0.5cm 0.25cm 1.25cm 0.5cm, width=1.0\linewidth]{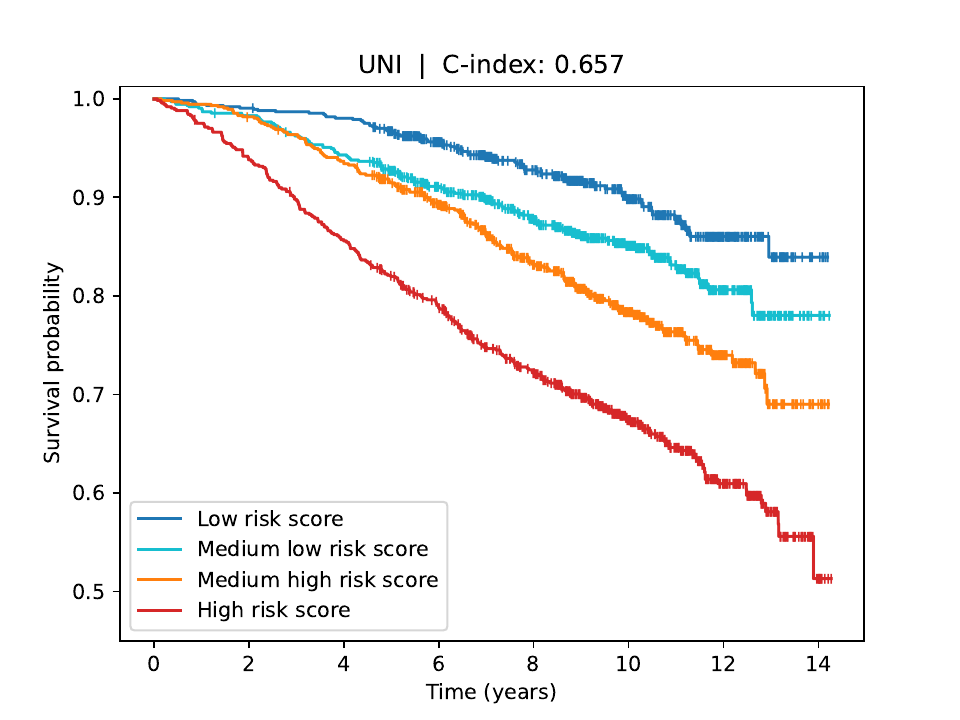}
        \end{subfigure}
        \begin{subfigure}[t]{0.33\textwidth}
            \centering%
            \includegraphics[clip, trim=0.5cm 0.25cm 1.25cm 0.5cm, width=1.0\linewidth]{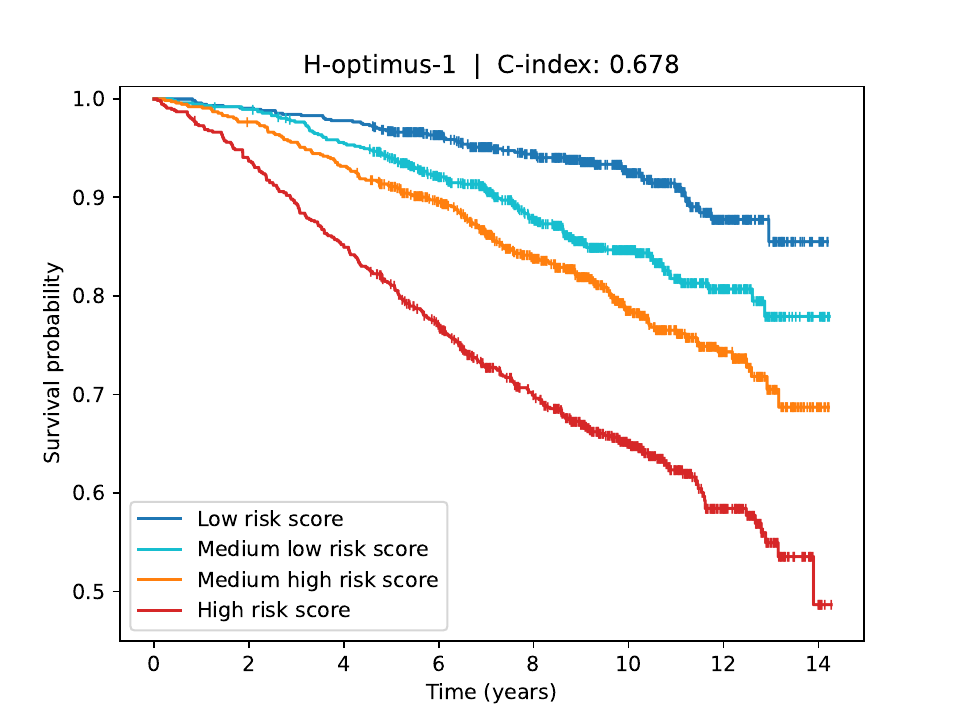}
        \end{subfigure}\vspace{0.0mm}
        \caption{\textbf{RFS -- All Patients}.}\vspace{0.0mm}
        \label{fig:km_plots_rfs_4groups_all-patients}
    \end{subfigure}
    \begin{subfigure}[t]{1.0\textwidth}
        \centering%
        \begin{subfigure}[t]{0.33\textwidth}
            \centering%
            \includegraphics[clip, trim=0.5cm 0.25cm 1.25cm 0.5cm, width=1.0\linewidth]{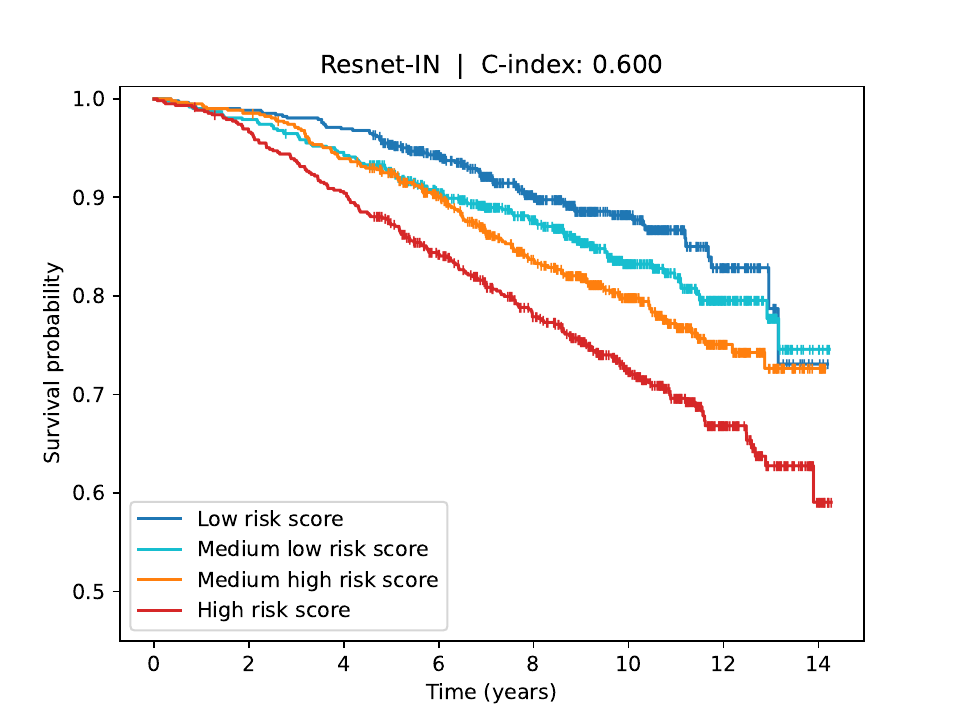}
        \end{subfigure}
        \begin{subfigure}[t]{0.33\textwidth}
            \centering%
            \includegraphics[clip, trim=0.5cm 0.25cm 1.25cm 0.5cm, width=1.0\linewidth]{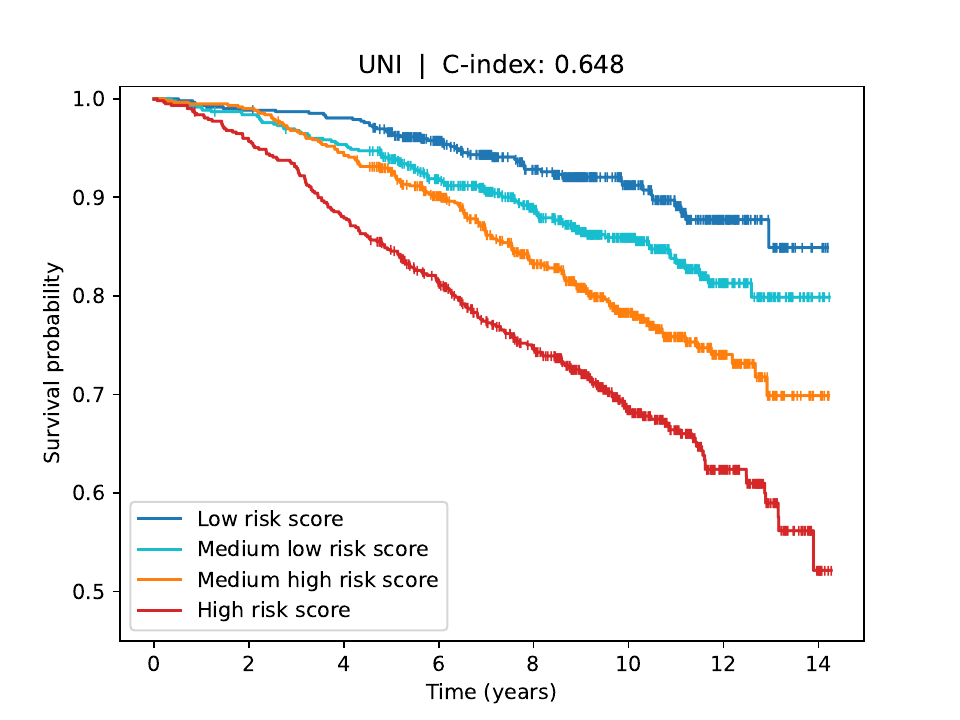}
        \end{subfigure}
        \begin{subfigure}[t]{0.33\textwidth}
            \centering%
            \includegraphics[clip, trim=0.5cm 0.25cm 1.25cm 0.5cm, width=1.0\linewidth]{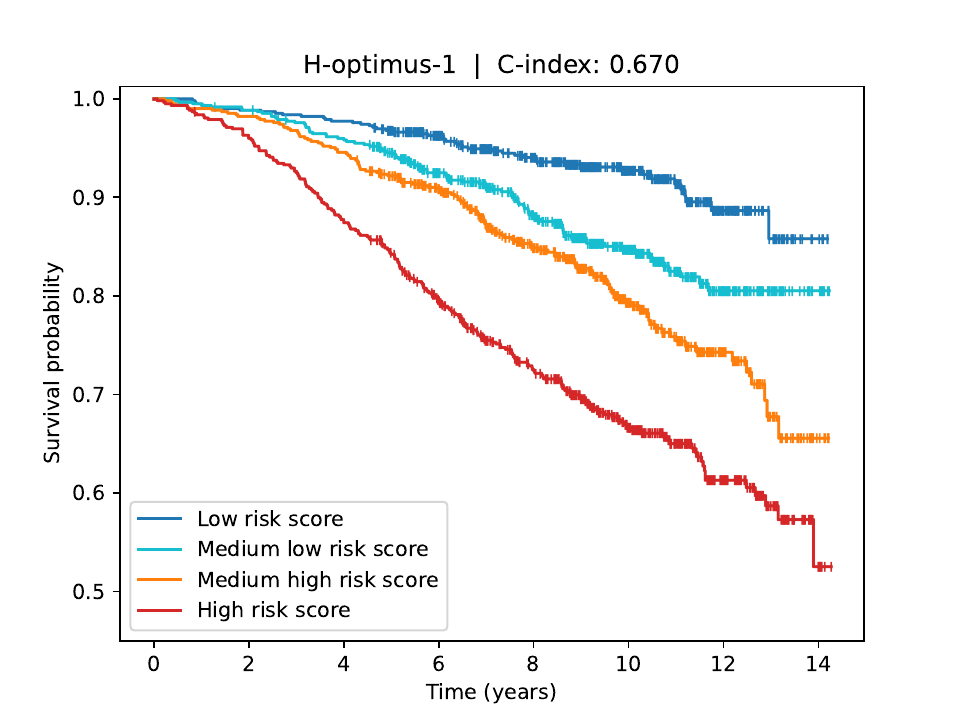}
        \end{subfigure}\vspace{0.0mm}
        \caption{\textbf{RFS -- Patient Subgroup (ER+ \& HER2-)}.}\vspace{0.0mm}
        \label{fig:km_plots_rfs_4groups_subgroup}
    \end{subfigure}
    \begin{subfigure}[t]{1.0\textwidth}
        \centering%
        \begin{subfigure}[t]{0.33\textwidth}
            \centering%
            \includegraphics[clip, trim=0.5cm 0.25cm 1.25cm 0.5cm, width=1.0\linewidth]{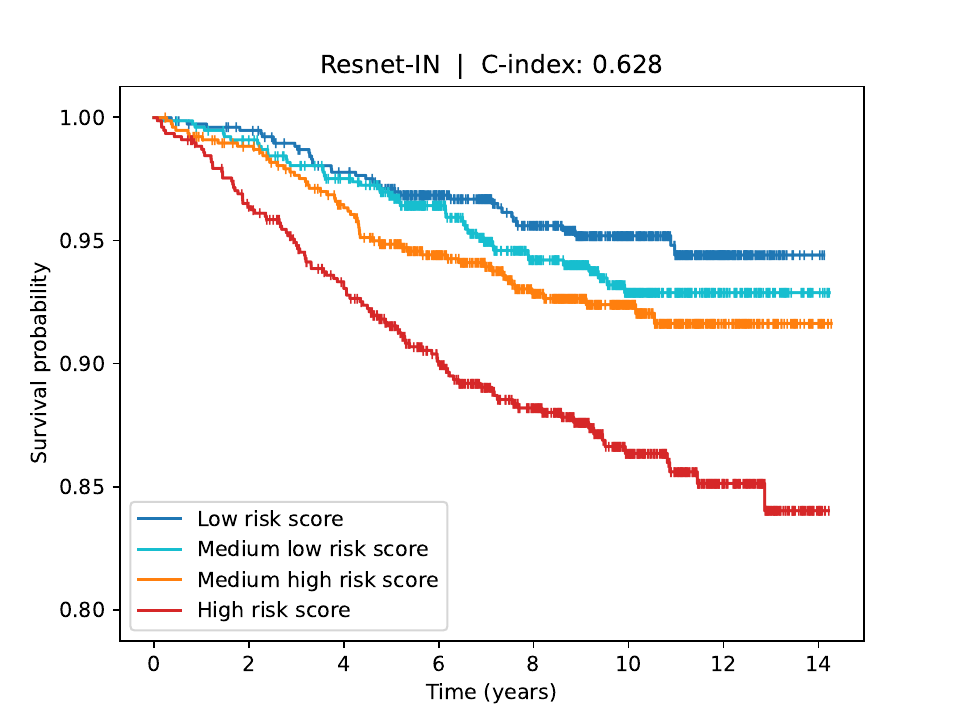}
        \end{subfigure}
        \begin{subfigure}[t]{0.33\textwidth}
            \centering%
            \includegraphics[clip, trim=0.5cm 0.25cm 1.25cm 0.5cm, width=1.0\linewidth]{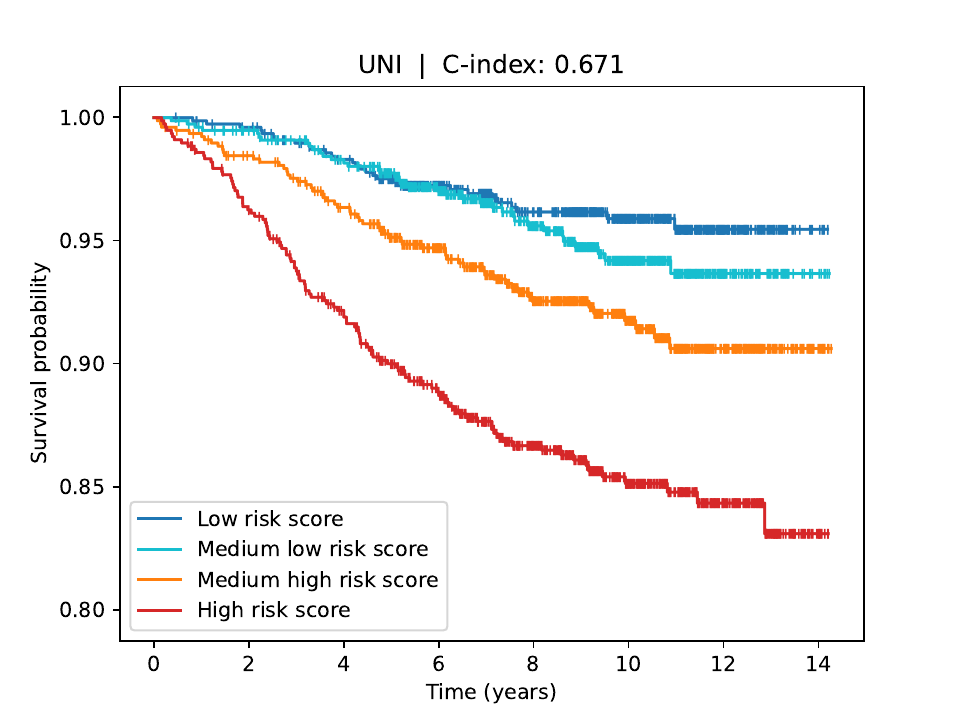}
        \end{subfigure}
        \begin{subfigure}[t]{0.33\textwidth}
            \centering%
            \includegraphics[clip, trim=0.5cm 0.25cm 1.25cm 0.5cm, width=1.0\linewidth]{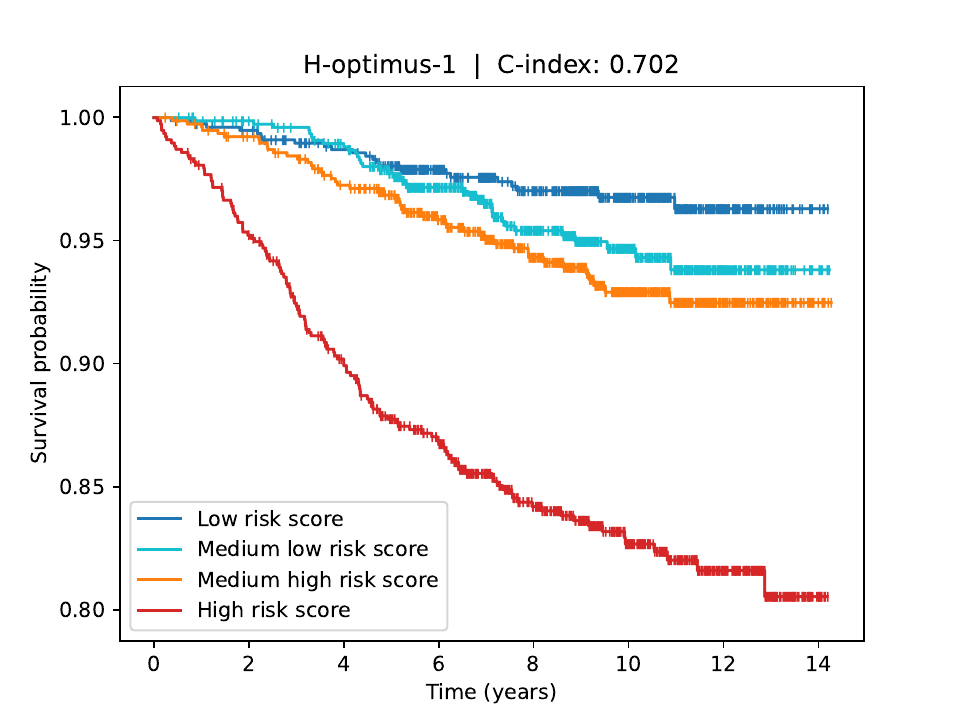}
        \end{subfigure}\vspace{0.0mm}
        \caption{\textbf{PFS -- All Patients}.}\vspace{0.0mm}
        \label{fig:km_plots_pfs_4groups_all-patients}
    \end{subfigure}
    \begin{subfigure}[t]{1.0\textwidth}
        \centering%
        \begin{subfigure}[t]{0.33\textwidth}
            \centering%
            \includegraphics[clip, trim=0.5cm 0.25cm 1.25cm 0.5cm, width=1.0\linewidth]{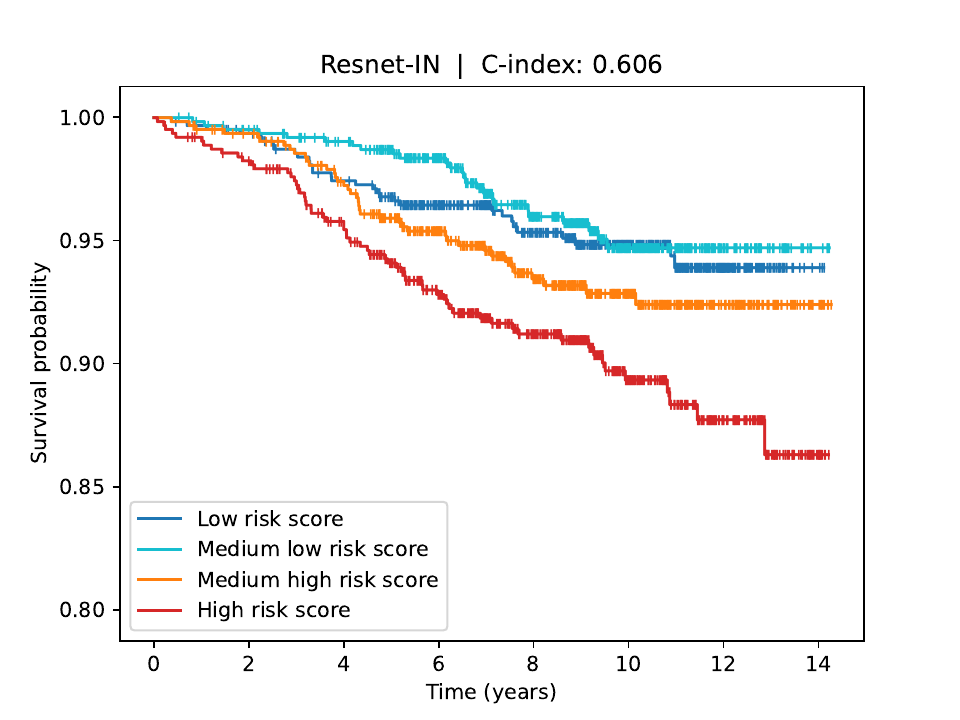}
        \end{subfigure}
        \begin{subfigure}[t]{0.33\textwidth}
            \centering%
            \includegraphics[clip, trim=0.5cm 0.25cm 1.25cm 0.5cm, width=1.0\linewidth]{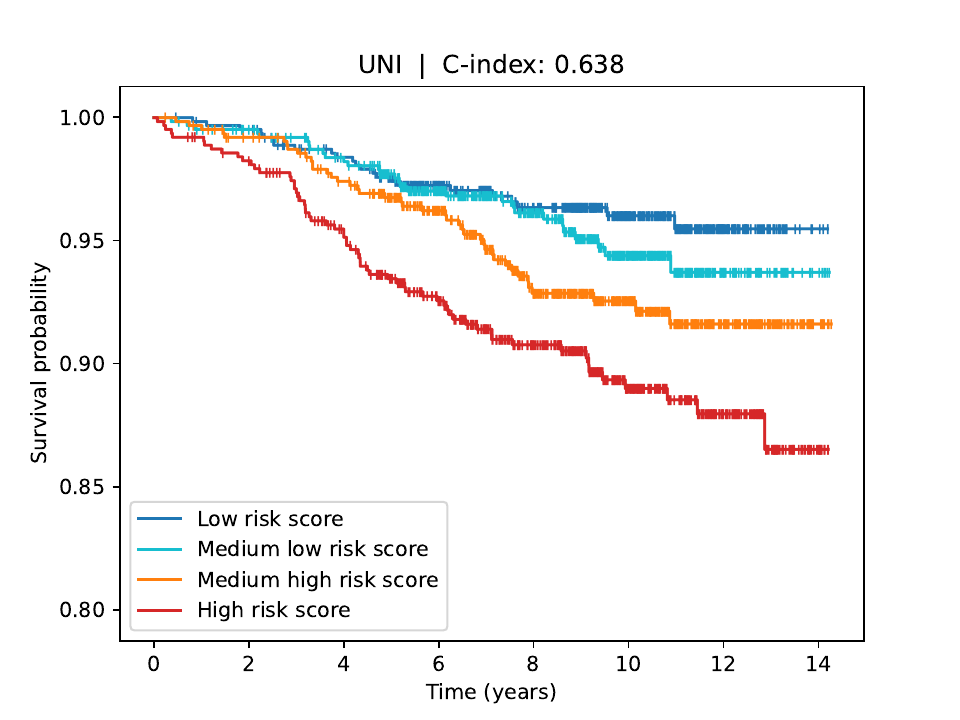}
        \end{subfigure}
        \begin{subfigure}[t]{0.33\textwidth}
            \centering%
            \includegraphics[clip, trim=0.5cm 0.25cm 1.25cm 0.5cm, width=1.0\linewidth]{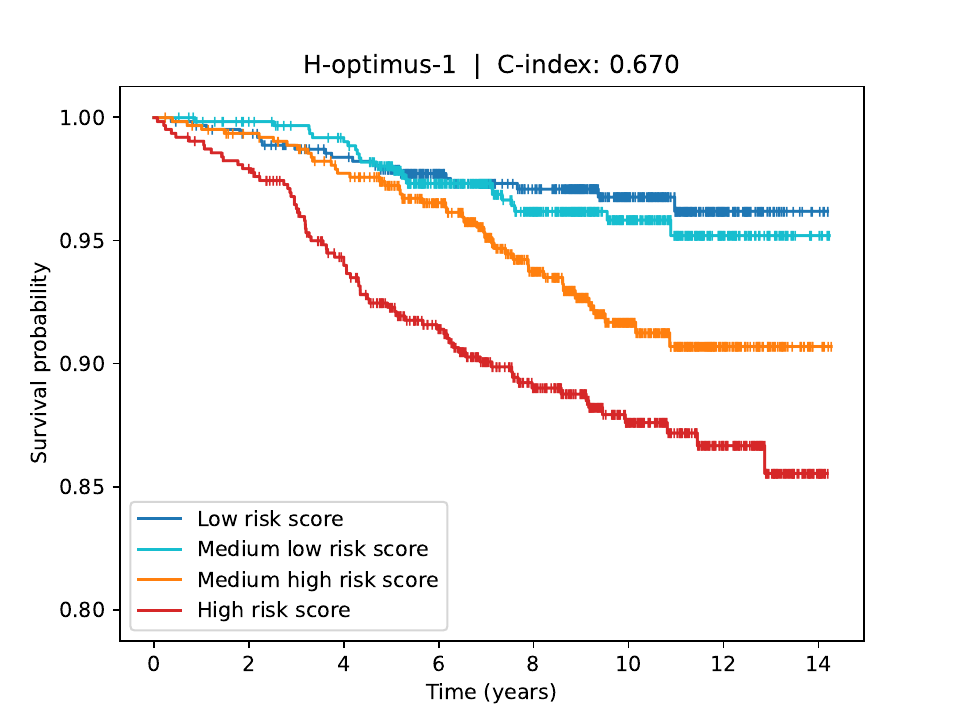}
        \end{subfigure}\vspace{0.0mm}
        \caption{\textbf{PFS -- Patient Subgroup (ER+ \& HER2-)}.}
        \label{fig:km_plots_pfs_4groups_subgroup}
    \end{subfigure}\vspace{-1.0mm}
  \caption{\textbf{Four-group Kaplan-Meier risk stratification for three representative models}. KM survival curves showing stratification into four risk groups for RFS and PFS, each assessed for the full cohort (`All Patients') and the `ER+ \& HER2-' patient subgroup. Results for \textit{Resnet-IN} (left column), \textit{UNI} (middle), and \textit{H-optimus-1} (right). Note the difference in range of the y-axis between RFS and PFS.}
  \label{fig:km_plots_4groups}
\end{figure*}

Despite these consistent trends, absolute performance differences between many models are relatively small. Confidence intervals for the C-index estimates show substantial overlap across most models. For example, in the \textit{RFS -- All Patients} setting, the top-ranked model exhibits non-overlapping 95\% confidence intervals only when compared to the two lowest-ranked models (CTransPath and Resnet-IN). A similar pattern is observed for \textit{RFS -- Patient Subgroup}, where non-overlapping intervals are only observed for the lowest-ranked model. For PFS, confidence intervals overlap across all models. Overall, these results indicate that differences in predictive performance between many models are modest despite consistent ranking patterns.

While model rankings are broadly consistent across the two survival endpoints (RFS and PFS), some variability is observed. In particular, CONCHv1.5 and Virchow rank quite low for RFS (8th and 10th in both settings, respectively) but substantially higher for PFS (top-two positions in both settings). Virchow2 also shows a shift in ranking, achieving 3rd place for RFS but 7th and 8th place for PFS. These deviations contrast with the otherwise stable ordering of models across endpoints, indicating that relative performance can vary depending on the specific survival task.

% Kaplan-Meier Risk Stratification
\subsubsection*{Survival Analysis Using the Kaplan-Meier Estimator}
Kaplan-Meier (KM) survival curves illustrate a survival function while accounting for right-censoring, and is used to demonstrate the ability of models to stratify patients into distinct risk groups. Figure~\ref{fig:km_plots_2groups_counts} shows stratification into two risk groups across all four evaluation settings, for Resnet-IN, UNI, and H-optimus-1. For each model and setting, the KM plots report log-rank test p-values, along with the number of patients at risk and the number of events over time (0-14 years). 

These three models are selected to represent the overall best-performing model (H-optimus-1), the overall worst-performing model (Resnet-IN), and the worst-performing model among PFMs (UNI), enabling direct visual comparison of risk stratification quality. KM plots for the remaining PFMs show broadly similar patterns across settings, with differences between models which are less pronounced than those observed between UNI and H-optimus-1.

Across all three models, consistent trends are observed in the degree of separation between risk groups across evaluation settings. The strongest separation is observed for \textit{RFS -- All Patients}, followed by \textit{RFS -- Patient Subgroup}, \textit{PFS -- All Patients}, and finally \textit{PFS -- Patient Subgroup}, as reflected by progressively larger log-rank p-values. Within each evaluation setting, differences between models are also evident. H-optimus-1 consistently shows the clearest separation between low- and high-risk groups, with the smallest log-rank p-values across all settings. UNI exhibits intermediate separation, while Resnet-IN shows the weakest separation, with larger p-values and more overlap between survival curves.

Figure~\ref{fig:km_plots_4groups} shows stratification into four risk groups across the same settings for the three representative models, providing a more detailed view of model behavior. Similar trends are observed across evaluation settings, with the clearest separation for \textit{RFS -- All Patients} and progressively weaker separation for \textit{RFS -- Patient Subgroup}, \textit{PFS -- All Patients}, and \textit{PFS -- Patient Subgroup}. In the latter setting, survival curves show increased overlap and less consistent ordering between adjacent risk groups, particularly at earlier time points.

Differences between models remain apparent in this setting. H-optimus-1 shows more distinct and consistently ordered risk groups over time, with clearer separation between curves. UNI exhibits moderate separation with some overlap between adjacent groups, while Resnet-IN shows the least consistent separation, with greater overlap and less stable ordering of risk groups. Notably, H-optimus-1 maintains clear separation across four risk groups in multiple evaluation settings, with well-ordered survival curves and sustained differences in survival probability over time. These patterns are consistent with the two-group analysis (Figure~\ref{fig:km_plots_2groups_counts}) and further highlight differences in the strength and stability of risk stratification across models.

Four-group KM plots including the number of patients at risk and the number of events over time are provided in Figure~\ref{fig:km_plots_4groups_counts_resnet-in}~-~\ref{fig:km_plots_4groups_counts_h-optimus-1} in the supplementary material, with corresponding plots for three risk groups shown in Figure~\ref{fig:km_plots_3groups_counts_resnet-in}~-~\ref{fig:km_plots_3groups_counts_h-optimus-1}.

Overall, KM analysis confirms that all models capture prognostic signal to some extent, including the lowest-performing model Resnet-IN. Clear differences in risk stratification are observed when comparing the three representative models, with the most pronounced contrast between H-optimus-1 and Resnet-IN, while differences between PFMs are more subtle. These results complement the C-index analysis by providing a qualitative view of model performance that is not fully captured by ranking metrics alone.

\begin{figure*}[t]
\centering
    \begin{subfigure}[t]{0.45\textwidth}
        \centering%
            \begin{tikzpicture}[scale=0.75, baseline]
                \begin{axis}[
                    % xmode=log,
                    % log ticks with fixed point,
                    xlabel={\% of SöS-BC-4 used for training},
                    ylabel={C-index ($\uparrow$)},
                    xmin=5, xmax=105,
                    ymin=0.5, ymax=0.7125,
                    xtick={10, 25, 50, 75, 100},
                    ytick={0.50, 0.55, 0.60, 0.65, 0.70},
                    yticklabels={0.50, 0.55, 0.60, 0.65, 0.70},
                    legend pos=south east,
                    ymajorgrids=true,
                    %xmajorgrids=true,
                    title={},
                    grid style=dashed,
                    y tick label style={
                        /pgf/number format/.cd,
                            fixed,
                            % fixed zerofill,
                            precision=2,
                        /tikz/.cd
                    },
                    every axis plot/.append style={thick},
                ]
                \addplot [Cerulean, mark=triangle*, line width=0.6125mm]
                 plot [error bars/.cd, y dir = both, y explicit]
                 table[row sep=crcr, x index=0, y index=1, y error index=2,]{
                10 0.5782 0.0266\\
                25 0.6177 0.0085\\
                50 0.6553 0.0033\\
                75 0.6683 0.0037\\
                100 0.6743 0.0009\\
                };
                \addlegendentry{H-optimus-1}
            
                \addplot [flamingopink, mark=square*, line width=0.6125mm]
                 plot [error bars/.cd, y dir = both, y explicit]
                 table[row sep=crcr, x index=0, y index=1, y error index=2,]{
                10 0.5506 0.0250\\
                25 0.5932 0.0195\\
                50 0.6337 0.0164\\
                75 0.6515 0.0099\\
                100 0.6524 0.0062\\
                };
                \addlegendentry{UNI}
    
                \addplot [gray, mark=*, line width=0.6125mm]
                 plot [error bars/.cd, y dir = both, y explicit]
                 table[row sep=crcr, x index=0, y index=1, y error index=2,]{
                10 0.5332 0.0154\\
                25 0.5694 0.0112\\
                50 0.5947 0.0097\\
                75 0.6075 0.0037\\
                100 0.6116 0.0015\\
                };
                \addlegendentry{Resnet-IN}
                \end{axis}
            \end{tikzpicture}\hspace{8.0mm}\vspace{-1.0mm}
        \caption{\textbf{RFS -- All Patients}.}\vspace{3.0mm}
        \label{fig:TODO1}
    \end{subfigure}
    \begin{subfigure}[t]{0.45\textwidth}
        \centering%
            \begin{tikzpicture}[scale=0.75, baseline]
                \begin{axis}[
                    % xmode=log,
                    % log ticks with fixed point,
                    xlabel={\% of SöS-BC-4 used for training},
                    ylabel={C-index ($\uparrow$)},
                    xmin=5, xmax=105,
                    ymin=0.5, ymax=0.7125,
                    xtick={10, 25, 50, 75, 100},
                    ytick={0.50, 0.55, 0.60, 0.65, 0.70},
                    yticklabels={0.50, 0.55, 0.60, 0.65, 0.70},
                    legend pos=south east,
                    ymajorgrids=true,
                    %xmajorgrids=true,
                    title={},
                    grid style=dashed,
                    y tick label style={
                        /pgf/number format/.cd,
                            fixed,
                            % fixed zerofill,
                            precision=2,
                        /tikz/.cd
                    },
                    every axis plot/.append style={thick},
                ]
                \addplot [Cerulean, mark=triangle*, line width=0.6125mm]
                 plot [error bars/.cd, y dir = both, y explicit]
                 table[row sep=crcr, x index=0, y index=1, y error index=2,]{
                10 0.5651 0.0200\\
                25 0.6036 0.0100\\
                50 0.6436 0.0029\\
                75 0.6590 0.0033\\
                100 0.6662 0.0024\\
                };
                % \addlegendentry{H-optimus-1}
                    
                \addplot [flamingopink, mark=square*, line width=0.6125mm]
                 plot [error bars/.cd, y dir = both, y explicit]
                 table[row sep=crcr, x index=0, y index=1, y error index=2,]{
                10 0.5414 0.0228\\
                25 0.5802 0.0192\\
                50 0.6209 0.0177\\
                75 0.6416 0.0117\\
                100 0.6434 0.0071\\
                };
                % \addlegendentry{UNI}

                \addplot [gray, mark=*, line width=0.6125mm]
                 plot [error bars/.cd, y dir = both, y explicit]
                 table[row sep=crcr, x index=0, y index=1, y error index=2,]{
                10 0.5260 0.0129\\
                25 0.5603 0.0080\\
                50 0.5839 0.0101\\
                75 0.5959 0.0057\\
                100 0.5998 0.0018\\
                };
                % \addlegendentry{Resnet-IN}
                \end{axis}
            \end{tikzpicture}\hspace{8.0mm}\vspace{-1.0mm}
        \caption{\textbf{RFS -- Patient Subgroup (ER+ \& HER2-)}.}\vspace{3.0mm}
        \label{fig:TODO2}
    \end{subfigure}
    \begin{subfigure}[t]{0.45\textwidth}
        \centering%
            \begin{tikzpicture}[scale=0.75, baseline]
                \begin{axis}[
                    % xmode=log,
                    % log ticks with fixed point,
                    xlabel={\% of SöS-BC-4 used for training},
                    ylabel={C-index ($\uparrow$)},
                    xmin=5, xmax=105,
                    ymin=0.5, ymax=0.7125,
                    xtick={10, 25, 50, 75, 100},
                    ytick={0.50, 0.55, 0.60, 0.65, 0.70},
                    yticklabels={0.50, 0.55, 0.60, 0.65, 0.70},
                    legend pos=south east,
                    ymajorgrids=true,
                    %xmajorgrids=true,
                    title={},
                    grid style=dashed,
                    y tick label style={
                        /pgf/number format/.cd,
                            fixed,
                            % fixed zerofill,
                            precision=2,
                        /tikz/.cd
                    },
                    every axis plot/.append style={thick},
                ]
                \addplot [Cerulean, mark=triangle*, line width=0.6125mm]
                 plot [error bars/.cd, y dir = both, y explicit]
                 table[row sep=crcr, x index=0, y index=1, y error index=2,]{
                10 0.5973 0.0380\\
                25 0.6281 0.0202\\
                50 0.6722 0.0147\\
                75 0.6873 0.0047\\
                100 0.6966 0.0064\\
                };
                % \addlegendentry{H-optimus-1}
            
                \addplot [flamingopink, mark=square*, line width=0.6125mm]
                 plot [error bars/.cd, y dir = both, y explicit]
                 table[row sep=crcr, x index=0, y index=1, y error index=2,]{
                10 0.5831 0.0350\\
                25 0.6155 0.0072\\
                50 0.6530 0.0220\\
                75 0.6623 0.0088\\
                100 0.6654 0.0090\\
                };
                % \addlegendentry{UNI}

                \addplot [gray, mark=*, line width=0.6125mm]
                 plot [error bars/.cd, y dir = both, y explicit]
                 table[row sep=crcr, x index=0, y index=1, y error index=2,]{
                10 0.5625 0.0331\\
                25 0.5892 0.0147\\
                50 0.6082 0.0150\\
                75 0.6198 0.0120\\
                100 0.6276 0.0013\\
                };
                % \addlegendentry{Resnet-IN}
                \end{axis}
            \end{tikzpicture}\hspace{8.0mm}\vspace{-1.0mm}
        \caption{\textbf{PFS -- All Patients}.}
        \label{fig:TODO3}
    \end{subfigure}
    \begin{subfigure}[t]{0.45\textwidth}
        \centering%
            \begin{tikzpicture}[scale=0.75, baseline]
                \begin{axis}[
                    % xmode=log,
                    % log ticks with fixed point,
                    xlabel={\% of SöS-BC-4 used for training},
                    ylabel={C-index ($\uparrow$)},
                    xmin=5, xmax=105,
                    ymin=0.5, ymax=0.7125,
                    xtick={10, 25, 50, 75, 100},
                    ytick={0.50, 0.55, 0.60, 0.65, 0.70},
                    yticklabels={0.50, 0.55, 0.60, 0.65, 0.70},
                    legend pos=south east,
                    ymajorgrids=true,
                    %xmajorgrids=true,
                    title={},
                    grid style=dashed,
                    y tick label style={
                        /pgf/number format/.cd,
                            fixed,
                            % fixed zerofill,
                            precision=2,
                        /tikz/.cd
                    },
                    every axis plot/.append style={thick},
                ]
                \addplot [Cerulean, mark=triangle*, line width=0.6125mm]
                 plot [error bars/.cd, y dir = both, y explicit]
                 table[row sep=crcr, x index=0, y index=1, y error index=2,]{
                10 0.5793 0.0288\\
                25 0.6049 0.0159\\
                50 0.6424 0.0179\\
                75 0.6544 0.0092\\
                100 0.6660 0.0099\\
                };
                % \addlegendentry{H-optimus-1}
    
                \addplot [flamingopink, mark=square*, line width=0.6125mm]
                 plot [error bars/.cd, y dir = both, y explicit]
                 table[row sep=crcr, x index=0, y index=1, y error index=2,]{
                10 0.5633 0.0174\\
                25 0.5926 0.0049\\
                50 0.6216 0.0234\\
                75 0.6317 0.0107\\
                100 0.6328 0.0115\\
                };
                % \addlegendentry{UNI}

                \addplot [gray, mark=*, line width=0.6125mm]
                 plot [error bars/.cd, y dir = both, y explicit]
                 table[row sep=crcr, x index=0, y index=1, y error index=2,]{
                10 0.5495 0.0186\\
                25 0.5742 0.0118\\
                50 0.5874 0.0132\\
                75 0.5996 0.0098\\
                100 0.6054 0.0012\\
                };
                % \addlegendentry{Resnet-IN}
                \end{axis}
            \end{tikzpicture}\hspace{8.0mm}\vspace{-1.0mm}
        \caption{\textbf{PFS -- Patient Subgroup (ER+ \& HER2-)}.}
        \label{fig:TODO4}
    \end{subfigure}
    \caption{\textbf{Effect of training data size on survival prediction performance for three representative models}. C-index performance for \textit{Resnet-IN}, \textit{UNI}, and \textit{H-optimus-1} as a function of the fraction of training data used (10\%, 25\%, 50\%, 75\%, 100\%) across all four evaluation settings (RFS and PFS, `All Patients' and `ER+ \& HER2-'). Results are reported as mean $\pm$ standard deviation (std) over five random seeds, where a new random subset of training patients is sampled for each seed at each data fraction. The PANTHER prototype estimation step is performed once on the full SöS-BC-4 dataset for each seed, after which the survival model is retrained on the corresponding subset.}
  \label{fig:subsets_h1_uni_resnetin}
\end{figure*}
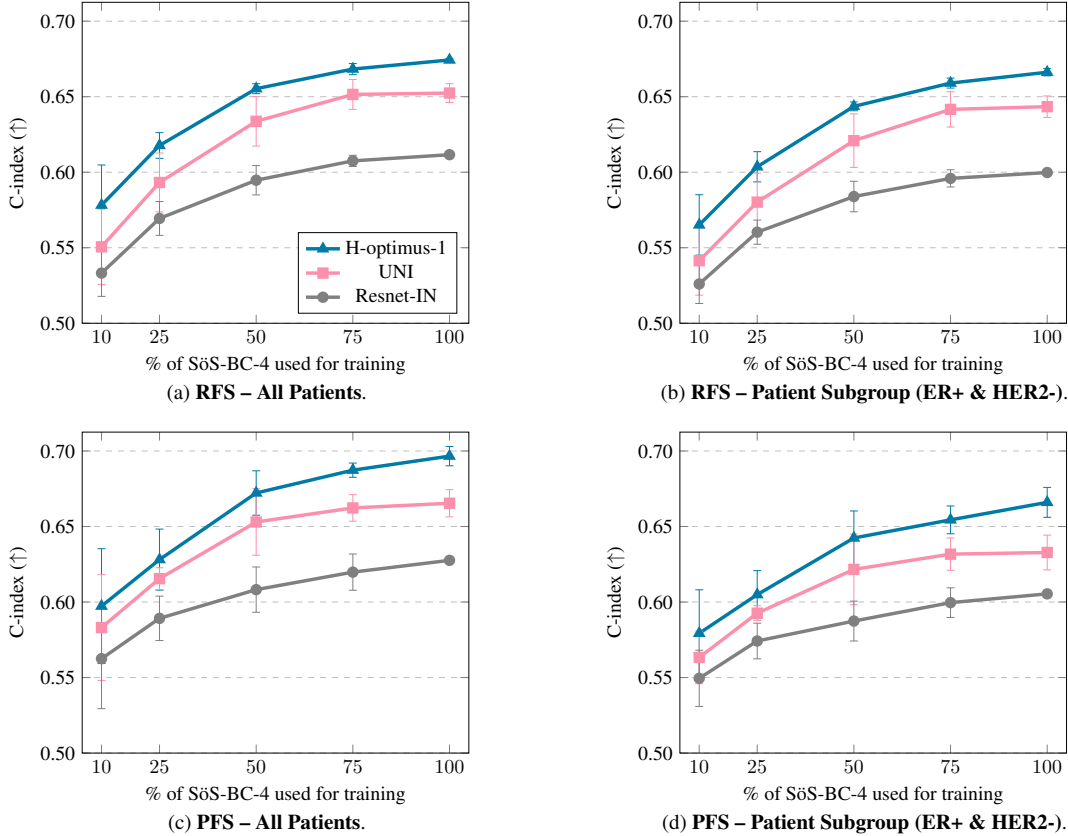

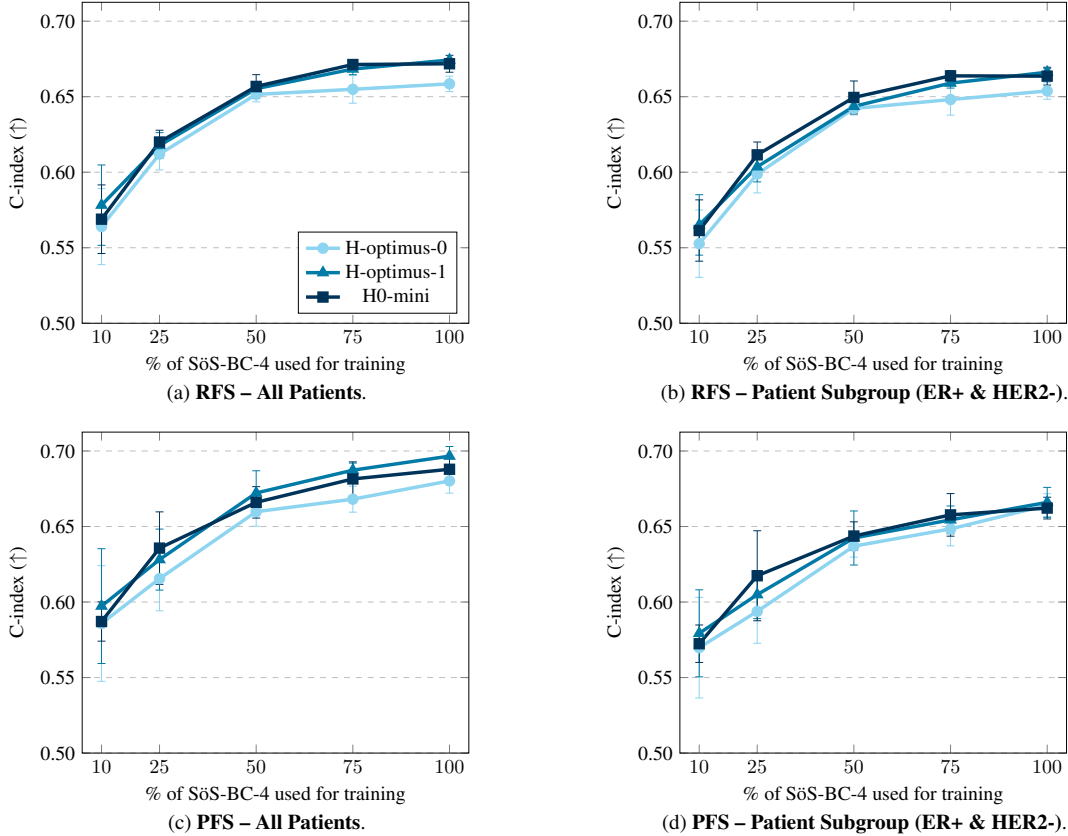
\begin{figure*}[t]
\centering
    \begin{subfigure}[t]{0.45\textwidth}
        \centering%
            \begin{tikzpicture}[scale=0.75, baseline]
                \begin{axis}[
                    % xmode=log,
                    % log ticks with fixed point,
                    xlabel={\% of SöS-BC-4 used for training},
                    ylabel={C-index ($\uparrow$)},
                    xmin=5, xmax=105,
                    ymin=0.5, ymax=0.7125,
                    xtick={10, 25, 50, 75, 100},
                    ytick={0.50, 0.55, 0.60, 0.65, 0.70},
                    yticklabels={0.50, 0.55, 0.60, 0.65, 0.70},
                    legend pos=south east,
                    ymajorgrids=true,
                    %xmajorgrids=true,
                    title={},
                    grid style=dashed,
                    y tick label style={
                        /pgf/number format/.cd,
                            fixed,
                            % fixed zerofill,
                            precision=2,
                        /tikz/.cd
                    },
                    every axis plot/.append style={thick},
                ]
                \addplot [skyblue2, mark=*, line width=0.6125mm]
                 plot [error bars/.cd, y dir = both, y explicit]
                 table[row sep=crcr, x index=0, y index=1, y error index=2,]{
                10 0.5640 0.0252\\
                25 0.6120 0.0105\\
                50 0.6515 0.0049\\
                75 0.6549 0.0092\\
                100 0.6585 0.0051\\
                };
                \addlegendentry{H-optimus-0}
                
                \addplot [Cerulean, mark=triangle*, line width=0.6125mm]
                 plot [error bars/.cd, y dir = both, y explicit]
                 table[row sep=crcr, x index=0, y index=1, y error index=2,]{
                10 0.5782 0.0266\\
                25 0.6177 0.0085\\
                50 0.6553 0.0033\\
                75 0.6683 0.0037\\
                100 0.6743 0.0009\\
                };
                \addlegendentry{H-optimus-1}
            
                \addplot [darkcerulean2, mark=square*, line width=0.6125mm]
                 plot [error bars/.cd, y dir = both, y explicit]
                 table[row sep=crcr, x index=0, y index=1, y error index=2,]{
                10 0.5689 0.0227\\
                25 0.6200 0.0078\\
                50 0.6568 0.0078\\
                75 0.6713 0.0022\\
                100 0.6718 0.0056\\
                };
                \addlegendentry{H0-mini}
                \end{axis}
            \end{tikzpicture}\hspace{8.0mm}\vspace{-1.0mm}
        \caption{\textbf{RFS -- All Patients}.}\vspace{3.0mm}
        \label{fig:TODO1_2}
    \end{subfigure}
    \begin{subfigure}[t]{0.45\textwidth}
        \centering%
            \begin{tikzpicture}[scale=0.75, baseline]
                \begin{axis}[
                    % xmode=log,
                    % log ticks with fixed point,
                    xlabel={\% of SöS-BC-4 used for training},
                    ylabel={C-index ($\uparrow$)},
                    xmin=5, xmax=105,
                    ymin=0.5, ymax=0.7125,
                    xtick={10, 25, 50, 75, 100},
                    ytick={0.50, 0.55, 0.60, 0.65, 0.70},
                    yticklabels={0.50, 0.55, 0.60, 0.65, 0.70},
                    legend pos=south east,
                    ymajorgrids=true,
                    %xmajorgrids=true,
                    title={},
                    grid style=dashed,
                    y tick label style={
                        /pgf/number format/.cd,
                            fixed,
                            % fixed zerofill,
                            precision=2,
                        /tikz/.cd
                    },
                    every axis plot/.append style={thick},
                ]
                \addplot [skyblue2, mark=*, line width=0.6125mm]
                 plot [error bars/.cd, y dir = both, y explicit]
                 table[row sep=crcr, x index=0, y index=1, y error index=2,]{
                10 0.5527 0.0223\\
                25 0.5988 0.0125\\
                50 0.6422 0.0043\\
                75 0.6481 0.0103\\
                100 0.6538 0.0055\\
                };
                % \addlegendentry{H-optimus-0}
                
                \addplot [Cerulean, mark=triangle*, line width=0.6125mm]
                 plot [error bars/.cd, y dir = both, y explicit]
                 table[row sep=crcr, x index=0, y index=1, y error index=2,]{
                10 0.5651 0.0200\\
                25 0.6036 0.0100\\
                50 0.6436 0.0029\\
                75 0.6590 0.0033\\
                100 0.6662 0.0024\\
                };
                % \addlegendentry{H-optimus-1}
                    
                \addplot [darkcerulean2, mark=square*, line width=0.6125mm]
                 plot [error bars/.cd, y dir = both, y explicit]
                 table[row sep=crcr, x index=0, y index=1, y error index=2,]{
                10 0.5614 0.0203\\
                25 0.6115 0.0085\\
                50 0.6495 0.0109\\
                75 0.6638 0.0034\\
                100 0.6636 0.0059\\
                };
                % \addlegendentry{H0-mini}
                \end{axis}
            \end{tikzpicture}\hspace{8.0mm}\vspace{-1.0mm}
        \caption{\textbf{RFS -- Patient Subgroup (ER+ \& HER2-)}.}\vspace{3.0mm}
        \label{fig:TODO2_2}
    \end{subfigure}
    \begin{subfigure}[t]{0.45\textwidth}
        \centering%
            \begin{tikzpicture}[scale=0.75, baseline]
                \begin{axis}[
                    % xmode=log,
                    % log ticks with fixed point,
                    xlabel={\% of SöS-BC-4 used for training},
                    ylabel={C-index ($\uparrow$)},
                    xmin=5, xmax=105,
                    ymin=0.5, ymax=0.7125,
                    xtick={10, 25, 50, 75, 100},
                    ytick={0.50, 0.55, 0.60, 0.65, 0.70},
                    yticklabels={0.50, 0.55, 0.60, 0.65, 0.70},
                    legend pos=south east,
                    ymajorgrids=true,
                    %xmajorgrids=true,
                    title={},
                    grid style=dashed,
                    y tick label style={
                        /pgf/number format/.cd,
                            fixed,
                            % fixed zerofill,
                            precision=2,
                        /tikz/.cd
                    },
                    every axis plot/.append style={thick},
                ]
                \addplot [skyblue2, mark=*, line width=0.6125mm]
                 plot [error bars/.cd, y dir = both, y explicit]
                 table[row sep=crcr, x index=0, y index=1, y error index=2,]{
                10 0.5858 0.0383\\
                25 0.6155 0.0213\\
                50 0.6598 0.0096\\
                75 0.6681 0.0086\\
                100 0.6802 0.0081\\
                };
                % \addlegendentry{H-optimus-0}

                \addplot [Cerulean, mark=triangle*, line width=0.6125mm]
                 plot [error bars/.cd, y dir = both, y explicit]
                 table[row sep=crcr, x index=0, y index=1, y error index=2,]{
                10 0.5973 0.0380\\
                25 0.6281 0.0202\\
                50 0.6722 0.0147\\
                75 0.6873 0.0047\\
                100 0.6966 0.0064\\
                };
                % \addlegendentry{H-optimus-1}
            
                \addplot [darkcerulean2, mark=square*, line width=0.6125mm]
                 plot [error bars/.cd, y dir = both, y explicit]
                 table[row sep=crcr, x index=0, y index=1, y error index=2,]{
                10 0.5871 0.0130\\
                25 0.6357 0.0240\\
                50 0.6660 0.0104\\
                75 0.6815 0.0113\\
                100 0.6879 0.0072\\
                };
                % \addlegendentry{H0-mini}
                \end{axis}
            \end{tikzpicture}\hspace{8.0mm}\vspace{-1.0mm}
        \caption{\textbf{PFS -- All Patients}.}
        \label{fig:TODO3_2}
    \end{subfigure}
    \begin{subfigure}[t]{0.45\textwidth}
        \centering%
            \begin{tikzpicture}[scale=0.75, baseline]
                \begin{axis}[
                    % xmode=log,
                    % log ticks with fixed point,
                    xlabel={\% of SöS-BC-4 used for training},
                    ylabel={C-index ($\uparrow$)},
                    xmin=5, xmax=105,
                    ymin=0.5, ymax=0.7125,
                    xtick={10, 25, 50, 75, 100},
                    ytick={0.50, 0.55, 0.60, 0.65, 0.70},
                    yticklabels={0.50, 0.55, 0.60, 0.65, 0.70},
                    legend pos=south east,
                    ymajorgrids=true,
                    %xmajorgrids=true,
                    title={},
                    grid style=dashed,
                    y tick label style={
                        /pgf/number format/.cd,
                            fixed,
                            % fixed zerofill,
                            precision=2,
                        /tikz/.cd
                    },
                    every axis plot/.append style={thick},
                ]
                \addplot [skyblue2, mark=*, line width=0.6125mm]
                 plot [error bars/.cd, y dir = both, y explicit]
                 table[row sep=crcr, x index=0, y index=1, y error index=2,]{
                10 0.5698 0.0333\\
                25 0.5938 0.0211\\
                50 0.6370 0.0073\\
                75 0.6484 0.0112\\
                100 0.6641 0.0076\\
                };
                % \addlegendentry{H-optimus-0}
                
                \addplot [Cerulean, mark=triangle*, line width=0.6125mm]
                 plot [error bars/.cd, y dir = both, y explicit]
                 table[row sep=crcr, x index=0, y index=1, y error index=2,]{
                10 0.5793 0.0288\\
                25 0.6049 0.0159\\
                50 0.6424 0.0179\\
                75 0.6544 0.0092\\
                100 0.6660 0.0099\\
                };
                % \addlegendentry{H-optimus-1}
    
                \addplot [darkcerulean2, mark=square*, line width=0.6125mm]
                 plot [error bars/.cd, y dir = both, y explicit]
                 table[row sep=crcr, x index=0, y index=1, y error index=2,]{
                10 0.5724 0.0124\\
                25 0.6174 0.0298\\
                50 0.6437 0.0094\\
                75 0.6577 0.0141\\
                100 0.6622 0.0072\\
                };
                % \addlegendentry{H0-mini}
                \end{axis}
            \end{tikzpicture}\hspace{8.0mm}\vspace{-1.0mm}
        \caption{\textbf{PFS -- Patient Subgroup (ER+ \& HER2-)}.}
        \label{fig:TODO4_2}
    \end{subfigure}
    \caption{\textbf{Effect of training data size on survival prediction performance for the three top-performing models}. C-index performance for \textit{H-optimus-0}, \textit{H-optimus-1}, and \textit{H0-mini} as a function of the fraction of training data used (10\%, 25\%, 50\%, 75\%, 100\%) across all four evaluation settings. The experimental setup is identical to Figure~\ref{fig:subsets_h1_uni_resnetin}, with results reported as mean $\pm$ std over five random seeds.}
  \label{fig:subsets_h0_h1_h0-mini}
\end{figure*}

\begin{figure*}[t]
\centering
    \begin{subfigure}[t]{0.33\textwidth}
        \centering%
        \includegraphics[clip, trim=1.5cm 0.95cm 1.85cm 0.95cm, width=1.0\linewidth]{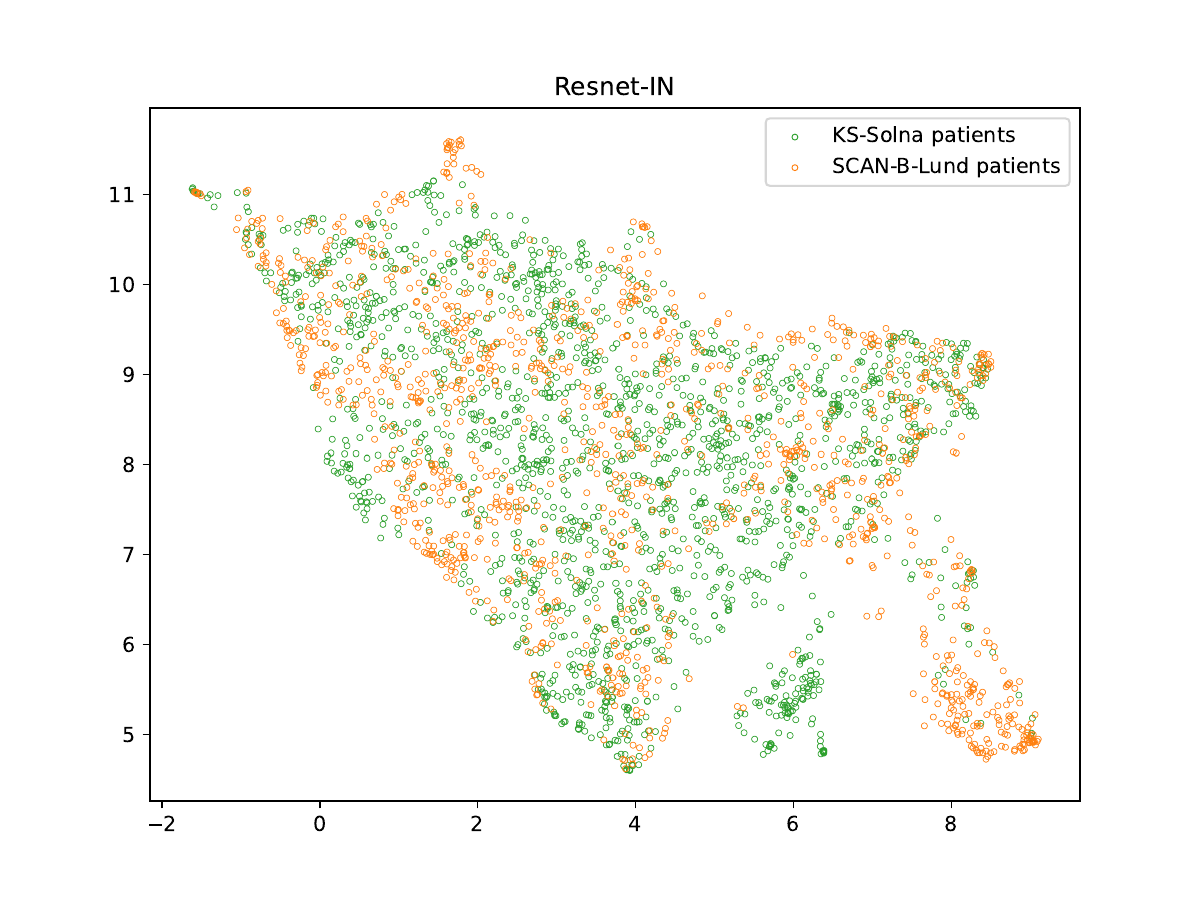}
    \end{subfigure}
    \begin{subfigure}[t]{0.33\textwidth}
        \centering%
        \includegraphics[clip, trim=1.5cm 0.95cm 1.85cm 0.95cm, width=1.0\linewidth]{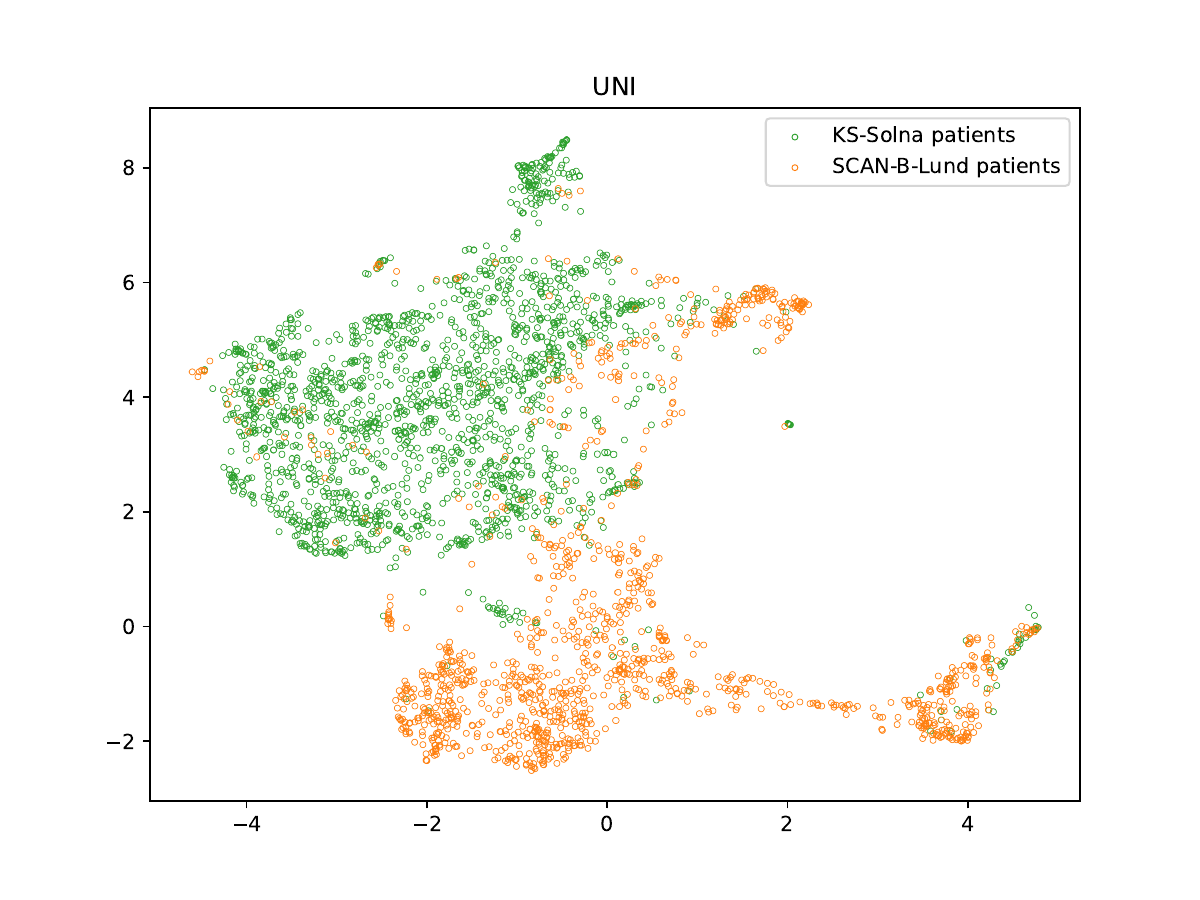}
    \end{subfigure}
    \begin{subfigure}[t]{0.33\textwidth}
        \centering%
        \includegraphics[clip, trim=1.5cm 0.95cm 1.85cm 0.95cm, width=1.0\linewidth]{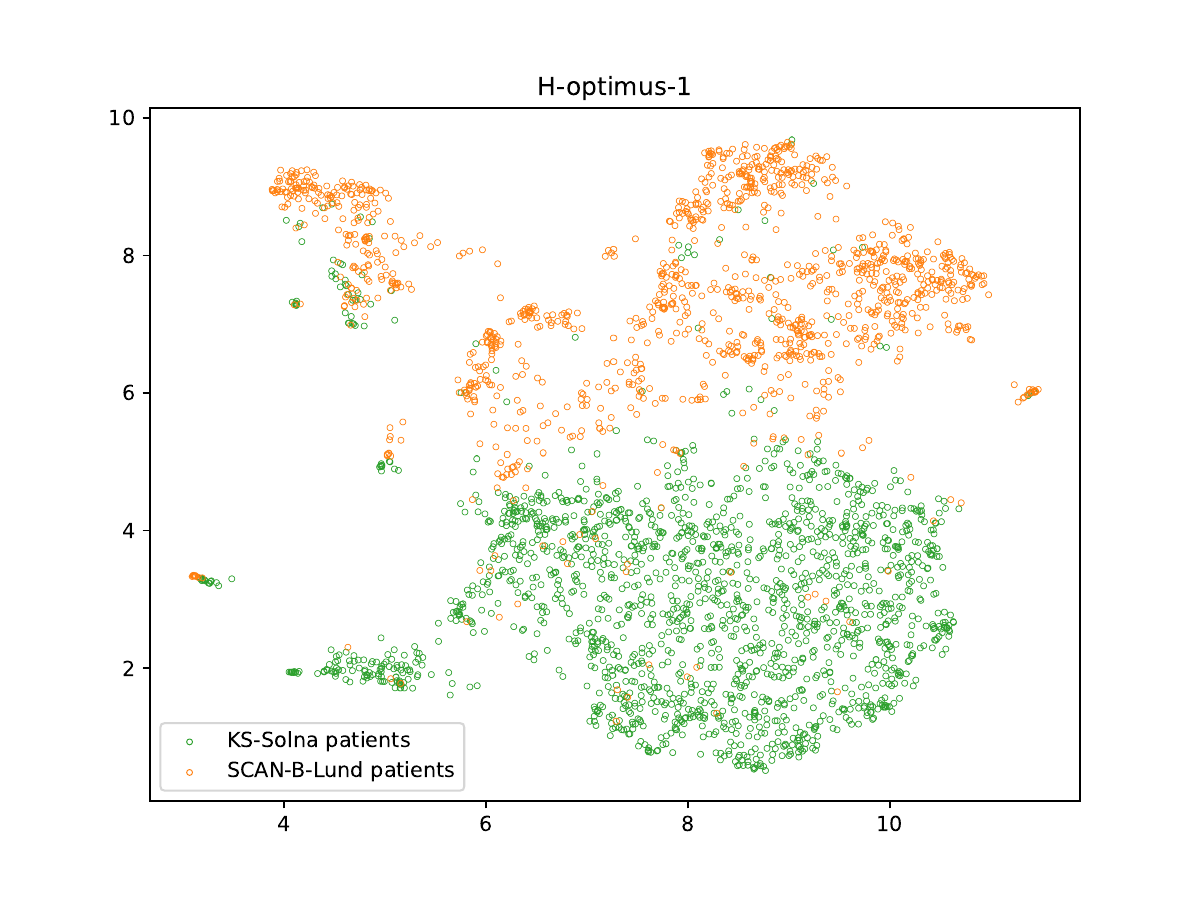}
    \end{subfigure}
    
    \begin{subfigure}[t]{0.33\textwidth}
        \centering%
        \includegraphics[clip, trim=1.5cm 0.95cm 1.85cm 0.95cm, width=1.0\linewidth]{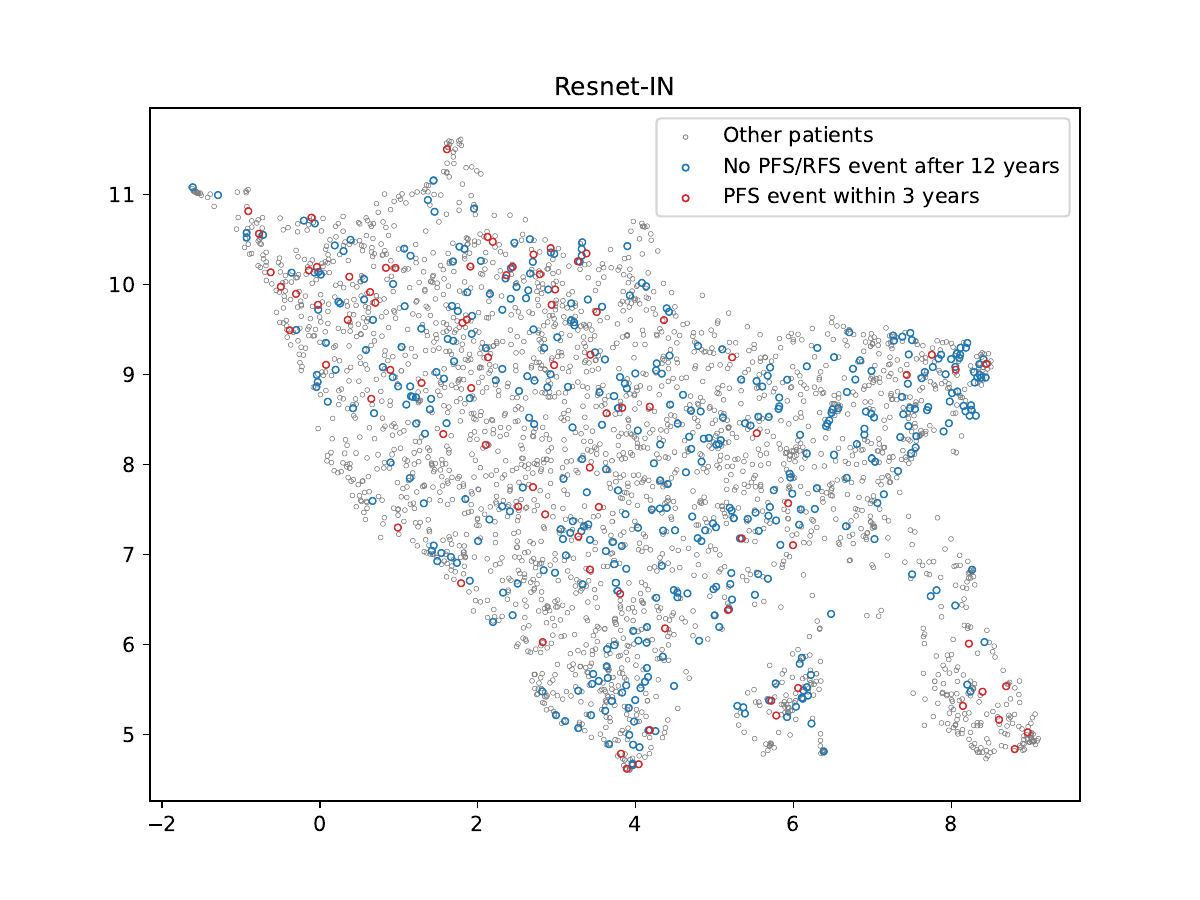}
    \end{subfigure}
    \begin{subfigure}[t]{0.33\textwidth}
        \centering%
        \includegraphics[clip, trim=1.5cm 0.95cm 1.85cm 0.95cm, width=1.0\linewidth]{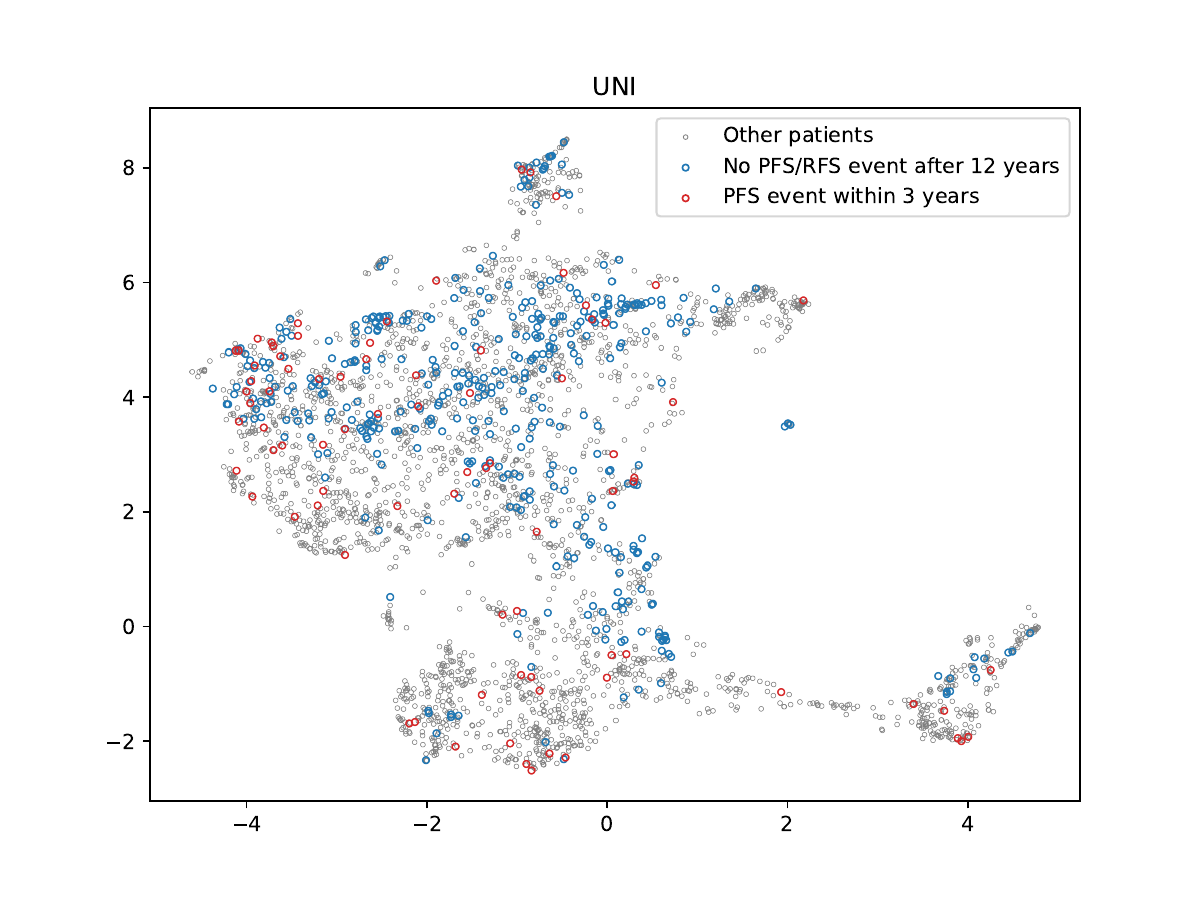}
    \end{subfigure}
    \begin{subfigure}[t]{0.33\textwidth}
        \centering%
        \includegraphics[clip, trim=1.5cm 0.95cm 1.85cm 0.95cm, width=1.0\linewidth]{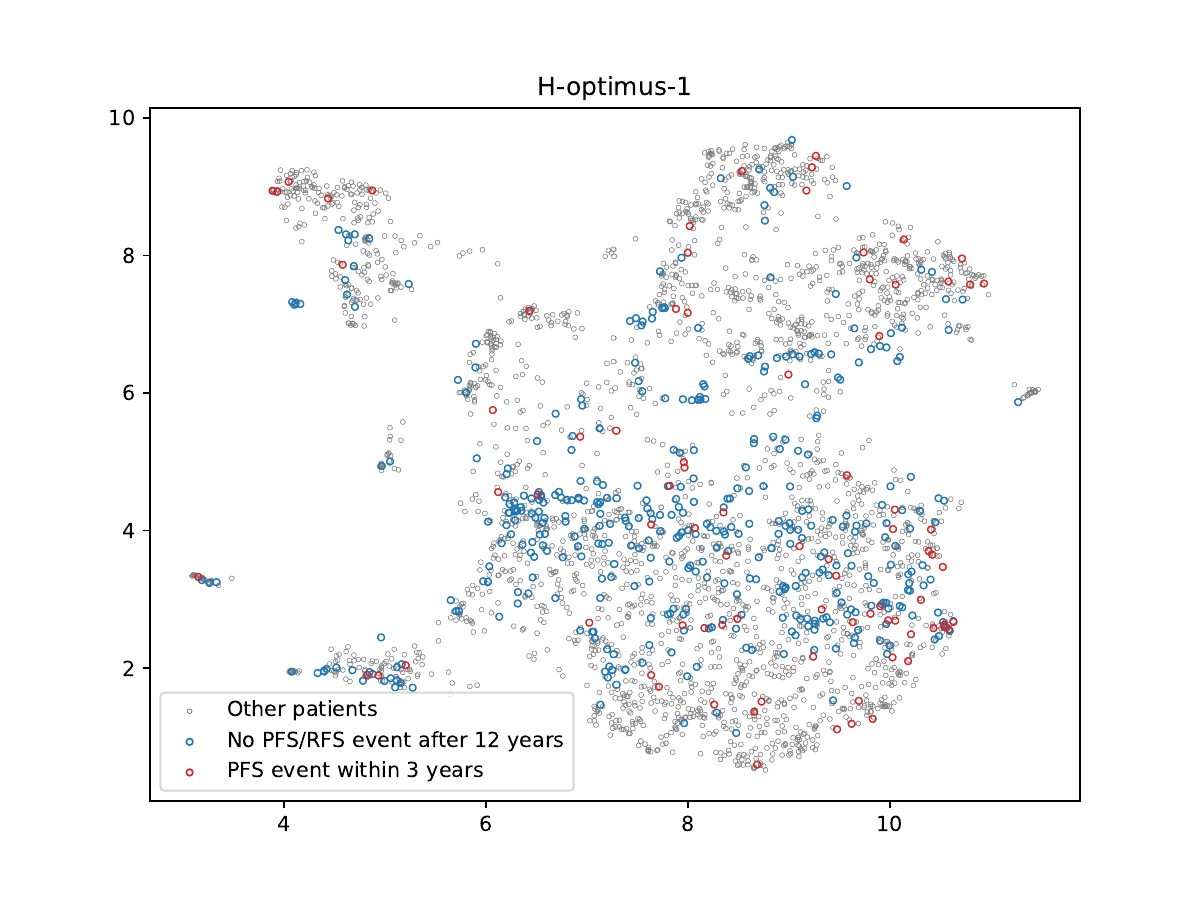}
    \end{subfigure}
    \caption{\textbf{UMAP visualizations of learned feature representations for three representative models}. UMAP projections of the mean patch-level feature vectors for \textit{Resnet-IN} (left column), \textit{UNI} (middle), and \textit{H-optimus-1} (right) on the combined KS-Solna and SCAN-B-Lund evaluation set, with one point per patient. \textbf{Upper row:} points are colored by dataset, \textcolor{ForestGreen}{\textbf{green}} for KS-Solna and \textcolor{goldenpoppy3}{\textbf{orange}} for SCAN-B-Lund. \textbf{Lower row:} points are colored by survival outcome, \textcolor{OrangeRed2}{\textbf{red}} for patients with a PFS event within three years (82 patients), \textcolor{Cerulean}{\textbf{blue}} for patients with no PFS/RFS event and a follow-up time of at least 12 years (407 patients), and \textcolor{gray}{\textbf{gray}} for all other patients (2{,}630 patients).}
  \label{fig:umap_plots}
\end{figure*}

\subsubsection*{Extended Model Analysis}
Figure~\ref{fig:subsets_h1_uni_resnetin} shows C-index performance across all four evaluation settings for Resnet-IN, UNI, and H-optimus-1 as a function of the amount of training data used from the SöS-BC-4 dataset (10\%, 25\%, 50\%, 75\%, 100\%). Across all subset sizes, consistent trends are observed in both absolute performance and relative ranking. H-optimus-1 achieves the highest performance across all settings and training set sizes, followed by UNI and Resnet-IN. Importantly, the relative ordering of models remains unchanged as the amount of training data varies, indicating that performance differences are robust across data regimes.

Performance improves for all models as the amount of training data increases. While the performance gap between models remains relatively stable across subset sizes, a slight widening of the gap between H-optimus-1 and Resnet-IN is observed. For example, in the \textit{RFS -- All Patients} setting, H-optimus-1 improves from a C-index of 0.578 at 10\% of the training data to 0.674 at 100\%, corresponding to a relative improvement of 16.6\%. In comparison, Resnet-IN improves from 0.533 to 0.612 (14.7\% relative improvement). A similar trend is observed for \textit{PFS -- All Patients}, where H-optimus-1 improves from 0.597 to 0.696 (16.6\%), while Resnet-IN improves from 0.562 to 0.627 (11.6\%). Comparable patterns are observed across the remaining settings.

Across all evaluation settings, H-optimus-1 trained on 25\% of the data achieves performance matching or exceeding that of Resnet-IN trained on the full dataset. Similarly, H-optimus-1 trained on 50\% of the data matches or exceeds the performance of UNI trained on 100\% of the data.
% H-optimus-1 at 25:
% 0.6177
% 0.6036
% 0.6281
% 0.6049
%%
% Resnet-IN at 100:
% 0.6116
% 0.5998
% 0.6276
% 0.6054
%%
% H-optimus-1 at 50:
% 0.6553
% 0.6436
% 0.6722
% 0.6424
%%
% UNI at 100:
% 0.6524
% 0.6434
% 0.6654
% 0.6328

Figure~\ref{fig:subsets_h0_h1_h0-mini} shows the same analysis of C-index performance across subset sizes for the three highest-ranked models overall: H-optimus-1, H0-mini, and H-optimus-0. These models exhibit very similar performance across all subset sizes and evaluation settings. Notably, the compact distilled model H0-mini consistently performs on par with, and in most cases even slightly better than, its teacher model H-optimus-0. Overall, the performance curves of these three models are closely aligned across all settings, with only minor differences in C-index values.

Finally, to further investigate the structure of learned feature representations, Figure~\ref{fig:umap_plots} shows UMAP~\citep{mcinnes2018umap} projections of the mean patch-level feature vectors for Resnet-IN, UNI, and H-optimus-1 on the combined KS-Solna and SCAN-B-Lund evaluation set, with one point per patient. In the upper row, points are colored by dataset (KS-Solna vs SCAN-B-Lund), while in the lower row, points are colored according to survival outcome. Specifically, points are colored red for patients with a PFS event within three years, blue for patients with no PFS/RFS event and a follow-up time of at least 12 years, and gray otherwise.

Clear clustering by dataset is observed for both UNI and H-optimus-1 in the first two UMAP components, indicating that these models capture systematic differences between WSIs from different cohorts, while Resnet-IN shows no clear separation of the two datasets. In contrast, clustering by survival outcome is not clearly observed for any of the models, suggesting that survival-related variability is not dominating the variability in the data. Despite the substantially better survival prediction performance of H-optimus-1 compared to Resnet-IN, early-event and event-free patients are not clearly separated in the feature space, as visualized by the first two UMAP components, of either model.

\section*{Discussion}

In this study, we present a large-scale, multi-cohort benchmark of PFMs for breast cancer survival prediction, providing a systematic comparison across widely used models. Several key insights emerge from our results.

First, while H-optimus-1 achieves the strongest overall performance, the absolute differences between top-performing models are relatively small. Across all four evaluation settings, multiple recent PFMs including H-optimus-1, H0-mini, H-optimus-0, CONCHv1.5, UNI2-h, and Virchow2 achieve C-index values within a narrow range, with substantially overlapping confidence intervals. This suggests that although architectural and training improvements lead to consistent ranking gains, these improvements are incremental rather than transformative. From a practical perspective, this implies that model selection should not be based solely on marginal performance differences, but also on factors such as computational efficiency, robustness, and access to models.

A particularly notable finding is the strong performance of the compact distilled model H0-mini, which slightly outperforms its teacher model H-optimus-0 while using fewer than 8\% of the parameters and enabling significantly faster feature extraction. This result is consistent across both full-data and subset analyses (Figure~\ref{fig:main_results}~\&~\ref{fig:subsets_h0_h1_h0-mini}), where H0-mini matches or exceeds H-optimus-0 across evaluation settings. Given the substantial computational cost associated with WSI processing, this highlights knowledge distillation as a highly promising strategy for developing efficient PFMs without sacrificing predictive performance. Notably, H0-mini is distilled using a small public dataset of just 6{,}000 WSIs, further emphasizing the potential of distillation to transfer representational knowledge efficiently. Taken together, these findings suggest that compact models such as H0-mini offer a favorable trade-off between performance and efficiency, and may represent a strong candidate for resource-constrained settings.

Across model families, we observe consistent generational improvements, with second-generation PFMs (H-optimus-1, CONCHv1.5, UNI2-h, Virchow2) outperforming their respective first-generation counterparts (H-optimus-0, CONCH, UNI, Virchow) according to the aggregated ranking. However, these gains remain modest despite substantial increases in pretraining scale. For example, H-optimus-1 is trained on twice as many WSIs as H-optimus-0 (1 million vs 0.5 million WSIs, Table~\ref{table:fms}), and Virchow2 similarly scales pretraining data compared to Virchow, yet both yield only incremental improvements in downstream performance. This suggests that scaling pretraining data alone may yield diminishing returns.

Similarly, model size alone is not a reliable predictor of performance. Despite being among the very largest models evaluated (1.1 billion parameters), Prov-GigaPath places only 9th in the overall ranking, while substantially smaller models such as H0-mini and CONCH (86 million parameters) achieve competitive or superior performance. This further reinforces the notion that aspects such as pretraining data quality and training strategy are more critical than raw model capacity.

Our results also provide tentative evidence regarding vision-language pretraining. CONCH and CONCHv1.5 perform competitively for prognostic stratification despite relatively smaller model sizes, suggesting that incorporating image-caption pairs and multimodal alignment objectives in the pretraining may improve representation learning efficiency. However, it remains unclear whether these gains arise from the multimodal training paradigm itself or from the scale and diversity of the underlying pretraining data. Furthermore, given the limited number of vision-language models evaluated and the lack of controlled comparisons, no definitive conclusions can be drawn. Future work should explore this direction more systematically, for example through ablation studies comparing vision-only and multimodal pretraining under controlled settings.

Consistent with expectations, the natural-image baseline Resnet-IN and early pathology-specific models (CTransPath, RetCCL) perform consistently worse than recent PFMs, confirming the importance of large-scale domain-specific pretraining. At the same time, it is notable that even Resnet-IN achieves non-trivial performance and can stratify patients into two risk groups (Figure~\ref{fig:km_plots_2groups_counts}), indicating that some prognostic morphological signals are sufficiently generic to be captured by natural-image features. However, while Kaplan-Meier analysis shows that all models capture prognostic signal to some extent, they differ in the strength and stability of risk stratification. In particular, the advantage of PFMs becomes more pronounced in more demanding settings, such as multi-group risk stratification (Figure~\ref{fig:km_plots_4groups}), where models like H-optimus-1 produce more clearly separated and consistently ordered survival curves. Notably, the ability to stratify patients into four distinct risk groups is non-trivial and provides a more stringent test of model quality than binary stratification. The fact that top-performing PFMs maintain clear separation in this setting suggests that they capture clinically meaningful gradations of risk.

Despite broadly consistent rankings across survival endpoints, we observe some task-dependent variability. CONCHv1.5 and Virchow rank relatively low for RFS but among the top performers for PFS. One likely explanation is the smaller number of PFS events relative to RFS, leading to higher variance and wider confidence intervals. These findings may also reflect differences in the types of prognostic signals captured by different models, although the current results are not sufficient to draw firm conclusions. Regardless, these results highlight the importance of evaluating models across multiple clinically relevant endpoints.

The data efficiency analysis (Figure~\ref{fig:subsets_h1_uni_resnetin}~\&~\ref{fig:subsets_h0_h1_h0-mini}) provides additional insight into model behavior. Performance improves steadily with increasing training data across models, with similar relative gains observed for both strong and weak feature extractors. This suggests that downstream survival modeling benefits consistently from increased supervision, largely independent of the choice of encoder. At the same time, stronger PFMs maintain a consistent performance advantage across all data regimes, and notably achieve competitive performance even with limited training data. For example, H-optimus-1 trained on 25\% of the data matches or exceeds the performance of Resnet-IN trained on the full dataset. These findings indicate that both better representations and larger labeled datasets contribute independently to improved survival prediction, and that further gains may be achievable by scaling either dimension.

From a data perspective, an important implication is that increasing the number of observed survival events, rather than simply the number of patients, is likely critical for improving model performance, particularly for endpoints such as PFS where event counts remain relatively low. This reflects the fact that time-to-event models rely primarily on observed events for learning when optimizing the Cox proportional hazards loss. Consequently, sufficiently long follow-up time, together with larger studies or aggregation of multiple cohorts, is expected to contribute to improved performance.

Finally, analysis of the learned feature space using UMAP (Figure~\ref{fig:umap_plots}) reveals that while PFMs capture dataset-specific structure, clear separation by survival outcome is not observed. Notably, even though models such as H-optimus-1 substantially outperform Resnet-IN in survival prediction, this improvement is not reflected in obvious clustering patterns. This suggests that prognostic signals are likely subtle, distributed, and not easily separable in low-dimensional projections, particularly when compared to more pronounced site- or cohort-specific differences.

Overall, our findings highlight both the progress and current limitations of pathology foundation models for survival prediction. While recent PFMs provide consistent improvements over earlier approaches, performance gains are mostly incremental, and multiple models achieve comparable results. In this context, efficiency emerges as a key consideration, particularly in resource-constrained settings, with compact distilled models such as H0-mini potentially offering a compelling balance between performance and scalability.

% Limitations / future work:
This study has several limitations. Although multiple independent datasets were used for evaluation, all cohorts originate from similar healthcare systems within a single country (Sweden) and may not fully capture global variability in clinical practice, staining protocols, or scanner hardware. In addition, all experiments are conducted on breast cancer cohorts, and results may not generalize to other cancer types or disease settings, where different morphological patterns and prognostic signals may influence model performance and relative rankings. We evaluate models using frozen feature extractors, within a unified pipeline based on PANTHER aggregation and an MLP survival head, to ensure a fair and controlled comparison across models. While this isolates the quality of pretrained representations, results and model rankings may differ under task-specific fine-tuning, end-to-end training, or alternative survival modeling approaches. The relatively small performance differences observed between many state-of-the-art PFMs, together with overlapping confidence intervals, suggest that further scaling of pretraining data or model size alone may yield diminishing returns. Alternative directions such as vision-language pretraining and knowledge distillation appear promising, with our results providing especially strong evidence for the effectiveness of distillation, but both require more systematic investigation in controlled settings. Distillation may be particularly valuable in resource-constrained settings due to its favorable efficiency–performance trade-off. At the same time, further work is needed to assess how well these gains generalize across tasks and applications, as distillation may in some cases lead to a loss of more fine-grained or task-specific information in the learned representations, even if such effects are not observed in the present evaluation. In particular, the strong performance of H0-mini, distilled using a small dataset of just 6{,}000 WSIs, raises important questions about the role of distillation data scale and composition, which should be explored in future work.

\emph{The main takeaways from our study are:}
(1) H-optimus-1 achieves the strongest overall performance, but absolute differences between top-performing PFMs are small and confidence intervals substantially overlap. While architectural and training improvements lead to consistent performance gains, these improvements are incremental rather than transformative.
(2) Across model families, consistent generational improvements are observed, with second-generation PFMs (H-optimus-1, CONCHv1.5, UNI2-h, Virchow2) outperforming their respective first-generation counterparts. However, these gains remain modest despite substantial increases in pretraining data scale, suggesting diminishing returns from scaling alone.
(3) Model size is not a reliable predictor of performance, as smaller and more efficient models can match or exceed much larger architectures, emphasizing the importance of training strategy and pretraining data quality over model scaling.
(4) The compact distilled model H0-mini achieves the second-best overall ranking and slightly outperforms its much larger teacher model H-optimus-0, while using less than 8\% of the parameters and enabling significantly faster feature extraction. This demonstrates that knowledge distillation can yield highly efficient PFMs without sacrificing predictive performance, making it a particularly promising approach for resource-constrained settings.
(5) Strong pretrained feature extractors and large labeled datasets contribute independently to improved survival prediction performance, with high-quality PFMs maintaining advantages also in low-data regimes. Increasing the number of observed survival events, for example through longer follow-up time, is likely to be key for further improvements.

\section*{Methods}

We benchmark pretrained PFMs for WSI-based breast cancer survival prediction using a unified experimental setup. The workflow consists of WSI preprocessing and patch extraction, patch-level feature extraction using frozen pretrained models, and slide-level survival prediction using the prototype-based aggregation approach PANTHER~\citep{song2024morphological}. Survival models are trained on SöS-BC-4 and evaluated on the independent external datasets KS-Solna and SCAN-B-Lund to assess generalization. The following sections describe the survival endpoints, datasets, preprocessing pipeline, survival model, evaluation framework, and the evaluated PFMs and baselines.

\subsubsection*{Survival Endpoints}
We evaluate two clinically relevant survival endpoints: recurrence-free survival (RFS) and progression-free survival (PFS). RFS is defined as the time from initial diagnosis to disease recurrence or death from any cause. Recurrence includes local recurrence, distant metastasis, or detection of contralateral tumors. Patients without an event are censored at the date of last follow-up. In contrast, PFS is defined as the time from initial diagnosis to disease recurrence. Death without documented recurrence is not counted as an event and is treated as censoring. 

Accordingly, RFS captures both recurrence and mortality events, whereas PFS focuses specifically on recurrence events. Evaluating both endpoints provides complementary perspectives: PFS focuses specifically on disease progression, while RFS includes mortality events and therefore yields a larger number of events for statistical analysis.

% "Progression-free survival (PFS) was the survival endpoint, defined as the time to local recurrence, distant metastasis or detection of contralateral tumours. The patients were followed from the initial date of diagnosis to the date of reported progression, or the last follow-up date (whichever came first)......."

% "Since the clinical NHG has high inter-rater variability, we utilise patient outcome (recurrence-free survival (RFS)), as our primary evaluation metric..... The RFS defined recurrence (i.e. local or distant metastasis, detection of contralateral tumours) or death as the event outcome. Patients were followed from the initial diagno- sis to the date of death/recurrence, emigration, or the last registration date, whichever occurred first....."

% "recurrence-free survival (RFS) as the survival endpoint that was defined as recurrence (i.e. local or distant metastasis, detection of contralateral tumours) or death as the event outcome. Patients were followed from the initial diagnosis to the date of death/recurrence, emigration, or the last registration date, whichever occurred first. In study II, we used progression-free survival (PFS) as the survival endpoint that was defined as the disease specific recurrences (i.e. local or distant metastasis, detection of contralateral tumours) as the event outcome. The patient follow-up time was defined from the initial diagnosis to the date of recurrence, emigration, or the last registration date, whichever occurred first"

\subsubsection*{Datasets}
We use three independent breast cancer datasets collected at Swedish clinical centers. In total, the three datasets comprise 5{,}434 patients, each with a corresponding H\&E-stained WSI and clinical follow-up data. WSIs were generated at 40$\times$ magnification from archived clinical routine resected tumor slides, using Hamamatsu NanoZoomer (S360 or XR) or Aperio GT 450 DX whole-slide scanners.

\paragraph{SöS-BC-4}
This is a retrospective observational cohort including patients diagnosed at Södersjukhuset (South General Hospital) in Stockholm, Sweden between 2012 and 2018~\citep{wang2022improved, sharma2024development}, with clinical outcome data retrieved from the Swedish National Registry for Breast Cancer (NKBC) in 2025. We use a subset of 2{,}315 patients with available follow-up data. This subset contains 351 RFS events and 144 PFS events, with a mean follow-up time of 7.7 years.

% (CHIME-Breast-KS-Solna-survival2025 (survival data updated in 2025))
\paragraph{KS-Solna}
CHIME breast KS-Solna is a population-representative retrospective cohort of primary breast cancer patients treated at the Karolinska University Hospital in Stockholm and diagnosed between 2009 and 2018~\citep{sharma2024validation}, with clinical outcome data retrieved from NKBC in 2025. We use a subset of 1{,}857 patients with available follow-up data, including 389 RFS events and 145 PFS events. The mean follow-up time is 9.1 years.

% (SCAN-B-Lund-survival2025 (survival data updated in 2025))
\paragraph{SCAN-B-Lund}
This is a subset of 1{,}262 patients enrolled in the prospective SCAN-B study~\citep{vallon2019cross}, diagnosed between 2010 and 2019 in Lund, Sweden~\citep{sharma2024development, sharma2024validation}. Clinical follow-up data was updated in 2025. The subset includes 226 RFS events and 88 PFS events, with a mean follow-up time of 8.1 years.

\paragraph{Train \& Evaluation Sets}
Models are trained using SöS-BC-4 and evaluated on the combined KS-Solna + SCAN-B-Lund datasets. This combined evaluation set contains 3{,}119 patients with 615 RFS events and 233 PFS events. Models are evaluated both on the full set and on the `ER+ \& HER2-' patient subgroup, consisting of 2{,}524 patients (80.9\% of the full set) with 475 RFS events (77.2\%) and 157 PFS events (67.4\%). All models are trained on the full set of 2{,}315 patients in SöS-BC-4.

\subsubsection*{WSI Processing \& Model Overview}
Each WSI is preprocessed using a standardized workflow. Tissue regions are first identified using Otsu's thresholding~\citep{otsu1979}, after which non-overlapping image patches of size $256 \times 256$ pixels are extracted at a resolution of $0.4536\,\mu$m/pixel (corresponding to 20$\times$ equivalent magnification for a reference slide scanner). Blurry patches are removed using the variance of Laplacian (VL) metric~\citep{pech2000diatom}, discarding patches with $\mathrm{VL} < 300$. All patches are also color normalized using the Macenko method~\citep{macenko2009method}, adapted for WSI-level color correction following \citet{wang2022improved}.

After preprocessing, each WSI $x$ is represented as a set of $P$ image patches $\{\tilde{x}_i\}_{i=1}^P$, where $P$ varies between slides. A pretrained PFM is then used as a frozen feature extractor to compute a feature vector $f(\tilde{x}_i)$ for each patch $\tilde{x}_i$. These patch-level feature vectors $\{f(\tilde{x}_i)\}_{i=1}^P$ are used as input to a slide-level survival model, which aggregates the patch-level features into a single WSI-level feature vector $f(x)$ and predicts a patient risk score $r(x)$.

\subsubsection*{Survival Model}
We perform slide-level survival prediction using PANTHER~\citep{song2024morphological}, a prototype-based aggregation approach for constructing compact, fixed-length slide representations from patch-level features. Given the set of $P$ patch-level feature vectors $\{f(\tilde{x}_i)\}_{i=1}^P$ extracted from a WSI $x$, PANTHER fits a Gaussian mixture model to estimate a small set of $C = 16$ morphological prototypes, where each prototype represents a recurring morphological pattern within the tissue.

The parameters of the mixture model ($C$ mixture probabilities, means, and diagonal covariance matrices) summarize the distribution of patch-level features and are concatenated to form a WSI-level representation $f(x)$ of dimension $D_{WSI} = C(1 + 2D_p)$, where $D_p$ is the patch-level feature dimension. For example, for a PFM with $D_p = 1536$, the resulting representation $f(x)$ has dimension $D_{WSI} = 49{,}168$. The dimensionality $D_{WSI}$ is constant across WSIs and independent of the number of patches $P$ extracted from each slide.

This WSI-level feature vector $f(x)$ captures both the appearance of morphological patterns and their relative abundance within the WSI $x$, and is used as input to a downstream survival predictor. The survival prediction model is implemented as a structured multilayer perceptron (MLP) head that processes each prototype representation independently before combining them into a final risk prediction, outputting a continuous risk score $r(x)$ for each patient.

We train separate survival models for RFS and PFS. Training is performed using the SöS-BC-4 dataset with random 5-fold cross-validation to determine the number of training epochs for each evaluated feature extractor. This is the only hyperparameter that is tuned individually for each feature extractor, while all other survival model hyperparameters are kept fixed to ensure a fair comparison. After selecting the number of training epochs, the model is retrained on the full SöS-BC-4 dataset. The AdamW optimizer~\citep{loshchilov2018decoupled} with a cosine learning rate schedule is used, optimizing the Cox proportional hazards loss~\citep{katzman2018deepsurv}.

To improve stability, both the prototype estimation and survival model training of PANTHER are repeated with five random seeds, and the resulting survival prediction models are ensembled by averaging standardized risk scores across seeds. The feature extractors are kept frozen, and only the MLP head of the survival model is updated during training.

% ("For training, we use weight decay of 1 × 10−5 and AdamW optimizer with a learning rate of 1 × 10−4 with the cosine decay scheduler")
% ("we use Cox proportional hazards loss [48] with a batch size of 64 patients")

% (PANTHER MLP model. Separately trained/evaluated models for RFS and PFS. Optimizing the number of epochs on SoS-BC-4-survival2025 random 5-fold cross-validation for each FM. Repeating both the clustering and model training on the full SoS-BC-4-survival2025 dataset with 5 random seeds. Creating an ensemble of these 5 trained models (using risk score normalized with the train mean/std). Bootstrapping)

% (IndivMLPEmb_SharedIndiv PANTHER model)

% (max number of sampled tiles per prototype in the clustering code: 2.5e+05, same for all models)

\subsubsection*{Evaluation Framework}
Model performance is primarily evaluated using the concordance index (C-index)~\citep{harrell1982evaluating}, which measures the agreement between predicted risk scores and observed survival times. Higher C-index values indicate better alignment between predicted risk ordering and actual patient outcomes. To estimate uncertainty, C-index values are computed using 10{,}000 bootstrap resamples with 95\% confidence intervals. We also perform Kaplan-Meier (KM)~\citep{kaplan1958nonparametric} survival analysis to assess the ability of models to stratify patients into two or more risk groups (Figure~\ref{fig:km_plots_2groups_counts}~\&~\ref{fig:km_plots_4groups}). Patients are divided into risk groups based on predicted risk scores, and differences between groups are assessed using the log-rank test.

All models are evaluated under four complementary settings: (1) RFS for the full patient cohort (`\textit{All Patients}'). (2) RFS for the subgroup of patients which are oestrogen receptor (ER)-positive and human epidermal growth factor receptor 2 (HER2)-negative (`\textit{ER+ \& HER2-}'). (3) PFS for `All Patients'. (4) PFS for `ER+ \& HER2-'. The patient subgroup `ER+ \& HER2-' represents the most common breast cancer subtype and has distinct biological and clinical characteristics, making it an important subgroup for prognostic modeling. We train separate survival models for RFS and PFS, but all models are trained on the full SöS-BC-4 dataset and then evaluated for both `All Patients' and `ER+ \& HER2-' on the combined KS-Solna + SCAN-B-Lund evaluation dataset. 

Models are ranked separately within each of the four evaluation settings based on their bootstrap mean C-index. The final overall model ranking (Table~\ref{table:main_results_rank}) is computed as the mean rank across the four settings, providing a robust comparison of model performance across both survival endpoints and patient populations.

In addition, we evaluate the effect of training data size by training survival models on random subsets of the SöS-BC-4 dataset comprising 75\%, 50\%, 25\%, and 10\% of the training data (Figure~\ref{fig:subsets_h1_uni_resnetin}~\&~\ref{fig:subsets_h0_h1_h0-mini}). For each subset size, results are reported as the mean $\pm$ standard deviation over five random seeds, where a new random subset of training patients is sampled for each seed. The PANTHER prototype estimation step is performed once on the full SöS-BC-4 dataset for each seed, after which the survival model is retrained using the corresponding sampled data subset. This analysis provides insight into the data efficiency and robustness of different PFMs.

\subsubsection*{Evaluated Models}
We evaluate thirteen pretrained patch-level feature extractors spanning multiple generations of representation learning in computational pathology. Table~\ref{table:fms} summarizes the architecture, parameter count, feature dimensionality, and pretraining datasets of all evaluated models.

\paragraph{Natural-Image Baseline}
We include Resnet-IN in the evaluation, a Resnet-50~\citep{he2016deep} model pretrained on the ImageNet dataset~\citep{russakovsky2015imagenet} of natural images, which has historically been used in early computational pathology pipelines. Resnet-IN serves as a simple reference baseline to quantify the benefit of large-scale pathology-specific pretraining, and is expected to be significantly outperformed by PFMs.

\paragraph{Early Pathology Models}
We also evaluate two early pathology-specific models trained directly on histopathology images using self-supervised learning: CTransPath~\citep{Wang2023ctranspath} and RetCCL~\citep{wang2023retccl}. Both models were trained on approximately 30{,}000 WSIs, which is substantially smaller than the datasets used to train state-of-the-art PFMs. These models represent an intermediate stage between generic natural-image encoders and more recent large-scale PFMs.

\paragraph{State-of-the-art PFMs}
The majority of evaluated models are PFMs trained on large collections of WSIs (ranging from 100{,}000 to 3.1 million WSIs) using self-supervised learning. These include Prov-GigaPath~\citep{gigapath2024}, UNI~\citep{chen2024uni}, Virchow~\citep{virchow2024} and H-optimus-0~\citep{hoptimus0}, as well as their more recent second-generation counterparts UNI2-h~\citep{UNI2h2024}, Virchow2~\citep{virchow22024}, and H-optimus-1~\citep{hoptimus1}. The second-generation models are trained on substantially larger datasets and incorporate improved training strategies, reflecting recent advances in large-scale representation learning for computational pathology.

\paragraph{Distilled Model}
We further evaluate H0-mini~\citep{filiot2025h0mini}, a compact distilled variant of H-optimus-0 designed to significantly reduce model size and computational cost while preserving representation quality. H0-mini is a ViT-Base vision transformer~\citep{dosovitskiy2021an} with 86 million parameters, corresponding to less than 8\% of the original ViT-Giant H-optimus-0 (1.1 billion parameters). The distillation is performed on a dataset of 43 million image patches extracted from 6{,}093 WSIs from the public TCGA dataset, covering 16 cancer types. This evaluation allows us to assess whether compact distilled models can retain the performance of substantially larger PFMs.

\paragraph{Vision-Language PFMs}
Finally, we include two multimodal vision-language PFMs, CONCH~\citep{conch2024} and its second-generation version CONCHv1.5~\citep{ding2025multimodal}. These models are trained using paired image-text supervision in addition to histopathology WSIs, representing an alternative training paradigm that leverages multimodal data with the aim of learning more generalizable representations.

\subsubsection*{Ethics Statement}
The study has approval from the Swedish Ethical Review Authority (2017/2106-31, with amendments 2018/1462-32 and 2019–02336). The study was performed in accordance with the Declaration of Helsinki. No additional informed consent was required in accordance with ethical approval in this non-interventional collection and analysis of data from patient records.
\section*{Data Availability}

The SöS-BC-4, KS-Solna, and SCAN-B-Lund datasets cannot be made publicly available due to restrictions relating to sensitive patient-related information.
\section*{Code Availability}

The code for this study is based on PANTHER which is available at \url{https://github.com/mahmoodlab/Panther}. Further implementation details are available from FKG upon reasonable request.
\section*{Acknowledgments}

The project was supported by funding from the Swedish Cancer Society (23 2905 Pj 01 H), VINNOVA (SwAIPP2), Swedish e-science Research Centre (SeRC) (eMPHasis project), Bröstcancerförbundet, Swedish Research Council (2024–06634, 2025-03411), and the AID4BC consortium supported by a WASP/DDLS NEST grant (KAW 2024.0159) from the Knut \& Alice Wallenberg Foundation to SciLifeLab for research in Data-Driven Life Science (DDLS) and the Wallenberg AI, Autonomous Systems and Software Program (WASP). 

DAC was funded by an NIHR Research Professorship; a Royal Academy of Engineering Research Chair; and the InnoHK Hong Kong Centre for Cerebro-cardiovascular Engineering (COCHE); and was supported by the National Institute for Health and Care Research (NIHR) Oxford Biomedical Research Centre (BRC) and the Pandemic Sciences Institute at the University of Oxford.

The authors acknowledge patients, clinicians, and hospital staff participating in the SCAN-B study; the staff at the central SCAN-B laboratory at the Division of Oncology, Lund University; the Swedish National Breast Cancer Quality Registry (NKBC); the Regional Cancer Center South; and the South Swedish Breast Cancer Group (SSBCG). SCAN-B was funded by the Swedish Cancer Society, the Mrs. Berta Kamprad Foundation, the Lund-Lausanne L2-Bridge/Biltema Foundation, the Mats Paulsson Foundation, and Swedish governmental funding (ALF).
\section*{Author Contributions}

FKG was responsible for project conceptualization, software implementation, preparation of figures and tables, and manuscript drafting. CB processed the survival datasets. JVC supported the use of the SCAN-B-Lund dataset. DAC acquired funding. MR supervised the project, contributed to the design of experiments and interpretation of results, and acquired funding. All authors contributed to manuscript revision and finalization.

\section*{Competing Interests}

MR is a co-founder and shareholder of Stratipath AB. All other authors declare no competing interests.
\section*{Declaration of Generative AI Use}

During the preparation of this manuscript, the authors used ChatGPT 5.3 to assist with language editing and drafting. All content produced using this tool was critically reviewed, edited and validated by the authors, who take full responsibility for the final content of the manuscript.

{
    \small
    \bibliographystyle{plainnat}
    \bibliography{references}
}

\clearpage
\appendix
\onecolumn

\renewcommand{\thefigure}{S\arabic{figure}}
\setcounter{figure}{0}

\renewcommand{\thetable}{S\arabic{table}}
\setcounter{table}{0}

\renewcommand{\theequation}{S\arabic{equation}}
\setcounter{equation}{0}

\subsection*{\centering{Benchmarking Pathology Foundation Models\\ for Breast Cancer Survival Prediction}}
\section*{\centering{Supplementary Material}}

\vspace{6.0mm}

\section{Supplementary Figures}
\label{appendix:figures}

\begin{figure*}[h]
\centering
    \begin{subfigure}[t]{0.495\textwidth}
        \centering%
        \includegraphics[clip, trim=0.35cm 1.25cm 0.25cm 0.0cm, width=0.95\linewidth]{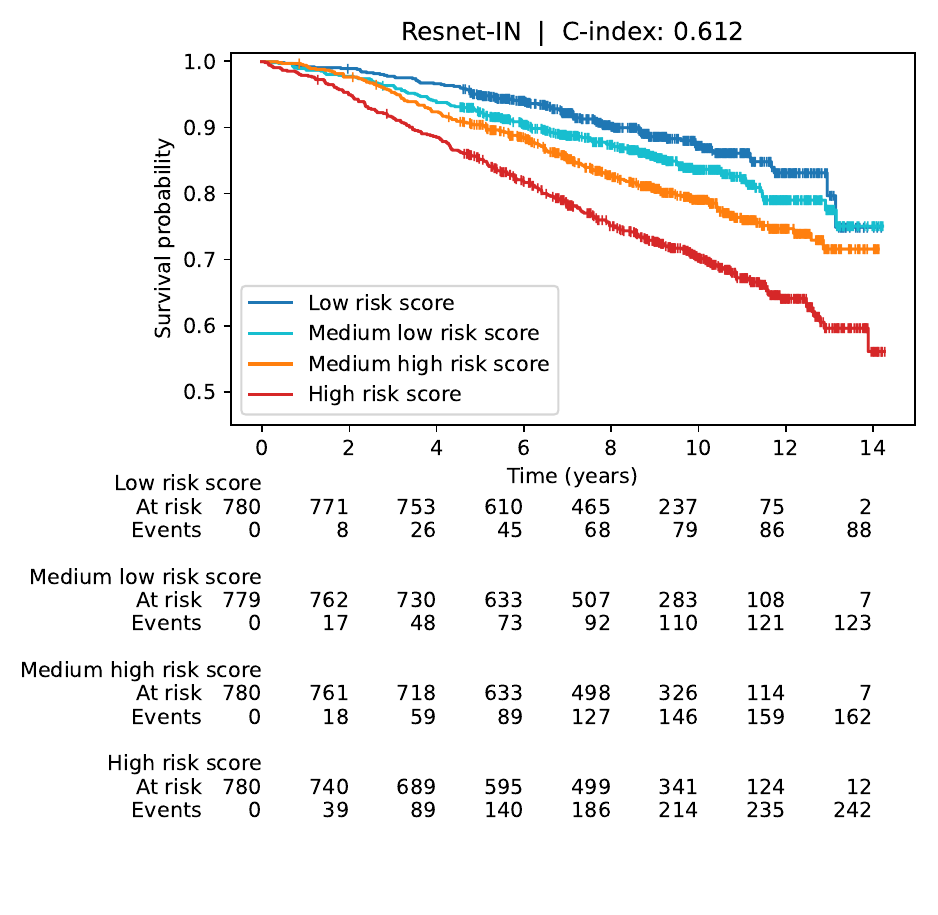}
        \caption{\textbf{RFS -- All Patients}.}\vspace{2.0mm}
        \label{fig:km_plots_rfs_4groups_counts_all-patients_resnet-in}
    \end{subfigure}
    \begin{subfigure}[t]{0.495\textwidth}
        \centering%
        \includegraphics[clip, trim=0.35cm 1.25cm 0.25cm 0.0cm, width=0.95\linewidth]{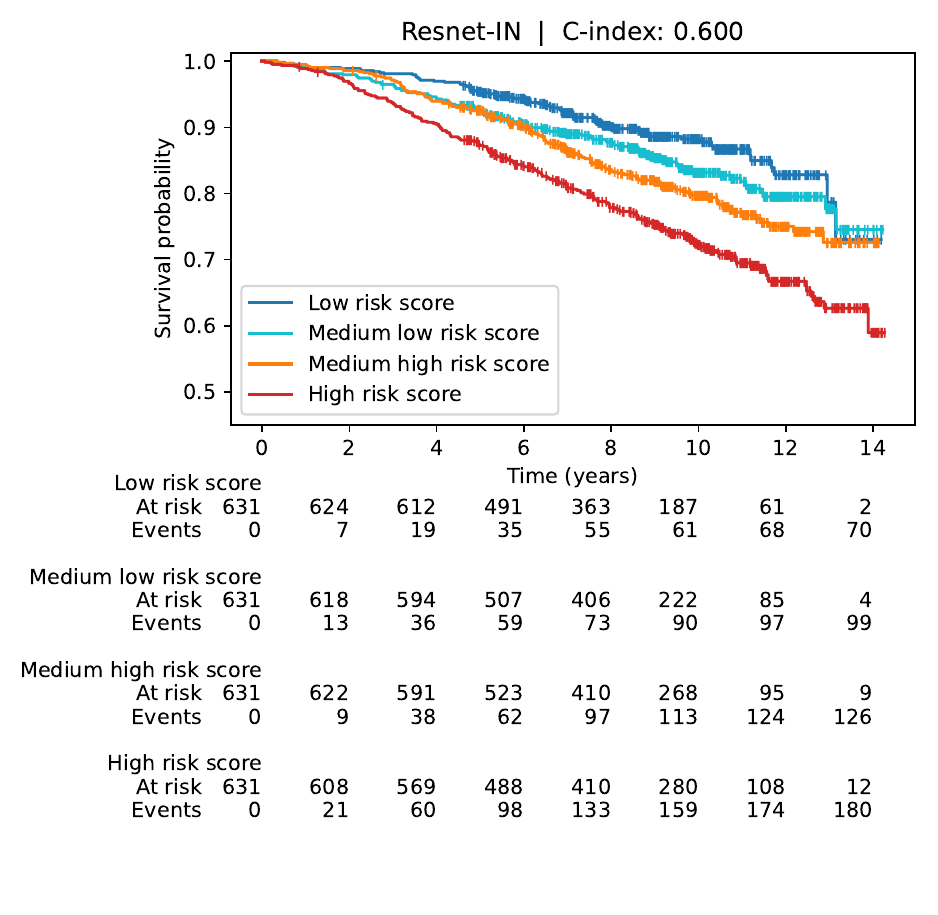}
        \caption{\textbf{RFS -- Patient Subgroup (ER+ \& HER2-)}.}\vspace{2.0mm}
        \label{fig:km_plots_rfs_4groups_counts_subgroup_resnet-in}
    \end{subfigure}
    \begin{subfigure}[t]{0.495\textwidth}
        \centering%
        \includegraphics[clip, trim=0.35cm 1.25cm 0.25cm 0.0cm, width=0.95\linewidth]{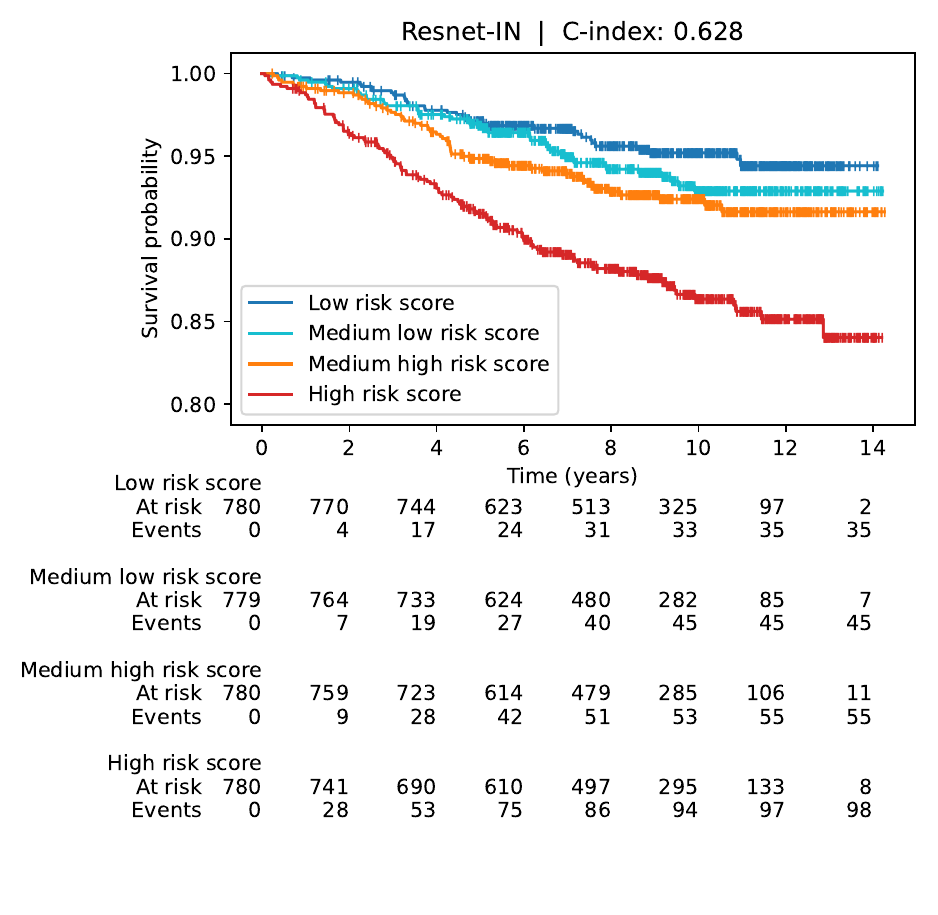}
        \caption{\textbf{PFS -- All Patients}.}
        \label{fig:km_plots_pfs_4groups_counts_all-patients_resnet-in}
    \end{subfigure}
    \begin{subfigure}[t]{0.495\textwidth}
        \centering%
        \includegraphics[clip, trim=0.35cm 1.25cm 0.25cm 0.0cm, width=0.95\linewidth]{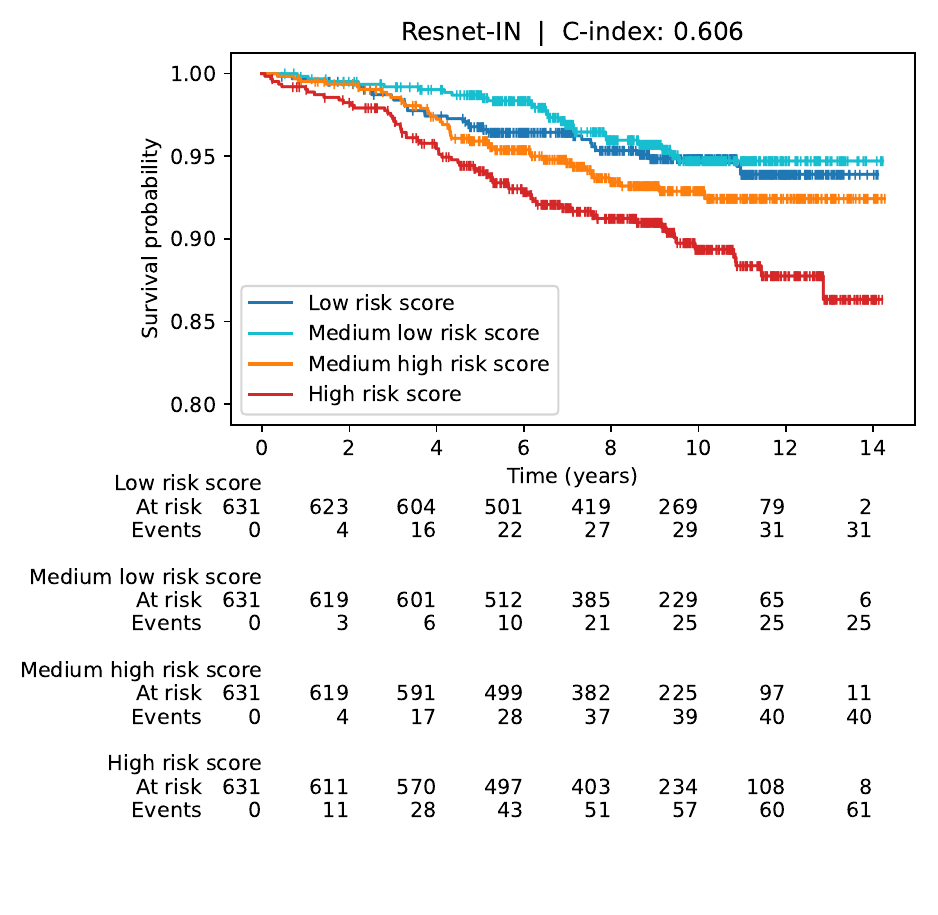}
        \caption{\textbf{PFS -- Patient Subgroup (ER+ \& HER2-)}.}
        \label{fig:km_plots_pfs_4groups_counts_subgroup_resnet-in}
    \end{subfigure}
    \caption{\textbf{Four-group Kaplan-Meier risk stratification with at-risk and event counts (Resnet-IN)}. KM survival curves corresponding to Figure~\ref{fig:km_plots_4groups} for \textit{Resnet-IN}, including the number of patients at risk and the number of events over time (0-14 years) for each of the four risk groups. Note the difference in range of the y-axis between RFS and PFS.}
  \label{fig:km_plots_4groups_counts_resnet-in}
\end{figure*}

\begin{figure*}[h]
\centering
    \begin{subfigure}[t]{0.495\textwidth}
        \centering%
        \includegraphics[clip, trim=0.35cm 1.25cm 0.25cm 0.0cm, width=0.95\linewidth]{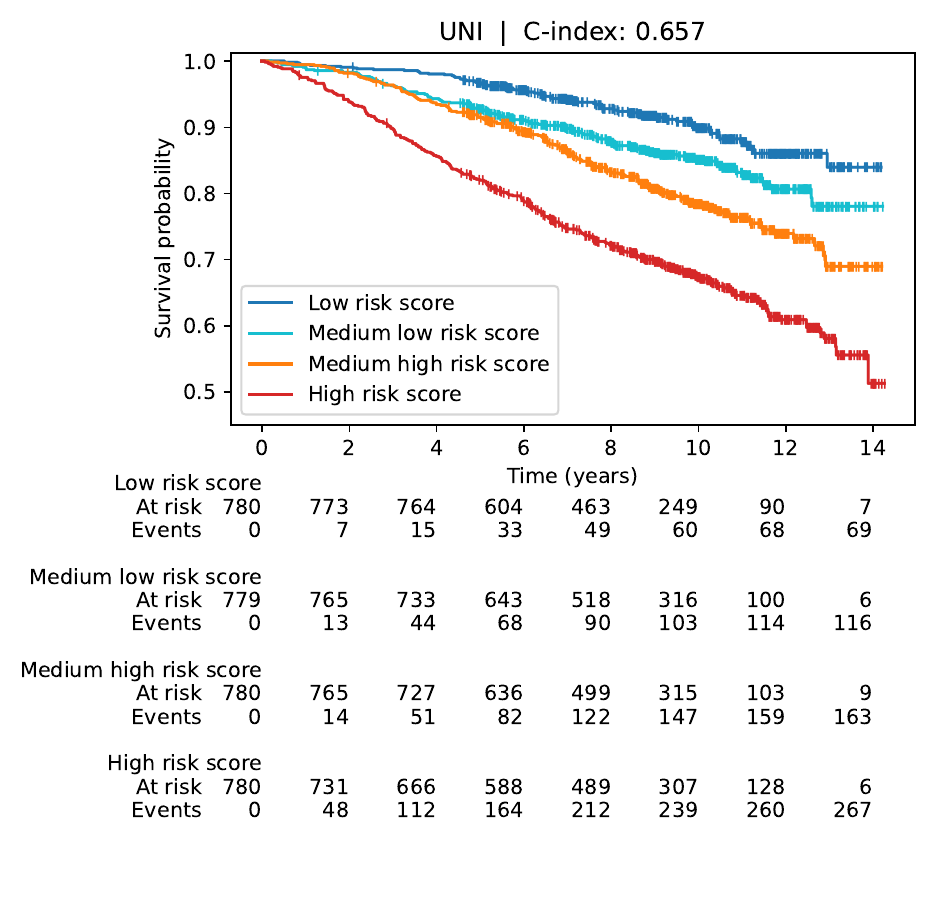}
        \caption{\textbf{RFS -- All Patients}.}\vspace{2.0mm}
        \label{fig:km_plots_rfs_4groups_counts_all-patients_uni}
    \end{subfigure}
    \begin{subfigure}[t]{0.495\textwidth}
        \centering%
        \includegraphics[clip, trim=0.35cm 1.25cm 0.25cm 0.0cm, width=0.95\linewidth]{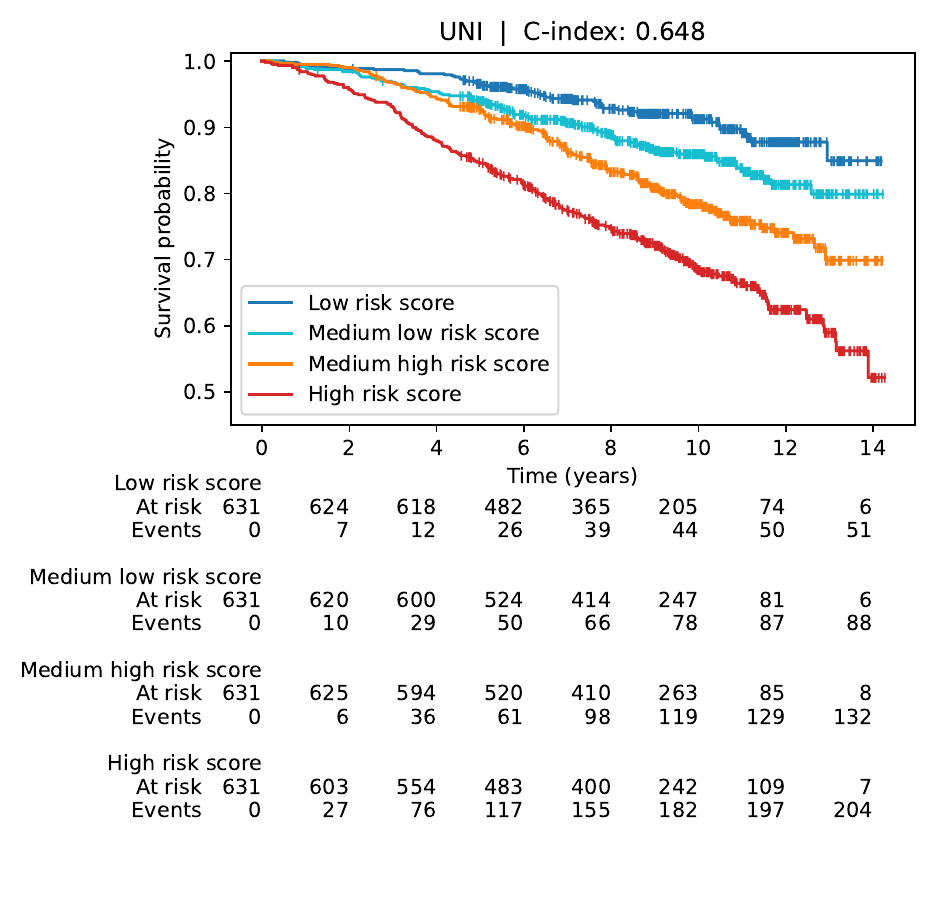}
        \caption{\textbf{RFS -- Patient Subgroup (ER+ \& HER2-)}.}\vspace{2.0mm}
        \label{fig:km_plots_rfs_4groups_counts_subgroup_uni}
    \end{subfigure}
    \begin{subfigure}[t]{0.495\textwidth}
        \centering%
        \includegraphics[clip, trim=0.35cm 1.25cm 0.25cm 0.0cm, width=0.95\linewidth]{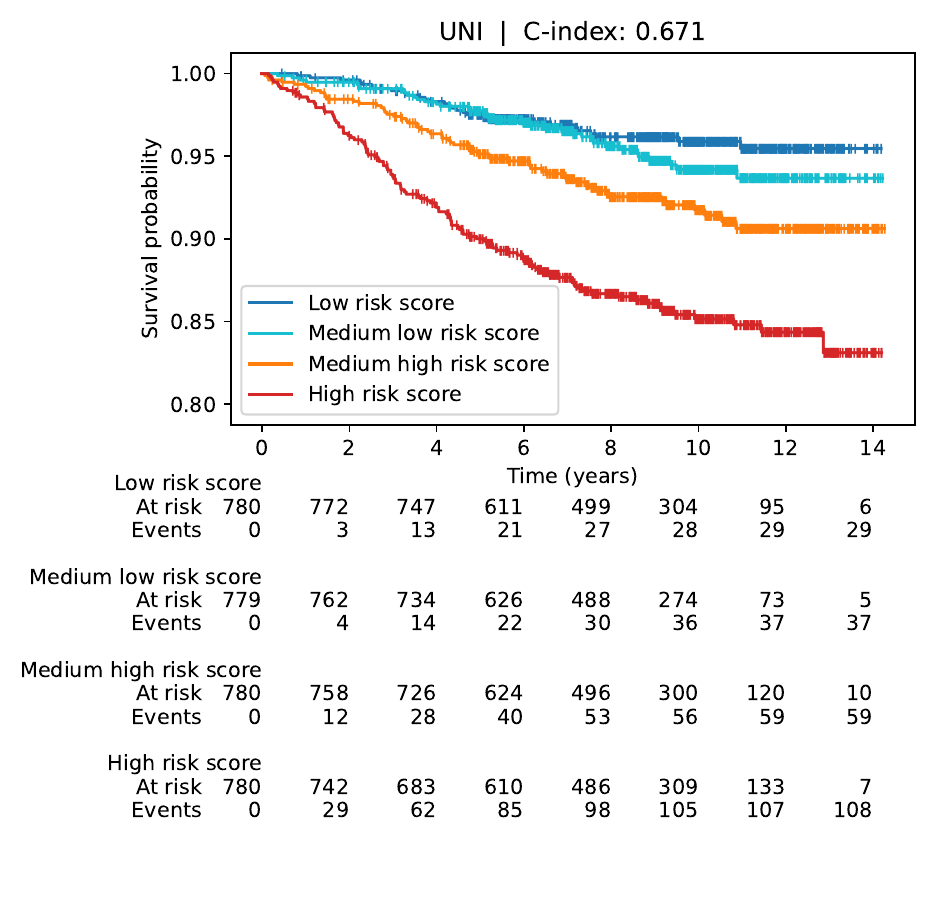}
        \caption{\textbf{PFS -- All Patients}.}
        \label{fig:km_plots_pfs_4groups_counts_all-patients_uni}
    \end{subfigure}
    \begin{subfigure}[t]{0.495\textwidth}
        \centering%
        \includegraphics[clip, trim=0.35cm 1.25cm 0.25cm 0.0cm, width=0.95\linewidth]{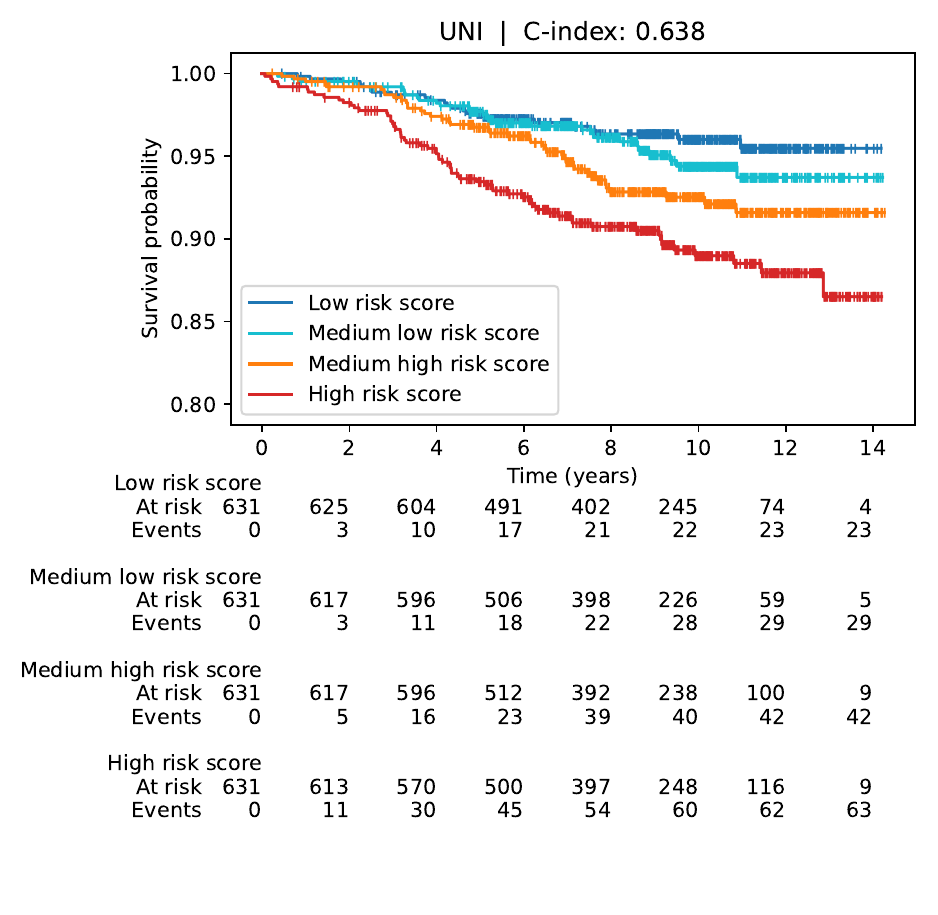}
        \caption{\textbf{PFS -- Patient Subgroup (ER+ \& HER2-)}.}
        \label{fig:km_plots_pfs_4groups_counts_subgroup_uni}
    \end{subfigure}
    \caption{\textbf{Four-group Kaplan-Meier risk stratification with at-risk and event counts (UNI)}. KM survival curves corresponding to Figure~\ref{fig:km_plots_4groups} for \textit{UNI}, including the number of patients at risk and the number of events over time (0-14 years) for each of the four risk groups. Note the difference in range of the y-axis between RFS and PFS.}
  \label{fig:km_plots_4groups_counts_uni}
\end{figure*}

\begin{figure*}[h]
\centering
    \begin{subfigure}[t]{0.495\textwidth}
        \centering%
        \includegraphics[clip, trim=0.35cm 1.25cm 0.25cm 0.0cm, width=0.95\linewidth]{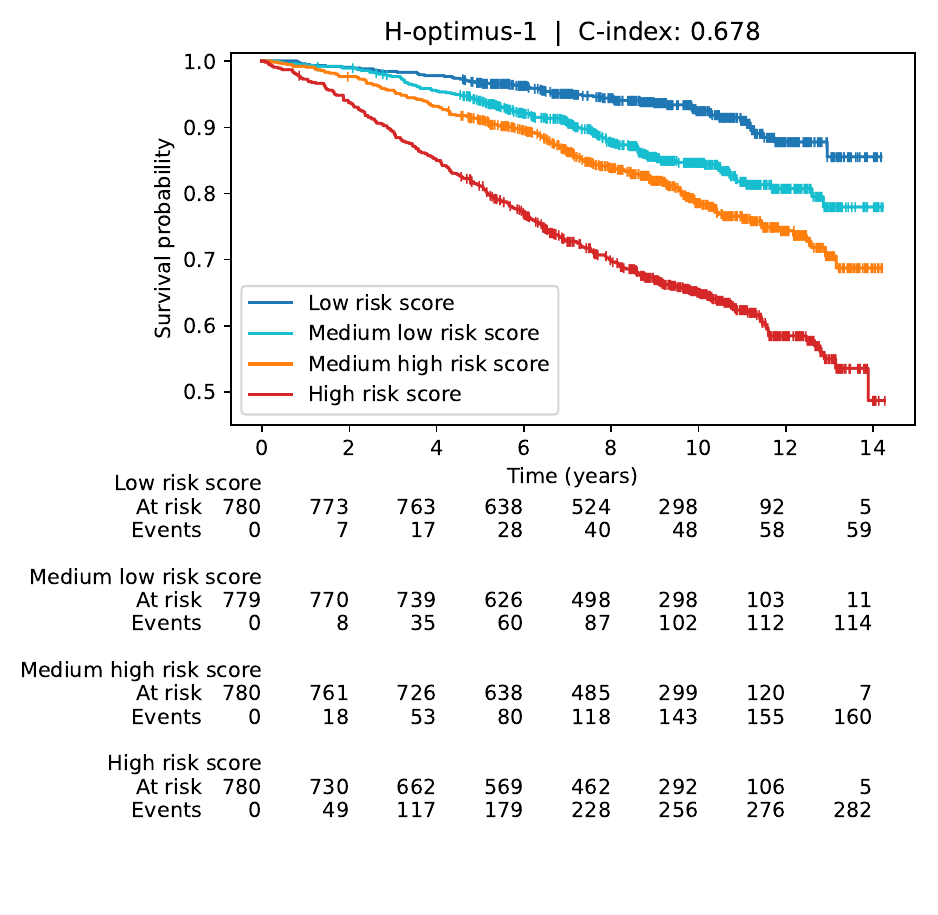}
        \caption{\textbf{RFS -- All Patients}.}\vspace{2.0mm}
        \label{fig:km_plots_rfs_4groups_counts_all-patients_h-optimus-1}
    \end{subfigure}
    \begin{subfigure}[t]{0.495\textwidth}
        \centering%
        \includegraphics[clip, trim=0.35cm 1.25cm 0.25cm 0.0cm, width=0.95\linewidth]{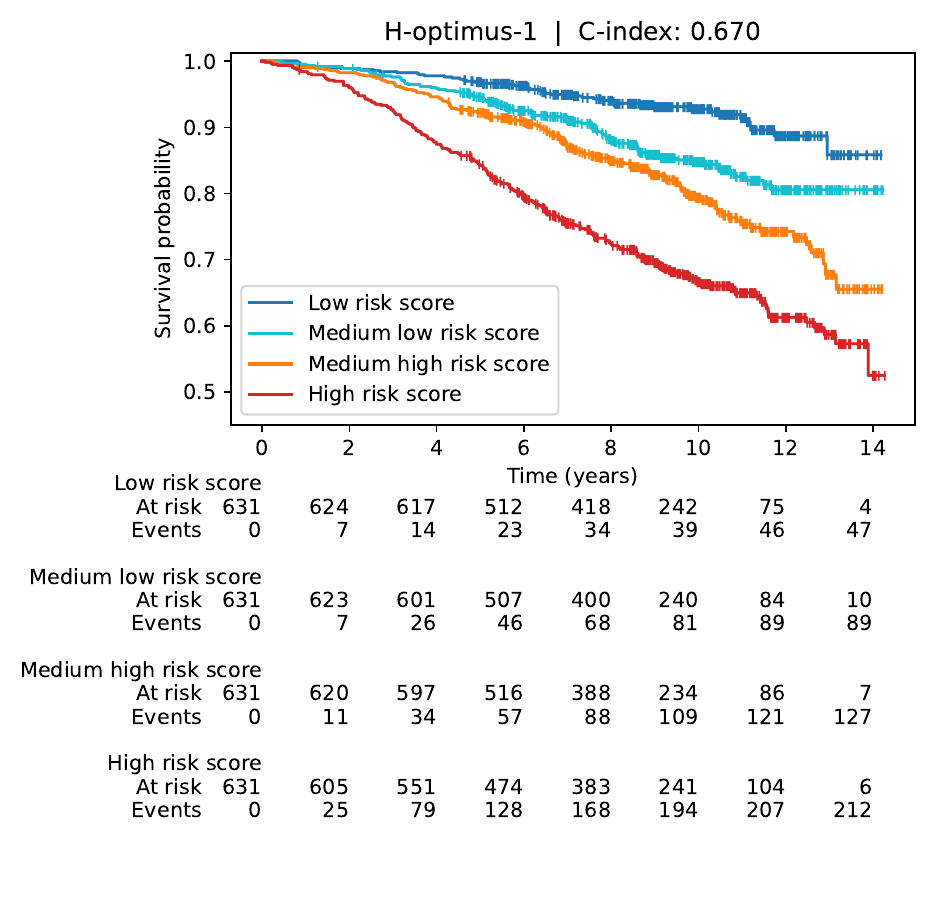}
        \caption{\textbf{RFS -- Patient Subgroup (ER+ \& HER2-)}.}\vspace{2.0mm}
        \label{fig:km_plots_rfs_4groups_counts_subgroup_h-optimus-1}
    \end{subfigure}
    \begin{subfigure}[t]{0.495\textwidth}
        \centering%
        \includegraphics[clip, trim=0.35cm 1.25cm 0.25cm 0.0cm, width=0.95\linewidth]{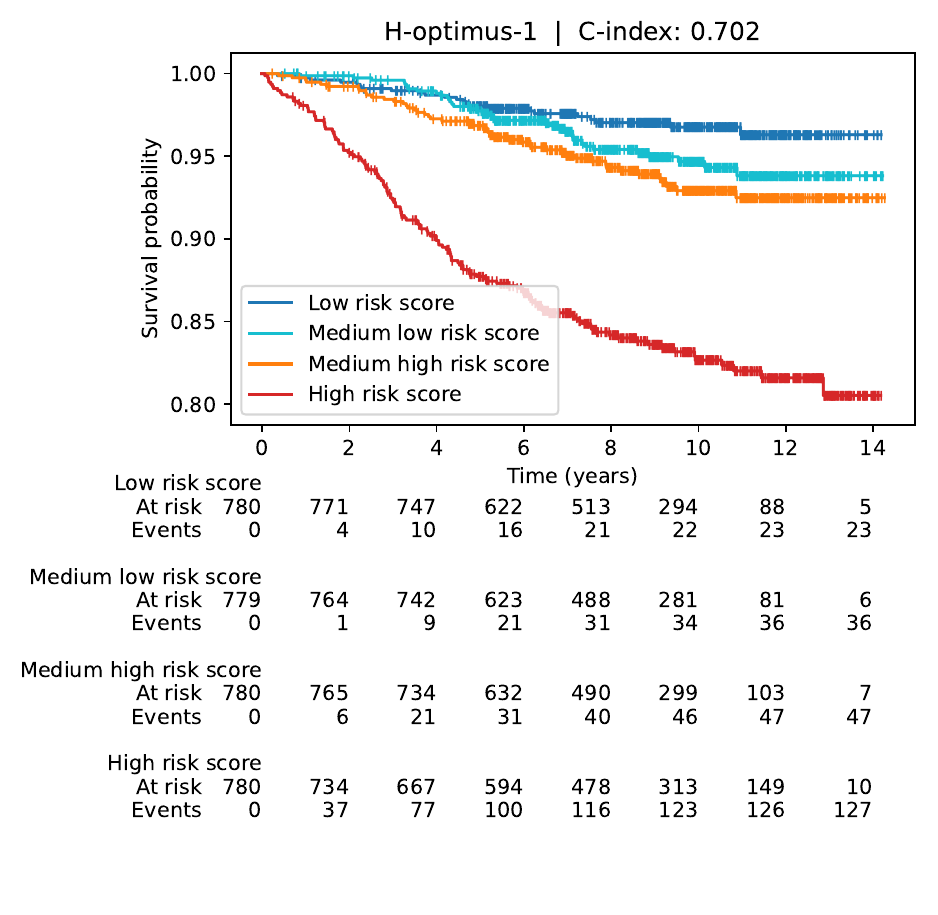}
        \caption{\textbf{PFS -- All Patients}.}
        \label{fig:km_plots_pfs_4groups_counts_all-patients_h-optimus-1}
    \end{subfigure}
    \begin{subfigure}[t]{0.495\textwidth}
        \centering%
        \includegraphics[clip, trim=0.35cm 1.25cm 0.25cm 0.0cm, width=0.95\linewidth]{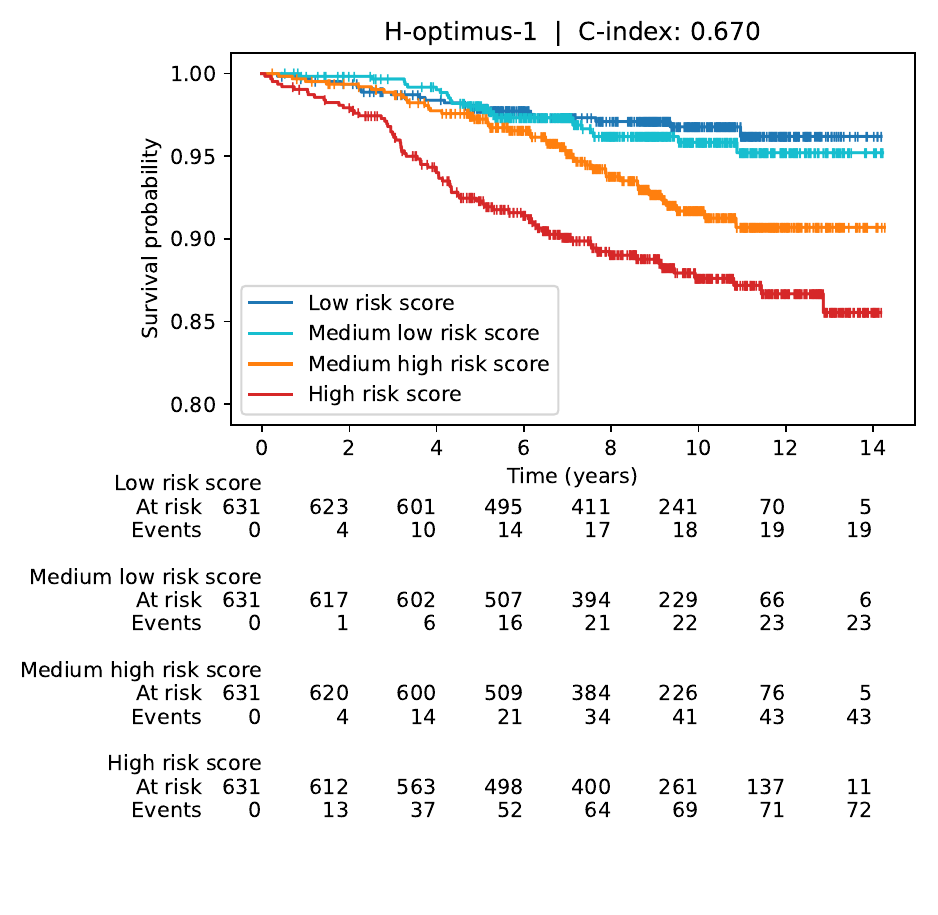}
        \caption{\textbf{PFS -- Patient Subgroup (ER+ \& HER2-)}.}
        \label{fig:km_plots_pfs_4groups_counts_subgroup_h-optimus-1}
    \end{subfigure}
    \caption{\textbf{Four-group Kaplan-Meier risk stratification with at-risk and event counts (H-optimus-1)}. KM survival curves corresponding to Figure~\ref{fig:km_plots_4groups} for \textit{H-optimus-1}, including the number of patients at risk and the number of events over time (0-14 years) for each of the four risk groups. Note the difference in range of the y-axis between RFS and PFS.}
  \label{fig:km_plots_4groups_counts_h-optimus-1}
\end{figure*}

\begin{figure*}[h]
\centering
    \begin{subfigure}[t]{0.495\textwidth}
        \centering%
        \includegraphics[clip, trim=0.4cm 1.25cm 0.25cm 0.0cm, width=0.925\linewidth]{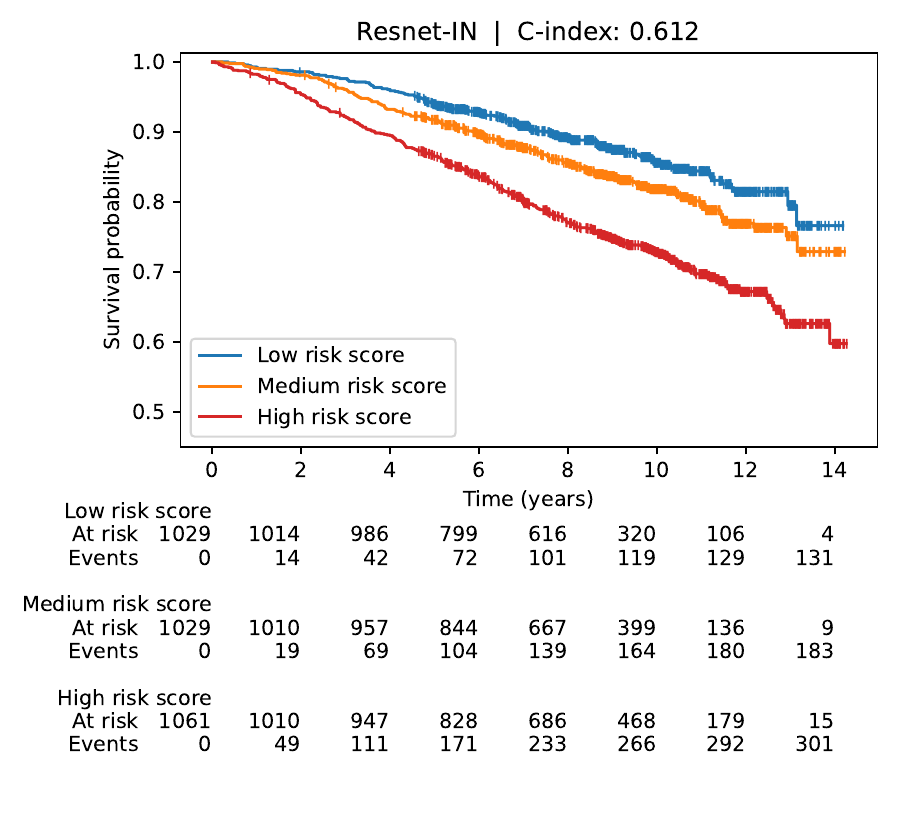}
        \caption{\textbf{RFS -- All Patients}.}\vspace{2.0mm}
        \label{fig:km_plots_rfs_3groups_counts_all-patients_resnet-in}
    \end{subfigure}
    \begin{subfigure}[t]{0.495\textwidth}
        \centering%
        \includegraphics[clip, trim=0.4cm 1.25cm 0.25cm 0.0cm, width=0.925\linewidth]{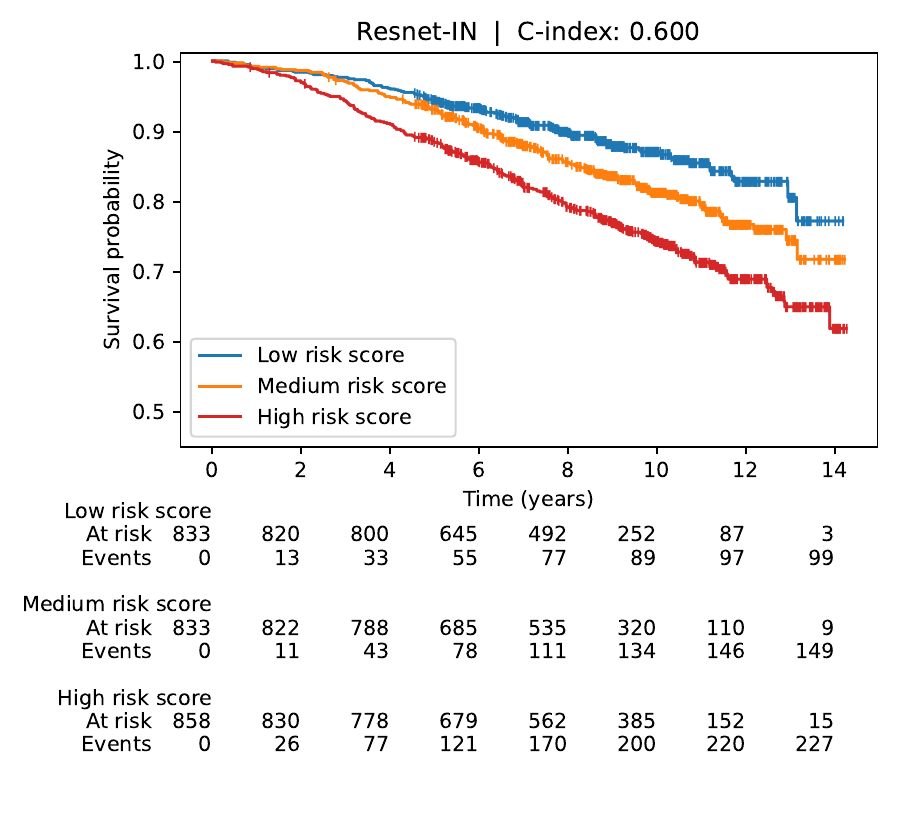}
        \caption{\textbf{RFS -- Patient Subgroup (ER+ \& HER2-)}.}\vspace{2.0mm}
        \label{fig:km_plots_rfs_3groups_counts_subgroup_resnet-in}
    \end{subfigure}
    \begin{subfigure}[t]{0.495\textwidth}
        \centering%
        \includegraphics[clip, trim=0.4cm 1.25cm 0.25cm 0.0cm, width=0.925\linewidth]{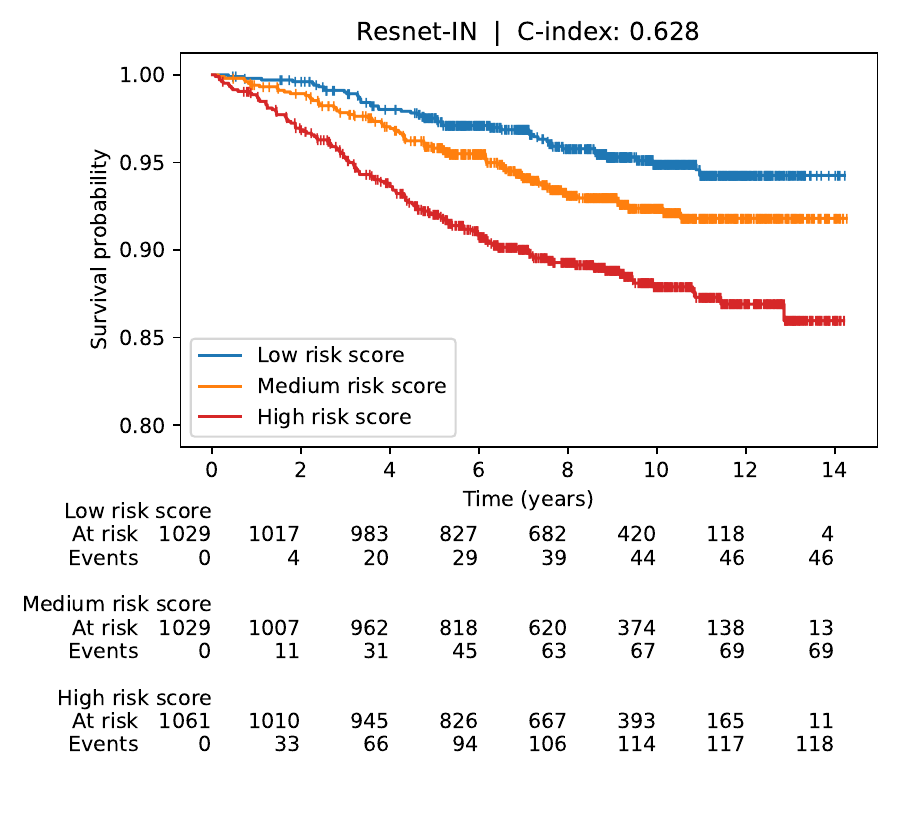}
        \caption{\textbf{PFS -- All Patients}.}
        \label{fig:km_plots_pfs_3groups_counts_all-patients_resnet-in}
    \end{subfigure}
    \begin{subfigure}[t]{0.495\textwidth}
        \centering%
        \includegraphics[clip, trim=0.4cm 1.25cm 0.25cm 0.0cm, width=0.925\linewidth]{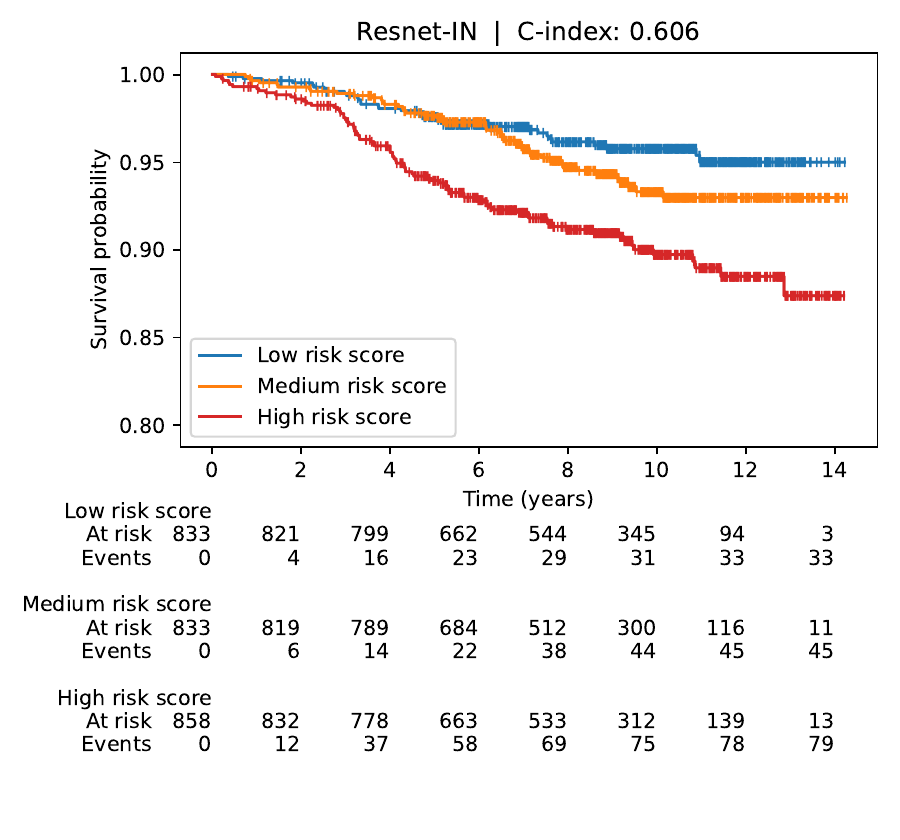}
        \caption{\textbf{PFS -- Patient Subgroup (ER+ \& HER2-)}.}
        \label{fig:km_plots_pfs_3groups_counts_subgroup_resnet-in}
    \end{subfigure}
    \caption{\textbf{Three-group Kaplan-Meier risk stratification with at-risk and event counts (Resnet-IN)}. KM survival curves corresponding to Figure~\ref{fig:km_plots_4groups_counts_resnet-in}, but with stratification into three risk groups instead of four, for \textit{Resnet-IN}. The plots include the number of patients at risk and the number of events over time (0-14 years) for each risk group. Note the difference in range of the y-axis between RFS and PFS.}
  \label{fig:km_plots_3groups_counts_resnet-in}
\end{figure*}

\begin{figure*}[h]
\centering
    \begin{subfigure}[t]{0.495\textwidth}
        \centering%
        \includegraphics[clip, trim=0.4cm 1.25cm 0.25cm 0.0cm, width=0.925\linewidth]{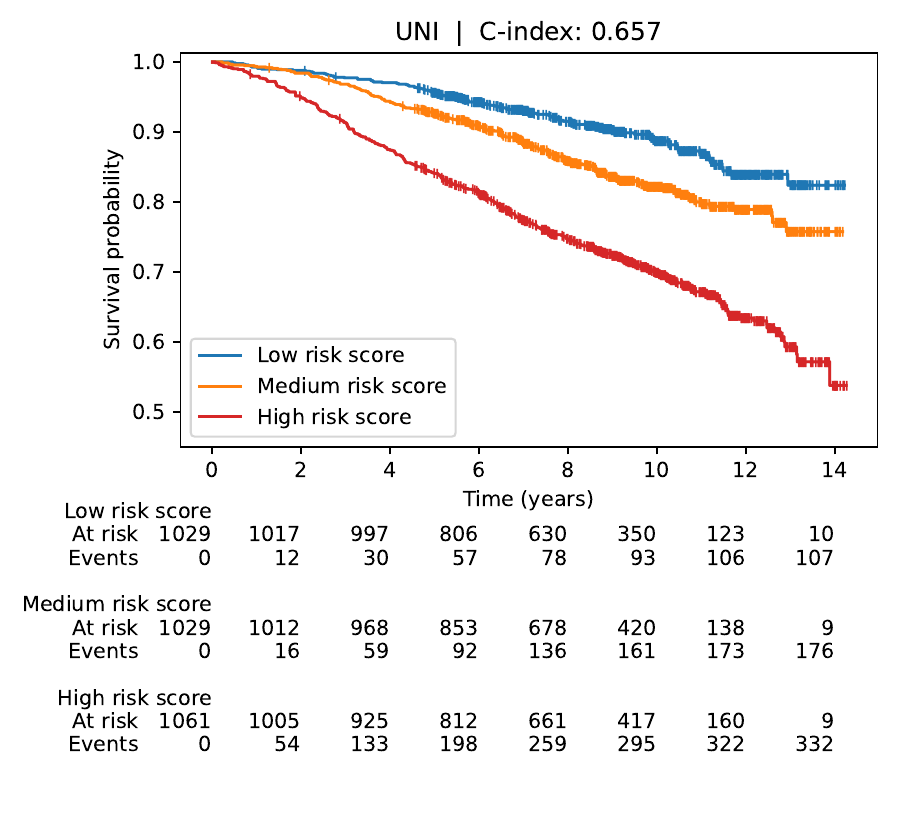}
        \caption{\textbf{RFS -- All Patients}.}\vspace{2.0mm}
        \label{fig:km_plots_rfs_3groups_counts_all-patients_uni}
    \end{subfigure}
    \begin{subfigure}[t]{0.495\textwidth}
        \centering%
        \includegraphics[clip, trim=0.4cm 1.25cm 0.25cm 0.0cm, width=0.925\linewidth]{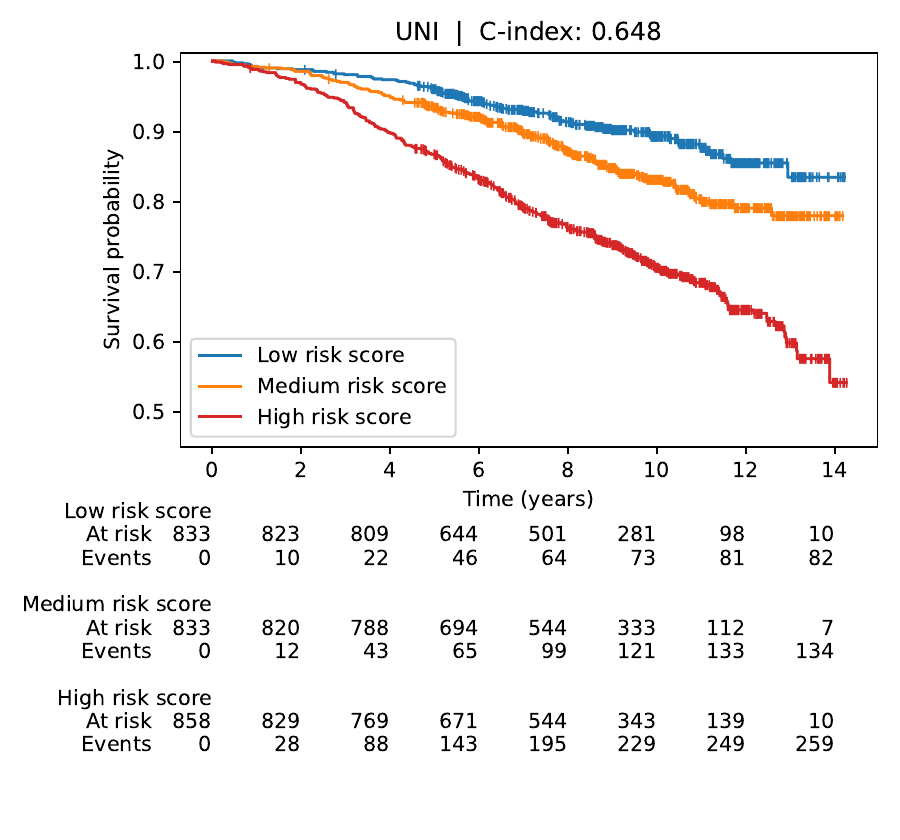}
        \caption{\textbf{RFS -- Patient Subgroup (ER+ \& HER2-)}.}\vspace{2.0mm}
        \label{fig:km_plots_rfs_3groups_counts_subgroup_uni}
    \end{subfigure}
    \begin{subfigure}[t]{0.495\textwidth}
        \centering%
        \includegraphics[clip, trim=0.4cm 1.25cm 0.25cm 0.0cm, width=0.925\linewidth]{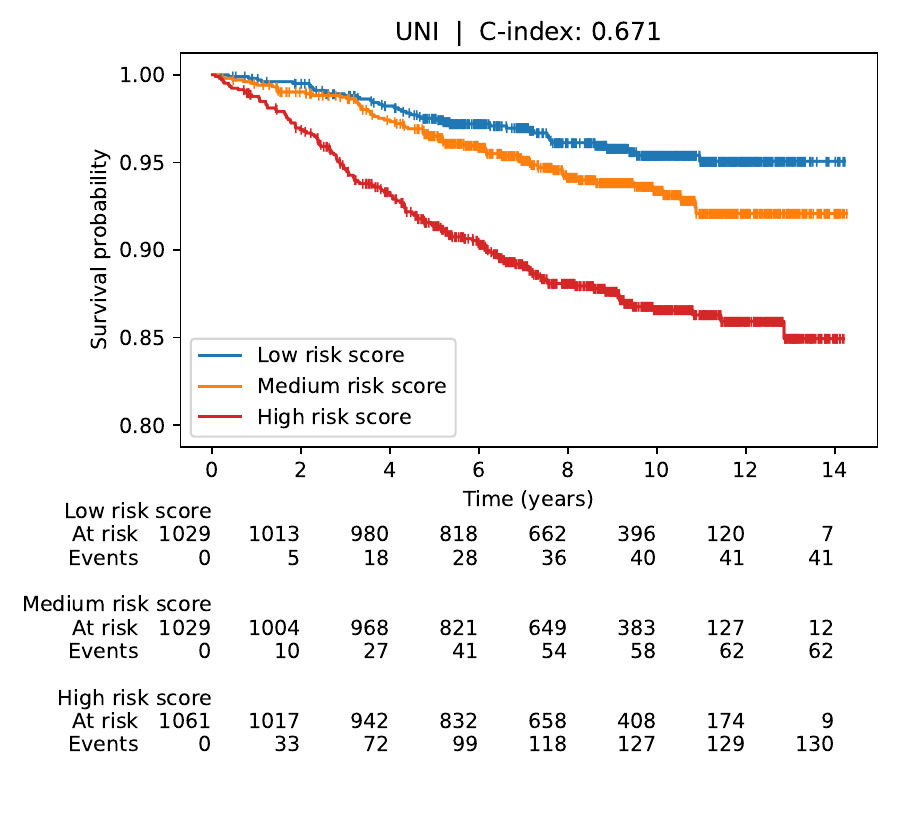}
        \caption{\textbf{PFS -- All Patients}.}
        \label{fig:km_plots_pfs_3groups_counts_all-patients_uni}
    \end{subfigure}
    \begin{subfigure}[t]{0.495\textwidth}
        \centering%
        \includegraphics[clip, trim=0.4cm 1.25cm 0.25cm 0.0cm, width=0.925\linewidth]{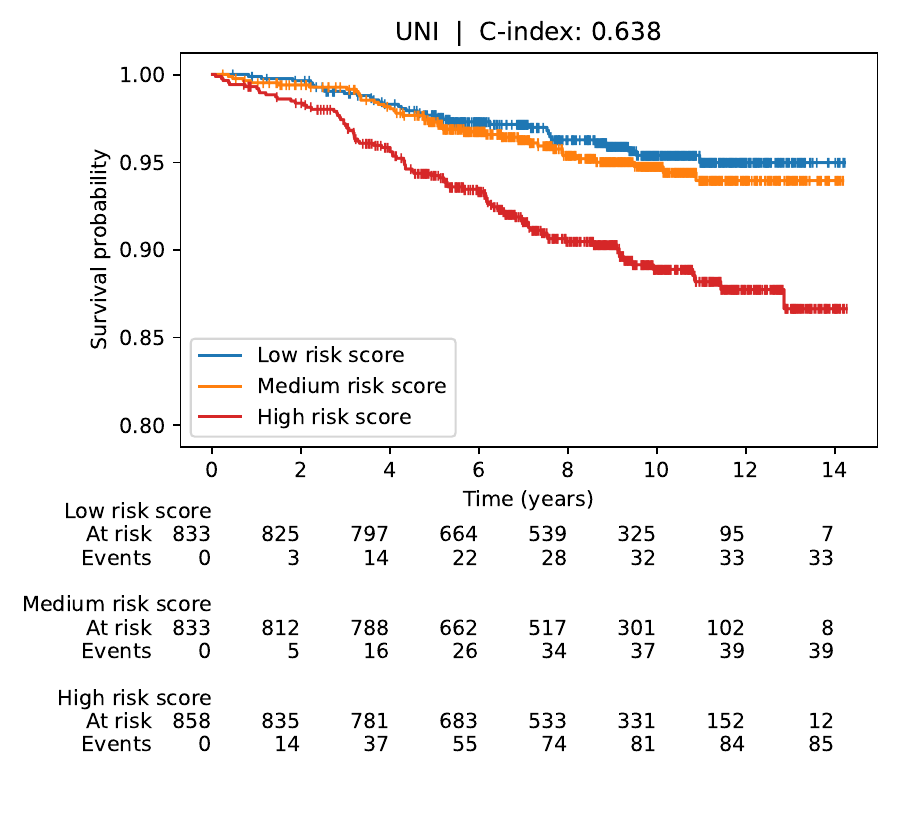}
        \caption{\textbf{PFS -- Patient Subgroup (ER+ \& HER2-)}.}
        \label{fig:km_plots_pfs_3groups_counts_subgroup_uni}
    \end{subfigure}
    \caption{\textbf{Three-group Kaplan-Meier risk stratification with at-risk and event counts (UNI)}. KM survival curves corresponding to Figure~\ref{fig:km_plots_4groups_counts_uni}, but with stratification into three risk groups instead of four, for \textit{UNI}. The plots include the number of patients at risk and the number of events over time (0-14 years) for each risk group. Note the difference in range of the y-axis between RFS and PFS.}
  \label{fig:km_plots_3groups_counts_uni}
\end{figure*}

\begin{figure*}[h]
\centering
    \begin{subfigure}[t]{0.495\textwidth}
        \centering%
        \includegraphics[clip, trim=0.4cm 1.25cm 0.25cm 0.0cm, width=0.925\linewidth]{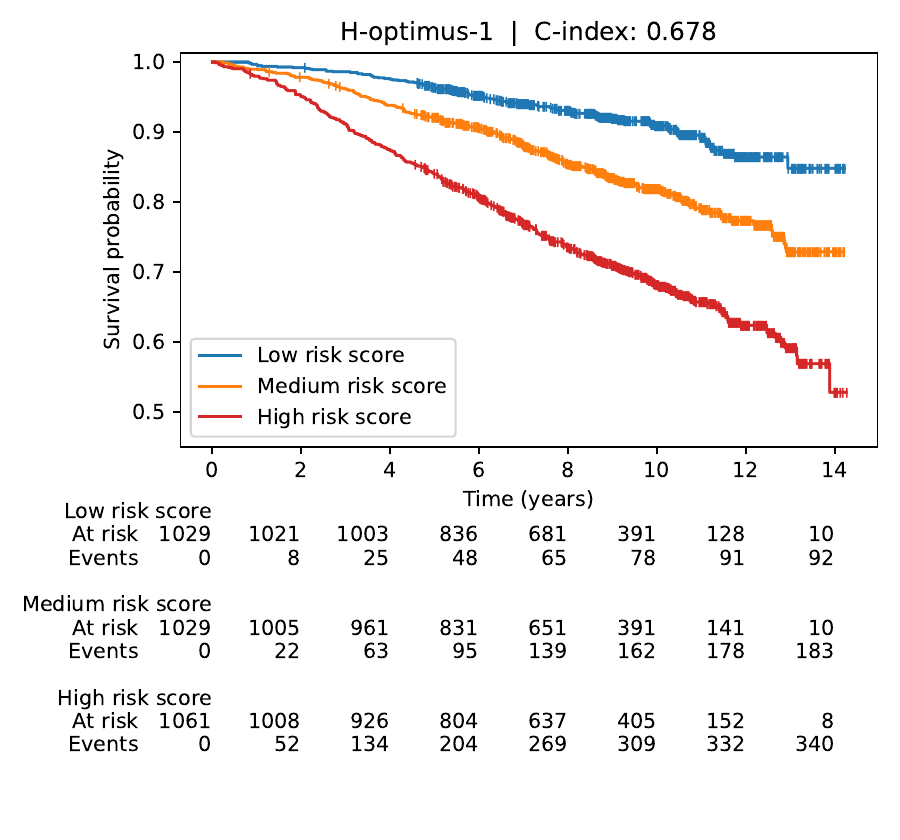}
        \caption{\textbf{RFS -- All Patients}.}\vspace{2.0mm}
        \label{fig:km_plots_rfs_3groups_counts_all-patients_h-optimus-1}
    \end{subfigure}
    \begin{subfigure}[t]{0.495\textwidth}
        \centering%
        \includegraphics[clip, trim=0.4cm 1.25cm 0.25cm 0.0cm, width=0.925\linewidth]{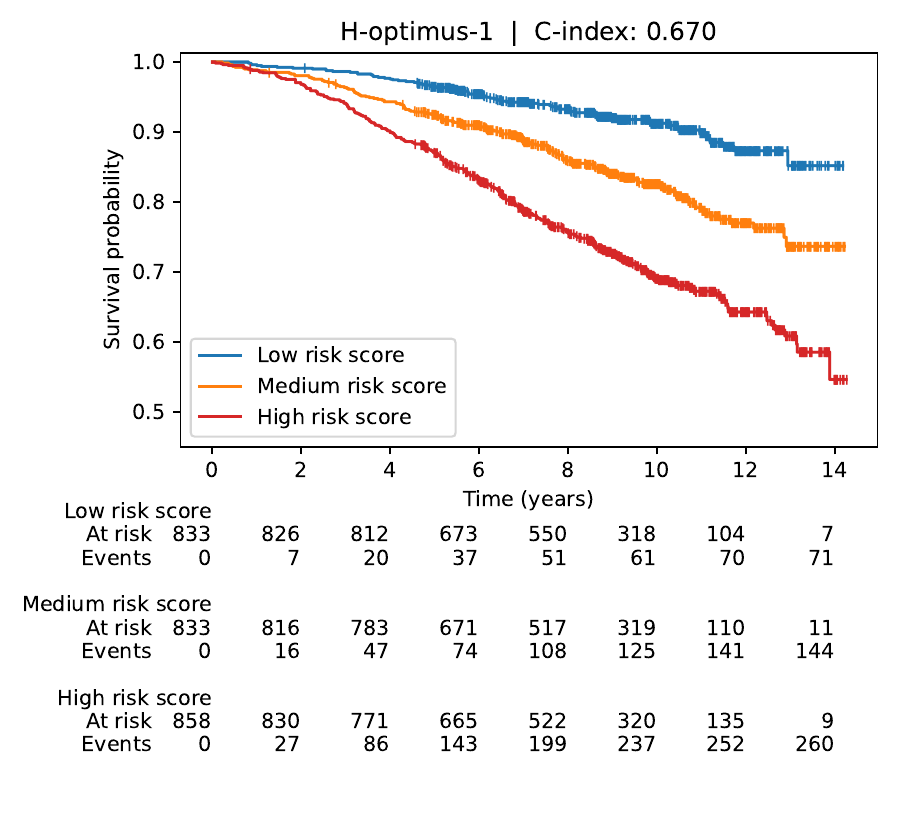}
        \caption{\textbf{RFS -- Patient Subgroup (ER+ \& HER2-)}.}\vspace{2.0mm}
        \label{fig:km_plots_rfs_3groups_counts_subgroup_h-optimus-1}
    \end{subfigure}
    \begin{subfigure}[t]{0.495\textwidth}
        \centering%
        \includegraphics[clip, trim=0.4cm 1.25cm 0.25cm 0.0cm, width=0.925\linewidth]{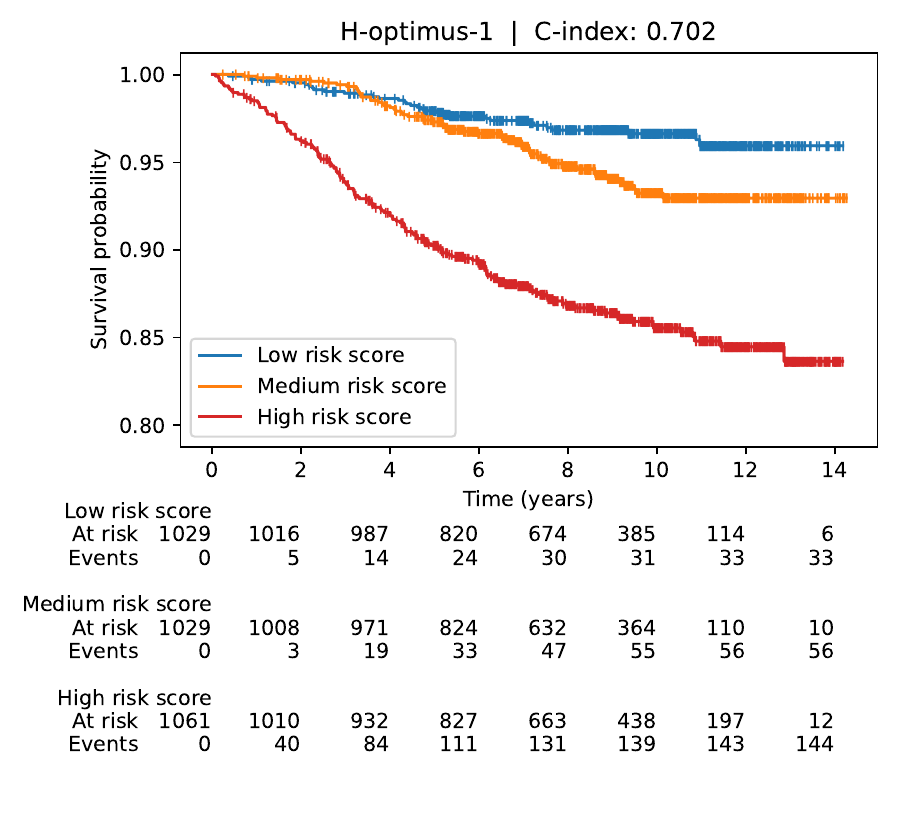}
        \caption{\textbf{PFS -- All Patients}.}
        \label{fig:km_plots_pfs_3groups_counts_all-patients_h-optimus-1}
    \end{subfigure}
    \begin{subfigure}[t]{0.495\textwidth}
        \centering%
        \includegraphics[clip, trim=0.4cm 1.25cm 0.25cm 0.0cm, width=0.925\linewidth]{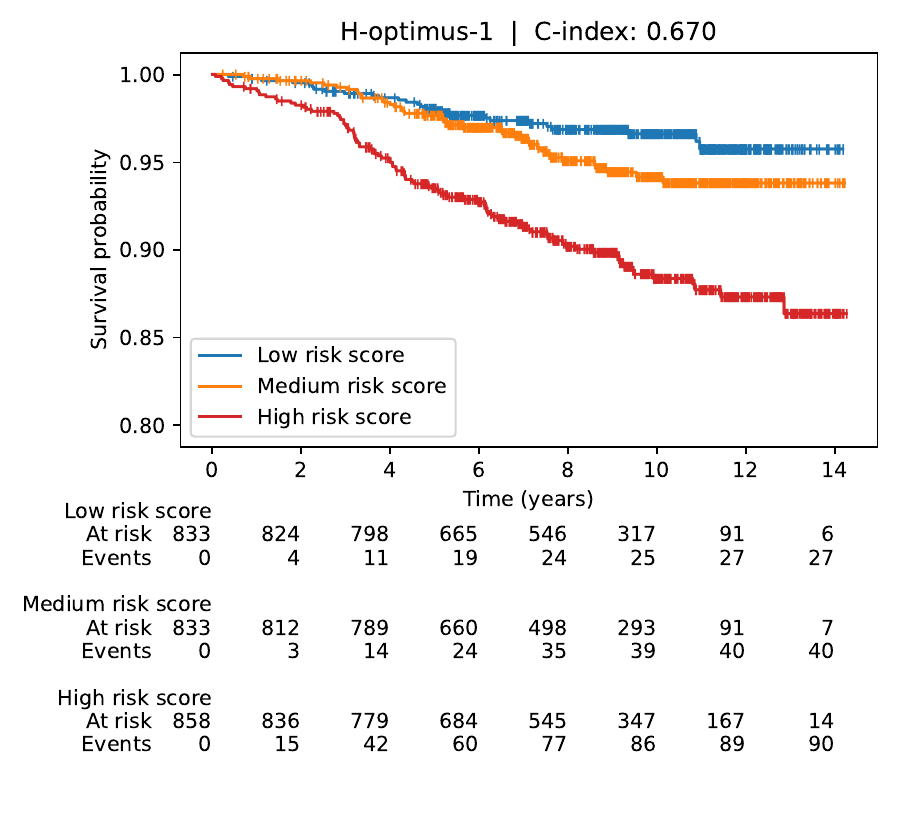}
        \caption{\textbf{PFS -- Patient Subgroup (ER+ \& HER2-)}.}
        \label{fig:km_plots_pfs_3groups_counts_subgroup_h-optimus-1}
    \end{subfigure}
    \caption{\textbf{Three-group Kaplan-Meier risk stratification with at-risk and event counts (H-optimus-1)}. KM survival curves corresponding to Figure~\ref{fig:km_plots_4groups_counts_h-optimus-1}, but with stratification into three risk groups instead of four, for \textit{H-optimus-1}. The plots include the number of patients at risk and the number of events over time for each risk group. Note the difference in range of the y-axis between RFS and PFS.}
  \label{fig:km_plots_3groups_counts_h-optimus-1}
\end{figure*}

This section contains Figure \ref{fig:km_plots_4groups_counts_resnet-in} - \ref{fig:km_plots_3groups_counts_h-optimus-1}.

\vspace{5.0mm}



\end{document}